%% file: TRC2022_Incident_duration_pred.tex
\PassOptionsToPackage{bookmarksnumbered,unicode,colorlinks=true,linktocpage=true}{hyperref}
\documentclass[a4paper,fleqn]{cas-sc}

\usepackage[numbers]{natbib}

\def\tsc#1{\csdef{#1}{\textsc{\lowercase{#1}}\xspace}}
\tsc{WGM}
\tsc{QE}
\tsc{EP}
\tsc{PMS}
\tsc{BEC}
\tsc{DE}

\usepackage{url}
\usepackage[nameinlink,capitalize]{cleveref}
\graphicspath{{Figures_final//}}

\usepackage{caption}
\usepackage{subcaption}

\DeclareGraphicsExtensions{.eps,.pdf,.png,.jpg,.tif,.tiff,.ps}

\usepackage{graphics} 
\usepackage{epsfig} 
\usepackage{float} 
\usepackage{graphicx}
\usepackage[]{algorithm2e}
\usepackage{algpseudocode}
\usepackage[justification=centering]{caption}

\usepackage{stix}

\usepackage{enumitem}
\usepackage{caption}
\usepackage{wrapfig}

\usepackage{adjustbox}

\usepackage{xspace}
\definecolor{navy}{rgb}{0.1, 0.1, 0.8}

\usepackage{soul}
\newcommand{\eat}[1]{}

\newcommand{\revA}[1]{#1}

\newcommand{\verify}[1]{#1}

\usepackage[many]{tcolorbox}

\begin{document}
\let\WriteBookmarks\relax
\def\floatpagepagefraction{1}
\def\textpagefraction{.001}
\shorttitle{Incident duration prediction using Machine Learning}
\shortauthors{Grigorev et~al.}


\title[mode = title]{Incident duration prediction using a bi-level machine learning framework with outlier removal and intra-extra joint optimisation}
                
%
%
%

\author[1]{Artur Grigorev}
\cormark[1]
\fnmark[1]
\ead{Artur.Grigorev@student.uts.edu.au}
\ead[url]{www.fmlab.org}
\credit{Writing - Original Draft, Methodology, Software, Investigation, Conceptualisation}
\address[1]{University of Technology Sydney, 61 Broadway Str, Sydney, Australia}

\author[1]{Adriana-Simona Mihaita}
\credit{Supervision, Project administration, Writing - Review \& Editing, Resources, Conceptualization}

\author[1]{Seunghyeon Lee}
\credit{Writing - Review \& Editing, Supervision}

\author[1]{Fang Chen}
\credit{Supervision}
\cortext[cor1]{Corresponding author}
%

\begin{abstract}
Predicting the duration of traffic incidents is a challenging task due to the stochastic nature of events. The ability to accurately predict how long accidents will last can provide significant benefits to both end-users in their route choice and traffic operation managers in handling of non-recurrent traffic congestion. This paper presents a novel bi-level machine learning framework enhanced with outlier removal and intra-extra joint optimisation for predicting the incident duration on three heterogeneous data sets collected for both arterial roads and motorways from Sydney, Australia and San-Francisco, U.S.A. Firstly, we use incident data logs to develop a binary classification prediction approach, which allows us to classify traffic incidents as short-term or long-term. We find the optimal threshold between short-term versus long-term traffic incident duration, targeting both class balance and prediction performance while also comparing the binary versus multi-class classification approaches \revA{using quantiled duration groups and varying threshold split}. Secondly, for more granularity of the incident duration prediction to the minute level, we propose a new Intra-Extra Joint Optimisation algorithm (IEO-ML) which extends multiple baseline ML models tested against several regression scenarios across the data sets. Final results indicate that: a) 40-45 min is the best split threshold for identifying short versus long-term incidents and that these incidents should be modelled separately, b) our proposed IEO-ML approach significantly outperforms baseline ML models in $66\%$ of all cases showcasing its great potential for accurate incident duration prediction. Lastly, we evaluate the feature importance and show that time, location, incident type, incident reporting source and weather at among the top 10 critical factors which influence how long incidents will last.

\end{abstract}

%


\maketitle



\newtcolorbox{mybox}[3][]
{
  colframe = #2!25,
  colback  = #2!10,
  coltitle = #2!20!black,  
  title    = {#3},
  #1,
}


\input{sections/1-introduction}
\input{sections/2-Data-sources}

\input{sections/3A-Methodology}

\input{sections/3B-ORM-Methodology}
\input{sections/5A-Classification-Results}

\input{sections/5B-Regression-Results}

\input{sections/5C-Regression-ORM}

\input{sections/6-Feature-importance}

\input{sections/7-Conclusions}

\newpage
\appendix
\section{Appendix A}
\label{appendix_A}

Providing additional results for the threshold variation along all data sets such as (Accuracy, Precision and Recall).  

\begin{figure}[h!]
\centering
\includegraphics[width=0.98\textwidth]{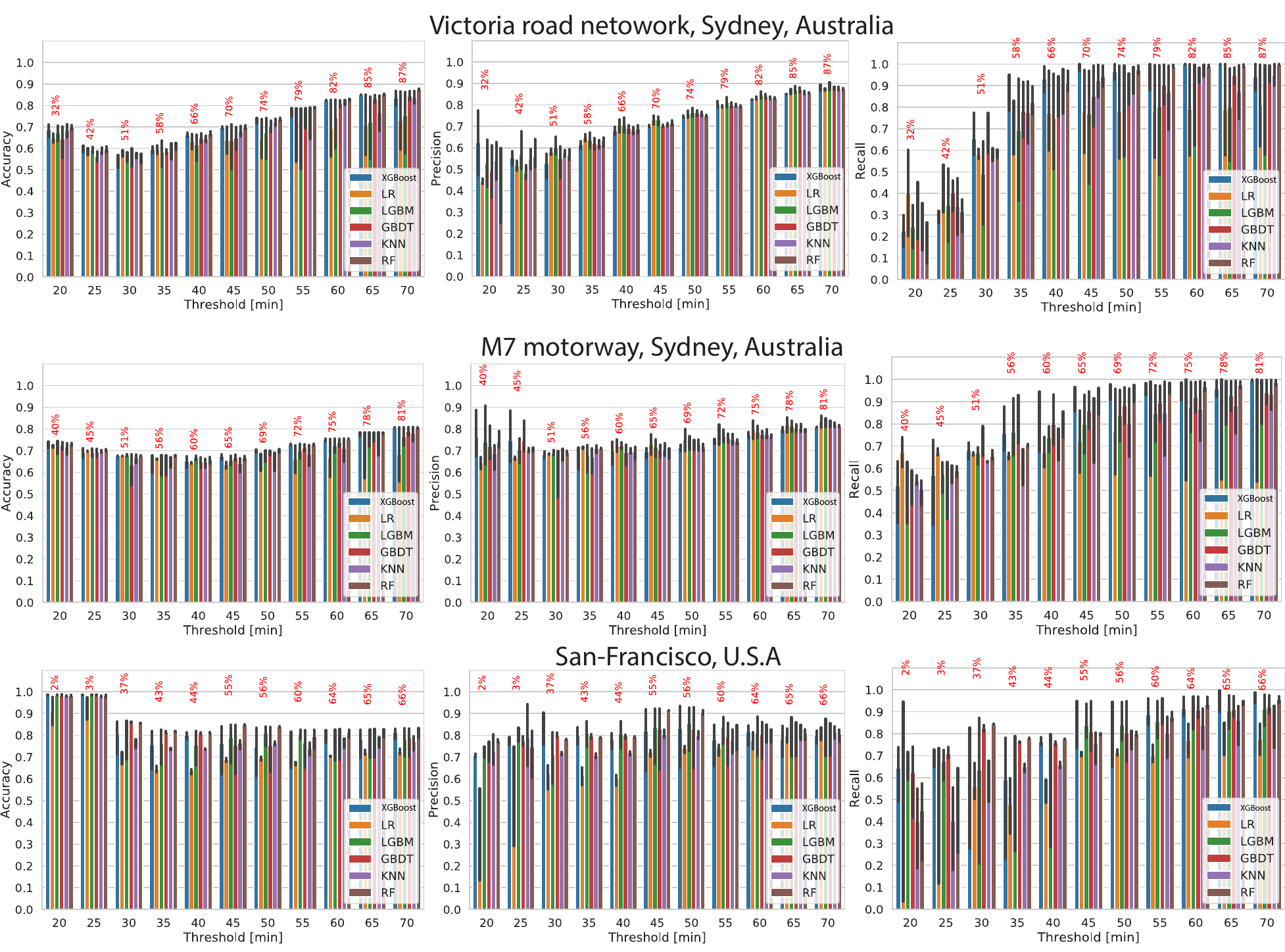}
\centering
\caption{Binary classification performance using varying incident duration threshold}
\label{fig:classif2}
\end{figure}

\newpage
\section{Appendix B}
\label{appendix_B}

Providing additional information with regards to the computational time of various baseline ML models across the three data sets. The findings indicate the RF and kNN seem to be the slowest models to train versus LGBM and XGBoost and LR which are faster from a computational time point of view. 
\begin{figure}[pos=h]
\centering
\includegraphics[width=0.32\textwidth]{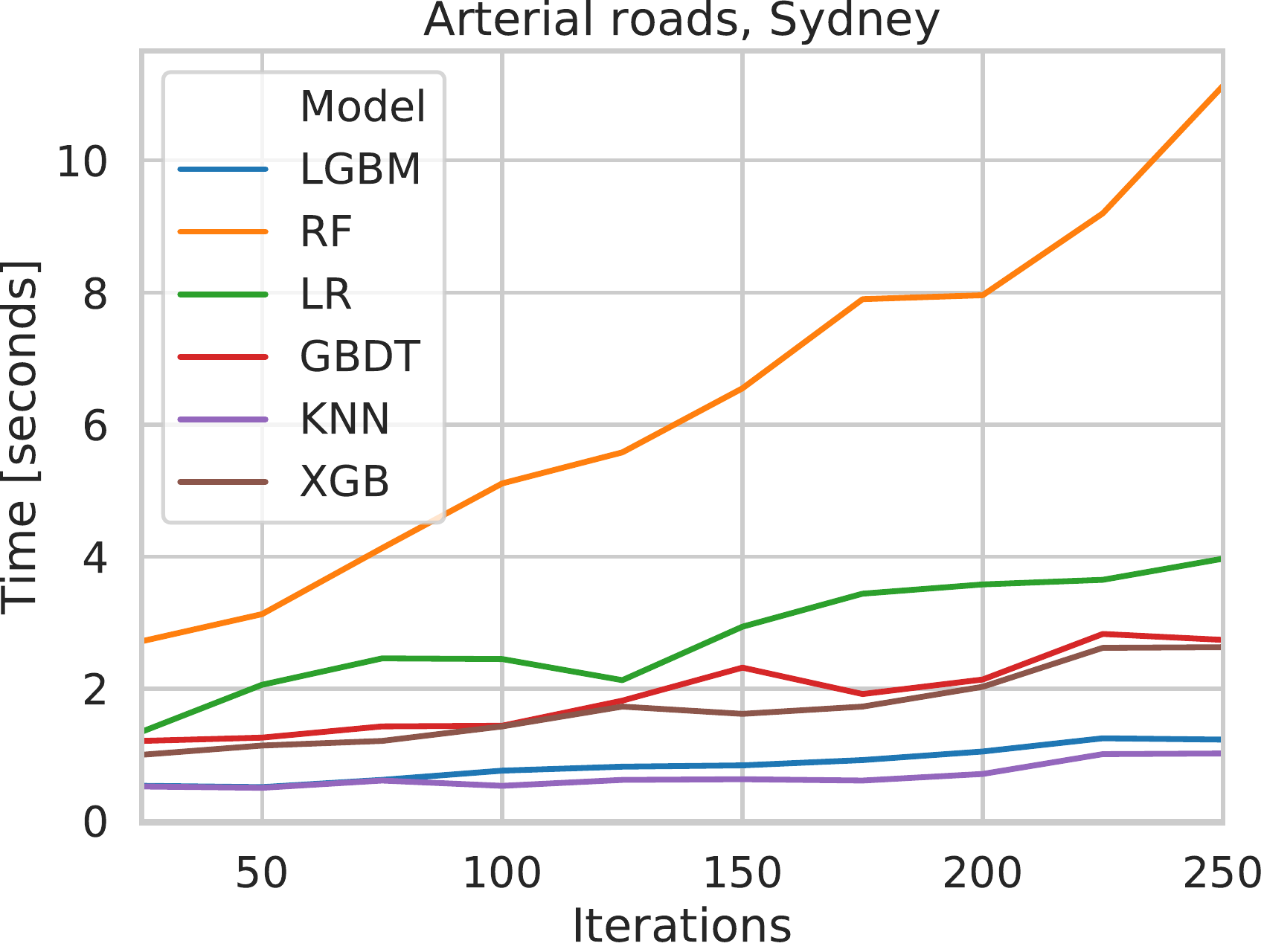}
\includegraphics[width=0.32\textwidth]{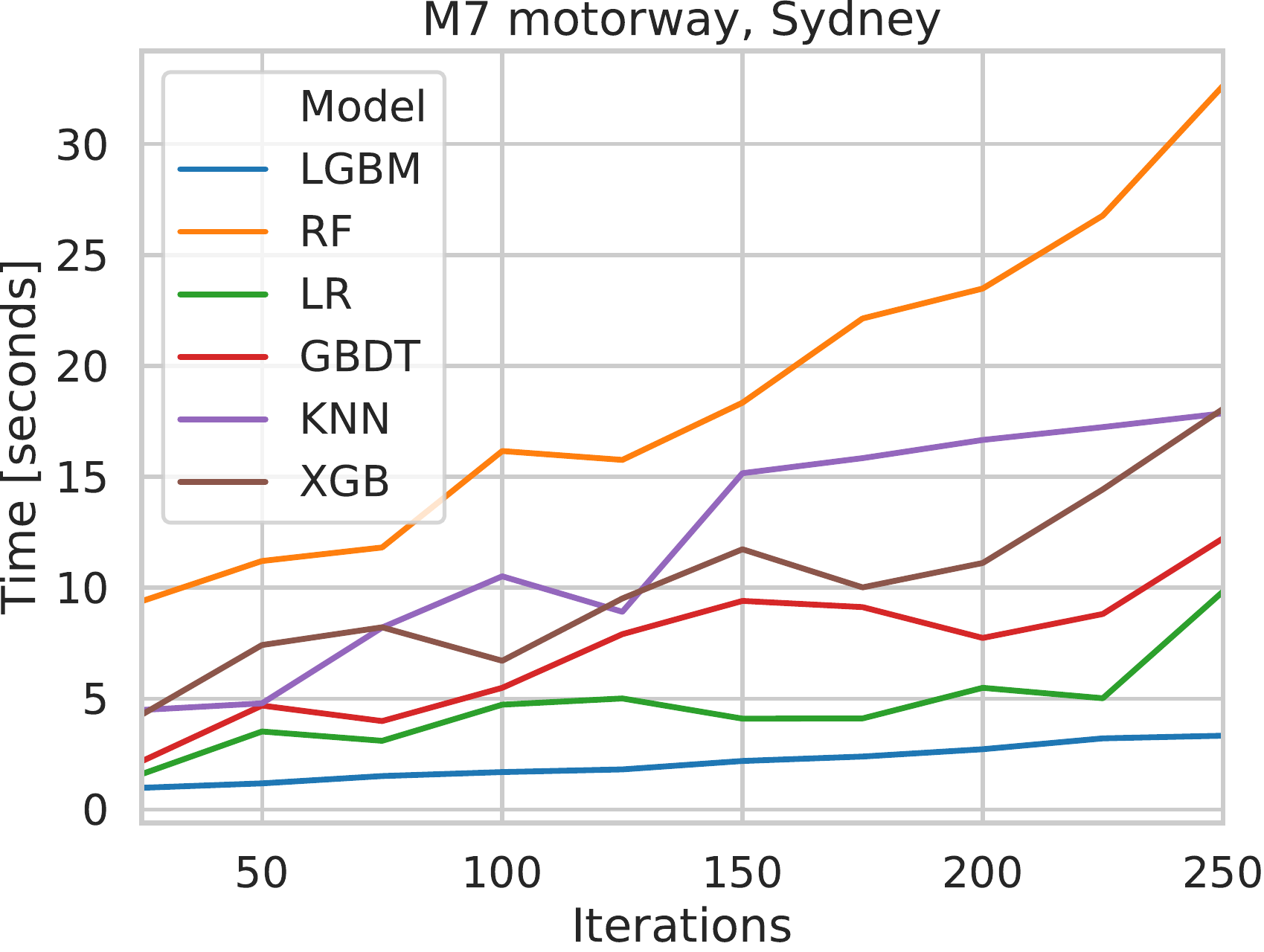}
\includegraphics[width=0.32\textwidth]{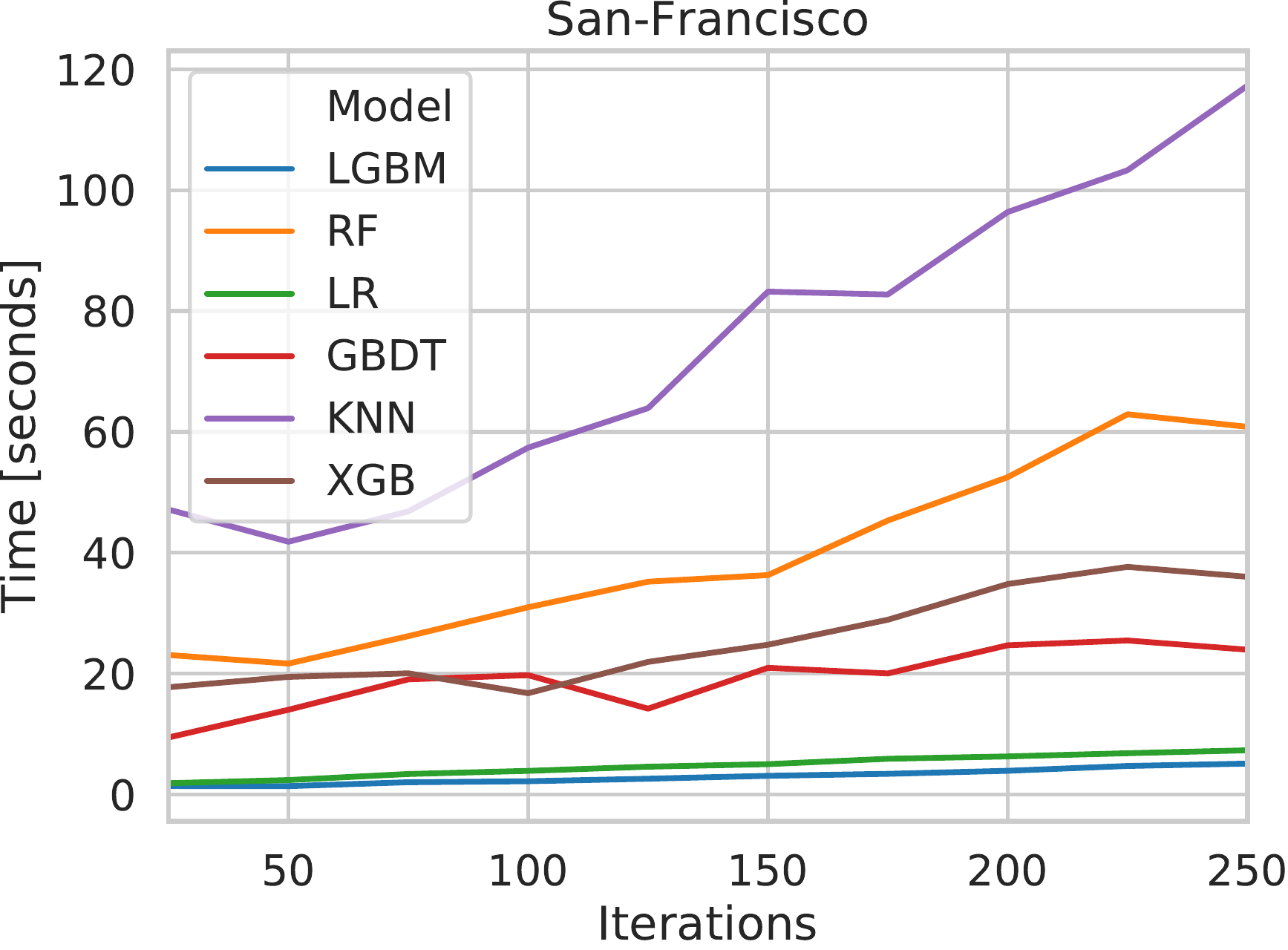}
\centering
\caption{Performance testing of ML models across three different data sets}
\label{fig:iters2}
\end{figure}
%
%
\printcredits

{ \small
\section*{ACKNOWLEDGMENT}
This work has been done as part of the ARC Linkage Project LP180100114. 
The authors are highly grateful for the support of Transport for NSW, Australia.
This research is funded by iMOVE CRC and supported by the Cooperative Research Centres program, an Australian Government initiative."
}

\bibliographystyle{cas-model2-names}

\bibliography{TRC2022_Incident_duration_pred}


\vspace{3mm}

\end{document}

%% file: sections/1-introduction.tex
%
\section{INTRODUCTION}\label{S1_Intro}

\subsection{Context}\label{S1_intro_context}
%


Traffic congestion is a significant concern for many cities around the world. Congestion arises due to various factors, including increased population, workforce concentration in central areas, or the lack of efficient public transport modes. Two forms of congestion are typically predominant: a) recurrent traffic congestion during peak hours when traffic demand exceeds the road capacity, and b) non-recurrent traffic congestion caused by unplanned events such as car accidents, breakdowns, weather, public manifestations etc. Previous studies have shown that almost $60\%$ of traffic congestion is due to non-recurrent incidents with a stochastic behaviour in space and time \cite{arc1}. In Australia, the number of road deaths per year has been reduced by $70\%$ since the 1970s. However, the annual economic cost of road crashes was estimated at \$27 billion per annum in 2017 \citep{arc2}.
Traffic Incident Management Systems (TIMS) collect data on traffic incidents, including information on different incident duration factors. Accurately predicting the total duration shortly after an incident took place could save operational costs and end-user time (through affecting the route planning). Moreover, the clearance time of accidents is highly related to the ongoing traffic congestion and several external factors with different weights of importance. Therefore, it is essential to estimate the incident factor importance to improve the accuracy of predictions. Most prior studies related to this topic concentrated on testing different machine learning models on specific road types like freeways or highways and focused primarily on different phases of the incident duration such as clearance time, recovery time, and the total incident duration \cite{LiOverview2018}. There is currently a lack of an advanced approach that can be applied on all road types, for all accident types and across various countries with different driving behaviour.

\subsection{Challenges and contribution}

The accuracy of predicting the incident duration is often determined more by the modelling methodology, the feature construction, and the result interpretation rather than by the model in use. In this work, we address several open questions or challenges concerning the prediction of the traffic incident duration.


\textbf{The first} challenge is to develop a universal bi-level framework applicable to different incident data sets reported on various road network layouts. The majority of prior works had studied the prediction of incident duration on specific types of roads (freeways or motorways) \citep{yubin}-\citep{chungyou}-\citep{hoj2}-\citep{zhan2}, where the data accuracy is higher than on arterial roads; as of 2018, very few applied the prediction strategies on normal arterial roads due to the high modelling complexity and a location mismatching; the majority of traffic incident duration analysis studies focus only on one type of road network (freeways, highways, etc.); this is revealed by a recent state-of-the-art paper published in \citep{LiOverview2018} which emphasises the difficulty of solving this problem for arterial roads and the lack of studies in this field. Our study proposes a framework capable of predicting the incident duration regardless of the road network or its complexity. 


\textbf{Secondly}, the majority of studies in the literature have concentrated on applying state of the art machine learning models mostly for classifying the incident severity \citep{Nguyen2017} or their duration\cite{LiOverview2018}. However, very few have treated the problem of outliers or imbalanced data classes. Our study addresses both of these issues by proposing a varying threshold procedure that can facilitate binary duration classification threshold selection by considering both class balance and model performance. We also test multi-class classification on data sets split into three equally-sized parts according to incident duration: short, medium or long term. Previous research studies were selecting incident duration thresholds by simple reasoning (e.g. choosing mean, median, percentiles, etc) \citep{kuang2019predicting}-\citep{Zou1}-\citep{li2015competing}-\citep{li2014}. We, on the contrary, test multiple different thresholds for three different data sets. Furthermore, we propose our own optimisation approach which we denote intra-extra joint optimisation (IEO) together with an outlier removal procedure (ORM) and advanced machine learning modelling.


\textbf{Thirdly}, we further solve the incident duration regression problem and also perform different regression scenarios to test the extrapolation performance of ML models on various incident data sets. 
We utilise thresholds selected during the classification threshold evaluation procedure to analyse the extrapolation performance by training ML models and making predictions on several duration subsets. It allows us to find the best ML model and the best extrapolation approach for the regression problem on each duration subset (e.g. short-term incidents) of each data set. For the regression problem, we also detect the most influential factors that affect the incident duration that traffic centres need to prioritise in order to predict incident duration with higher accuracy. Our end goal is to improve the extrapolation ability of machine learning models on the task of incident duration prediction and find the best modelling approach for short-term and long-term incidents.






\textbf{Lastly}, the majority of studies are primarily focusing on choosing a single winning algorithm or approach that works for a specific case study. Unfortunately, we show that the performance of ML models is highly affected by the data set and the chosen methodology: data quality, the available features, and the additional parameter tuning and optimisation techniques applied in this work. We try to develop the universal framework for traffic incident duration prediction applicable to different traffic incident data sets. We choose and adapt the best modelling approaches to each data set and show how this can affect the accuracy and performance of the models. This method allows high flexibility that can be applied for classification and regression predictions on various network types and different data sets. 



\revA{
The most similar research to the current work was published in \cite{kuang2019predicting} and relied only on one data set, one method for classification (Bayesian network), one method for regression (K-nearest neighbours), and authors selected static threshold (30 min) to alleviate the class-imbalance problem. This current paper provides a significant contribution by advancing on multiple aspects from a large pallet of machine learning models to multiple data sets with unique features, up to outlier removal and joint optimisation.}

\revA{
\revA{\textbf{Overall, our main paper contributions are the following}:
\begin{itemize}
    \item to the best of our knowledge, this is the first research study proposing a bi-level prediction framework using a large pallet of several machine learning models applied for both incident duration classification and regression, with the scope of predicting the incident duration on different road types across two different countries (Australia and U.S.A.). Overall, our methodology is agnostic of the location, the network, or the size of the network and can be adapted to any new incident log data set that can be made available. 
    \item we propose a binary versus multi-class classification approach in order to find the best optimal threshold to identify short versus long-term incidents via both quantile analysis and varying threshold data split.
    \item we propose a novel intra/extra joint optimisation algorithm that integrates baseline ML models with outlier removal and hyper-parameter optimisation techniques across the validation cycle.
    \item we propose several extrapolation scenarios of analysing the impact of missing logs in the precision of the prediction model and reflect on what type of logs should be best used for tailoring to the prediction problem needs. 
    \item we conduct a feature importance selection sing the SHAP method, which allows graphical interpretation of variables impact on the model output, before we conclude on the most important factors affecting traffic incidents.
\end{itemize}
 }

Overall, this research lays the foundation stone of bi-level predictive methodologies regarding the traffic incident duration and can provide accurate information for both the end-user route choice modelling as well as for the operational centres which need to optimise their operations under non-recurrent traffic congestion. Moreover, this work contributes to our ongoing objective to build a real-time platform for predicting traffic congestion and to evaluate the incident impact during peak hours (see our previous works published in \citep{Arterial2019}-\citep{shafiei2020short}-\citep{mao2021boosted}).


}

The paper is organised as follows: Section 1 discusses related works, Section 2 presents the data sources available for this study, Section 3 showcases the methodology, Section 4 presents the numerical results for binary and multi-class classification tasks,  Section 5 presents the numerical results of the regression part of the framework, Section 6 details on the feature importance evaluation and Section 7 is reserved for conclusions and future perspectives.


\subsection{Related works}
\paragraph{Incident data interpretation:}
The definition of traffic incident duration phases is provided in the Highway Capacity Manual \cite{Clearance2011}, and it consists of the following time-intervals: 
1) \textbf{incident detection time} which is the time interval between the incident occurrence and its reporting, 
2) \textbf{incident response time} standing for the time interval between the incident reporting and the arrival of the first investigator at the location of the accident, 
3) \textbf{incident clearance time} representing the time interval between the arrival of the first investigator and the clearance of the incident, 
4) \textbf{incident recovery time} indicating the time interval between the clearance of the incident and the return of traffic flows to normal conditions.

The \textbf{total incident duration} is the time interval between the first incident log, and the returning of traffic flows to normal conditions. In our work, we use the term \textbf{incident duration} for the time lapse between the detection of an incident and the clearance of the incident, as officially reported in traffic logs provided by local traffic authorities. Therefore we do not include the incident recovery time as this information is not recorded in the three data sets provided. 
However, different phases of traffic incident duration (e.g. clearance, recovery time) can be modelled individually upon availability; this type of research is rare because of the complexity of data collection for traffic incidents and small amounts of recorded traffic incidents in real-life datasets \cite{LiOverview2018,Clearance2011}.

When it comes to the data interpretation in the literature, the incident duration distribution has been modelled as log-normal \cite{sullivan1997new} and more recently as log-logistics distribution \cite{chung2010modeling,smith2001forecasting}. In a recent study, \cite{Haule}, incident clearance time and the total impact duration were modelled using Weibull, log-normal, log-logistic distributions and compared using the Akaike information criterion (AIC) criteria; findings have revealed that log-logistic distribution was outperforming other distributions. As distribution utilisation is highly related to the specificity of each data set, for this study, in which we use three different data sets, we further apply a comparison among several distribution modelling choices by using the AIC criteria.

\revA{

According to \cite{waliheterogeneity}, different statistical methods were applied to model traffic incidents: 1) fixed parameter regression  2) random parameter regression 3) quantile regression. In this study, log-transformation of the target variable (incident duration) also applied.
Random parameter regression found to give better statistical fit for incident duration models than fixed parameter regression, and therefore provide more accurate predictions of incident durations. It also highlighted that fixed-parameter regression model may give non-accurate incident duration predictions due to over/under- estimation of dependency between variables and incident durations. Also, there were no substantial difference found between fixed parameter regression and quantile regression in the case of 2015 Virginia incident data set.
The benefit of quantiled regression is the ability to model
the relationship of any quantile (rather than only average incident duration) of the incident duration vector with a set of explanatory variables \cite{khattak2016modeling}. Ordinary Least Squares model can provide the predicted mean of the incident duration. On the contrary, quantile regression provides estimates for every quantile, which represented as a conditional distribution of incident durations, without providing single value as the incident duration prediction. Quantile regression coefficients represent the change in the incident duration in a given quantile category in relation to independent variables. Similar to this approach, variable importance can be estimated within each traffic incident duration group.

}


\paragraph{Machine Learning for incident duration prediction:}

While several statistical modelling techniques have been applied previously, more recently, new approaches in machine learning (ML) modelling have emerged as a more advanced way of predicting the incident duration due to their capacity to easily account for new data sources, as well as for removing the linearity assumptions between features and the predicted class \cite{Hojati2014}. Examples of such approaches are: artificial neural networks (ANNs) \cite{lopes2013dynamic}, genetic algorithms \cite{lee2010computerized}, support vector machines (SVMs) \cite{valenti2010comparative}, k-Nearest-Neighbours (kNNs) \cite{wen2013traffic} and decision-trees (DTs) \cite{he2013incident}. The recently proposed Gradient-Boosted Decision Trees (GBDTs) have been shown to provide superior prediction performance when compared to Random Forests, SVMs and ANNs \cite{ma2017prioritizing}. However, it is known that GBDT can easily over-fit when the prediction target has a long-tail distribution, as is the case of the traffic incident duration distribution \cite{ma2017prioritizing}. XGBoost \cite{chen2016xgboost} is another decision-tree enhancement method that has gained popularity recently in the machine learning community due to its tree boosting capability, loss function regularisation and adaptive learning rate. It was employed in several international competitions, winning 17 out of the 29 Kaggle competitions singled out on the 2015 Kaggle blog; it was also employed by every team in the top-10 in the 2015 KDDCup \cite{bekkerman2015present} for solving various problems such as store sales prediction, web text classification, hazard risk prediction, and product categorisation. XGBoost's popularity is also due to its scalability (it can run on a single machine, as well as on distributed and paralleled clusters), its capacity to handle sparse data and its ability to handle instance weights in approximate tree learning (see the recent paper published by \cite{chen2016xgboost} where authors proposed an end-to-end tree boosting system with cache-aware and sparsity learning features). While each of these methods has its advantages and disadvantages, building a fast and reliable prediction framework that could be applied for real-time operations represents a true challenge.

One of the recent research studies \cite{kuang2019predicting} presented a two-step approach for traffic incident duration prediction. A cost-sensitive Bayesian network was used to perform binary classification of traffic incidents by choosing a threshold of 30 minutes and then performing regression for each class using kNN. While the approach is functional, one major drawback for the classification problem is to manually choose the class split threshold, as it can lead to severe class imbalance; to overcome this issue, in our study, we perform both a fixed and a varying threshold set-up to find the best class balance for our classification models; even-more, we propose as well a comparison with a multi-class classification approach and debate on the benefits and drawbacks of using classifiers for such problems; we also enhanced more advanced regression models together with outlier removal procedures that would provide a better and more precise prediction of the incident duration precondition in minutes. Overall, the cost sensitivity of incorrect classification can be further extended to the cost-based regression metrics. We propose our enhanced ML models with a proposed intra and extra joint optimisation technique and outlier removal procedure to have even more precise predictions.

In one of the recent research studies on applying machine learning, which was related to the classification of driving state, multiple hyper-optimised ML models were tested, and entire feature space was visualised using t-SNE for entire feature space visualisation \citep{yi2019machine}. RandomForest provided the highest prediction accuracy, but more advanced tree-based models exist that utilise gradient boosting, which we will be using in our research (e.g. gradient boosted decision trees). 

To verify the performance of advanced tree-based methods \revA{(as LGBM - Light Gradient Boosted Model)}, additional conventional ML models can be used \citep{chen2020predicting}. We decided to also include LGBM and compare it to conventional ML models with non-tree based models (KNN and Logistic Regression).

\paragraph{On the feature selection:}
It is generally not enough to use all the possible features for the regression analysis of traffic incident durations. Using a high amount of features combined with a small data set size can lead to over-fitting. Some features can be helpful or useless, more or less critical, while others do not impact the prediction results significantly. 
\revA{
By performing a feature importance analysis, we can recommend to traffic management facilities to record the most critical data and omit redundant data related to traffic incidents. 
For example, one can increase the prediction accuracy by using as additional features the weather conditions, which were found to play a significant role in some research studies (e.g. during the summer and autumn seasons in Washington – USA in 2009, the preparation time of the rescue team was higher on freeways \cite{Mecha2013}). In some countries with cold weather, the response times can be much higher, while in regions with sunny weather most of the year, the weather impact on the intervention team can be neglected. Overall, the weather impact on the traffic incident duration prediction needs needs customised via a data-driven feature important analysis. 
} 
Peak hours were the most influencing feature on response team preparation delay, which was found to be linked to response procedures (the goal of the response team was to resolve incidents during peak hours as soon as possible). A research study using Beijing traffic incidents data from 2008 \cite{HBM2014} found the importance of "peak hour" value for the response team travel time and clearance time, but not for the intervention team preparation time. Our study conducts a feature importance ranking based on the best performing ML models we have proposed and provides a detailed overview of their impact. Different approaches to feature importance estimation use tree-based models (e.g. Random Forest, Light Gradient Boosted Machines - LGBM, extreme gradient bosoting models - XGBoost). For example, one can use produced decision trees from the tree-ensemble model \cite{chen2020predicting}. A data-driven approach was used to perform information fusion from different sources \cite{abou2020real}, which involved the use of Gini-index extracted from Random Forests as a method to estimate feature importance. Nevertheless, the single random model can have a noticeable variance in data mapping when there is a weak connection between features and the target variable by making the feature importance value dependent on the random seed for the model. The Shapley Additive explanation (SHAP) \cite{lundberg2017unified} provides a more advanced approach for feature importance estimation because it fuses estimation from multiple models trained across many different subsets (which selected both feature-scale and index-scale) of the dataset. These studies motivated the utilisation of the Shap Values for our feature importance ranking across three different data sets, all with different features and incident information.

\paragraph{On the future application of our research: }

In comparison with other work, the research proposed in our paper comes not only with a significant prediction capability for all types of incident data sets with various features, but it can be further extended for solving the route scheduling problem within traffic simulation modelling, which will incorporate the adaptation of agents to occurring traffic incidents. 
Apart from analysing the effects of traffic control measures \cite{knapen2014within}, it is possible to analyse the effect of additional information such as the predicted incident duration, which can be performed both for scheduling and online rescheduling of dynamic agent re-routing. Furthermore, simulation can be performed with and without such information to estimate the possible benefits of the incident duration prediction modelling within the traffic system. Also, using an online rescheduling procedure requires the simulation to be performed at the level of dynamic agents within a micro-simulation model, which could benefit from new re-routing schemes when traffic disruptions occur along the route.


%% file: sections/2-Data-sources.tex
\section{DATA SOURCES}\label{2-Data-sources}
\begin{figure}[h]
\centering
\includegraphics[width=0.95\textwidth]{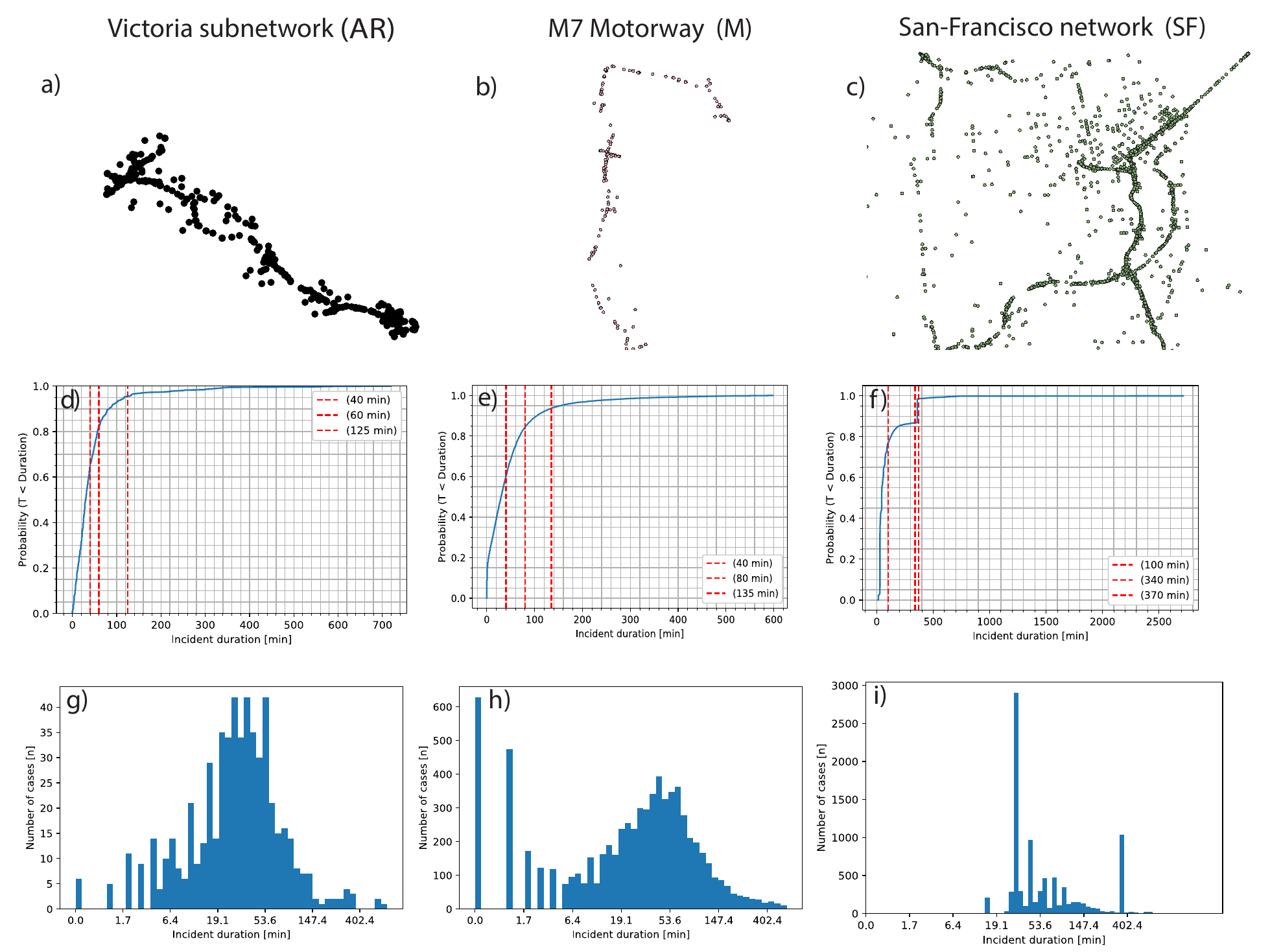}
\caption{Data profiling for all data sets in our study: Victoria Rd (A) - a) network mapping, d) ecdf - empirical cumulative distribution function g) distribution plot; M7 motorway (M) - b) network mapping, e) ecdf h) distribution plot; San Francisco (SF) - c) network mapping, f) ecdf i) distribution plot.}
\label{fig:dataprofiling}
\end{figure}

In order to test the efficiency of the proposed bi-level framework, we have used three different data sets from two different countries: Australia and U.S.A. The three data sets represent incident logs from an arterial road suburb in Sydney, a motorway in Sydney, Australia, and a road area from San Francisco, U.S.A. The data sets are all recorded by different means and allow us to explore the impact of the prediction framework across various types of road networks. The three data sets are represented in \cref{fig:dataprofiling} and are detailed as follows.

\paragraph{\textbf{Victoria Rd - arterial network, Sydney:}}

The first data set (dataset AR) contains one-year incident logs from the Victoria arterial road from Sydney, Australia (in 2017) (see \cref{tab:my_label} for a summary of features, \revA{in which the $+$ symbol under each data set column and for each line indicates whether that variable is present or not in the data set - for example, the TZName variable is present in the Arterial Roads data set but not in the M motorway data set}). It contains information on 5,134 traffic incidents with different incident types (e.g. hazards, breakdowns, accidents) and subtypes (e.g. work zone, accident with truck). Our current study focuses on 574 ``Accidents'' since these induce the longest clearance time in the current subnetwork according to the traffic management centre (TMC). Traffic 'Accidents' have a mean duration of 44.59 minutes and a maximum of 719 minutes. 

Weather data represented as average daily temperature (in Celsius) and precipitation rate (in millimetres) are obtained from the Observatory Hill station in Northern Sydney, which is the closest station to the analysis area. Public holiday data represented as boolean values for public and regional holidays in 2017 in New South Wales, Australia.
The area geometry features contain the sector ID as defined by TMC, the code of the official area where the accident occurred (as defined by the Bureau of Transport and Statistics), and supplementary information such as section capacity, section speed limit, and the number of lanes. These features are available for all road sections in the Victoria sub-network, and they were extracted from the official traffic simulation model of the Victoria network, developed in Aimsun and previously used by the authors for conducting an incident impact analysis and traffic prediction \citep{wen2018integrated}.

\begin{table}[ht]
    \centering
    \begin{adjustbox}{max width=\textwidth}

    \begin{tabular}{l|l|l|l|l}
        \textbf{Variable} & AR & M & \textbf{Values} & \textbf{Description} \\
        \hline
        Location & + & + & $\mathbb{N}, \mathbb{N}$ & ${X,Y}$ in GDA Lambert coordinates\\
        Hour of day & + & + & $\{0,1, \ldots ,23\}$ & - \\ 
        Peak Hour & + & + & $\{1,0\}$ & Value is 1 if hour belongs to $\{7 \ldots 9\}$ or $\{16 \ldots 18\}$ hour interval \\
        Day of the week & + & + & $\{1 \ldots 5\}$ & Weekday numbers from Monday to Friday \\
        Weekend & + & + & $\{0,1\}$ & Value is 1 for Saturday and 0 for Sunday \\
        Month of the Year & + & + & $\{1,2, \ldots ,12\}$ & - \\
        Incident Subtype & + & + & $\{Bus, car, bicycle, animals, etc.\}$ & Field indicating cause of incident \\
        Affected lanes & + & + & $\{1, 2, 3, 4, All lanes, breakdown, no data\}$ & Number of lanes affected by the accident\\
        Direction & + & + & E, W, N, S, E-W, N-S, One/Both  & Affected traffic direction \\
        Incident Source & + & + & $\{ICEMS/ISENTRY, OPERATOR, etc\}$ & Source of the incident report \\
        Unplanned & + &   & $\{0,1\}$ & Value is 1 if incident is planned, 0 otherwise \\
        Average Temperature & + & + & $\{11.13 C - 32.4 C \}$ & Average temperature for the time of the incident \\
        Rainfall & + & + & $\{0 - 85mm\}$ & Rainfall for the time of the incident\\
        Public holidays & + & + & $\{0,1\}$ & Value is 1 if days is a public holiday\\
        Sector ID & + & + & R+ & Defined by TMC\\
        TZName & + &   & R+ & Traffic zone name as Defined by the Bureau of Transport Statistics\\
        Section ID & + &   & R+ & Road section on which the incident occurred\\
        Section Speed & + &   & $R+ [Km/h]$ & Section speed limit\\
        Section Lanes & + &   & $\{1,2,3,4,5,6\}$ & Number of section lanes\\
        Section class & + &   & $R+$ & As defined by TMC\\
        Street ID & + &   & $R+$ & As defined by TMC\\
        Intersection ID & + &   & $R+$ & As defined by TMC\\
        Distance from CBD & + &   & $R+$ & distance between the traffic incident and the city CBD\\
        Section Capacity & + &   & $\{0 \ldots 3100\ vehicles/hour\}$ & Maximum flow capacity of the section
    \end{tabular}
    \end{adjustbox}    
    \caption{Traffic incident features for Sydney Arterial roads (AR) and M7 motorway (M).}
    \label{tab:my_label}
\end{table}

\paragraph{\textbf{M7 motorway, Sydney:}}

The second data set is a motorway data set (data set M), consisting of 7,194 traffic accidents along the M7 motorway in Sydney, Australia, during the same year 2017. The mean duration of motorway accidents is 47.2 minutes, with a maximum duration of \revA{598 minutes (9.96 hours).}
This data set also includes weather data (average daily temperature and precipitation). This set of features is similar to the arterial roads data set AR without the geometric features of the lanes (section lanes, section class), intersection ID, distance from the central business district (CBD); this is due to the complexity of mapping of a traffic incident to a correct location along the motorway. We make the observation that for both Data set AR and M, the traffic flow information of the affected road sections was omitted for this study since we found previously no significant improvement to the prediction accuracy \citep{Arterial2019}.

\paragraph{\textbf{San-Francisco road network:}}

The last data set is from San-Francisco, U.S.A. (data set SF) and includes information on accidents from all types of roads in the city. It is part of a more considerable initiative entitled "A Countrywide Traffic Accident Dataset", recently released in 2021, which contains \revA{1.5} million accident reports collected for almost 4.5 years since March 2016 \citep{moosavi2019countrywide}. The SF data set contains 49 features describing the accidents as detailed in \citep{moosavi2019countrywide} (due to a large table of feature, we refer the reader to the cited paper and not duplicate this feature information). This study focuses on the "accident'' type duration prediction as being the most severe one. We extract and use 8,754 accident records related to the San-Francisco area. As observed from \cref{fig:dataprofiling} c), a significant part of the accidents occurred along the ``US-101'' highway and ``John F. Foran'' Freeway. Accidents have a mean duration of 100 minutes and a max duration of 2,715 minutes.

\paragraph{\textbf{Data sets profiling:}}
Each data set undergoes a profiling procedure by investigating the empirical cumulative distribution functions (ECDF) - as plotted in \cref{fig:dataprofiling} d), e), f), and their equivalent log-space distribution plots (as represented in \cref{fig:dataprofiling} g), h), i). The ECDF function presents thresholds of data behaviour (marked in red) across each data set which reveal indicative thresholds of a different behaviour around specific incident duration (see for example \cref{fig:dataprofiling}d) versus \cref{fig:dataprofiling}f) where the first inflection point is around 40min for data set AR versus 100min for data set SF. Findings reveal significant anomalies representative of each data set. For example, data set AR contains a reduced amount of traffic accidents with small incident duration (zero or less than 4 min), data set M contains an increased number of accidents with zero or one-minute duration, while the data set SF despite not presenting any short term incident duration below 17 minutes, it contains a large number of incidents of 29 and 360 minutes which raises the question of either these are outliers in the data set or simply reveal a road network behaviour in terms of incident management in the area; this also might indicate that it will present unique behaviour under the prediction framework and that different processing techniques needs to be applied for this data set. We also observe that the incident duration is long-tail distributed, which is likely to pose difficulties for prediction algorithms due to the presence of extreme values (either small or large).

%% file: sections/3A-Methodology.tex
\section{METHODOLOGY}\label{S3-Methodology}

\begin{figure}[h]
\centering
\includegraphics[width=0.8\textwidth]{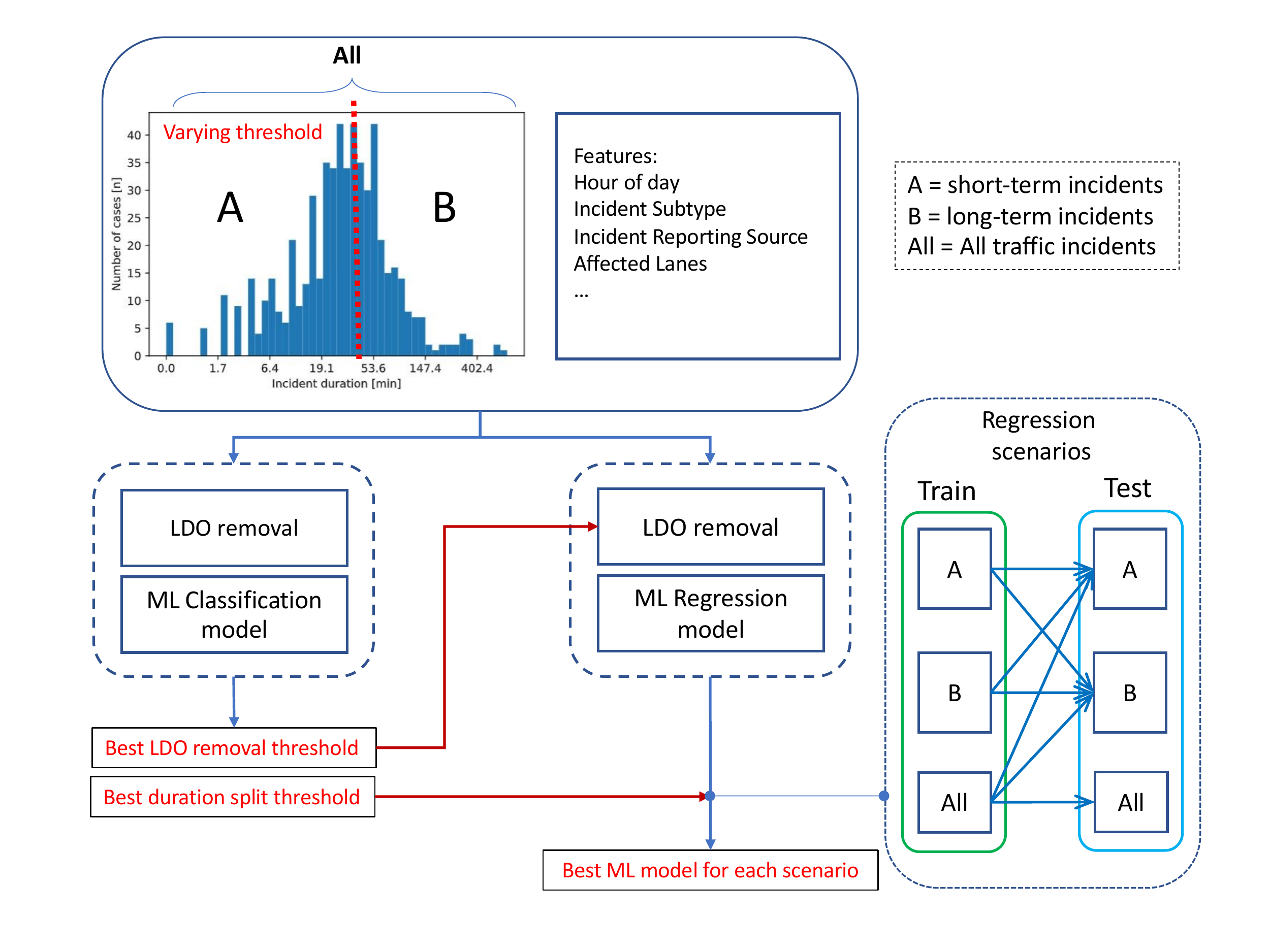}
\centering
\caption{The proposed bi-level modelling framework for traffic incident duration prediction.}
\label{fig:scheme}
\end{figure}

Clearing accidents in a short time represents a high priority task for traffic management centres (TMC) worldwide. For example, in New South Wales, Australia, the target clearance time for traffic incidents is 45 minutes, but this limit might differ in other countries. Therefore, in the rest of this paper, we will refer to this threshold as ``incident clearance threshold ($T_c$)'' and any incidents cleared before this threshold (e.g. $< 45$ min) as "short-term";  incidents which lasted more than the clearance threshold (e.g. $>= 45$ min) will be referred to as ``long-term'' traffic incidents. A unique threshold will be derived for each dataset and will be discussed further in this paper. The methodology of this paper has its origins in our previous work applied only for arterial roads \citep{Arterial2019}, which we further extend and improve via the joint optimisation and outlier detection enhancements of the prediction framework. The methodology we propose for modelling the incident duration prediction problem is using a bi-level prediction framework combining a classification and regression modelling, as represented in \cref{fig:scheme}. This approach has been constructed by considering the real-time operational goals of TMC and providing short duration prediction into the life-cycle of the incident management. 

Based on the initial traffic incident information, the first step is the deployment of a fast classification method which would only predict whether the accident will be either short-term (subset A) or long-term (subset B) - see incoming data set from \cref{fig:scheme} where the data is split in two parts based on $T_c$). Next, we test various duration thresholds and select the optimal $T_c^o$, which provides a good class balance and classification performance for each dataset.

Once the optimal $T_c^o$ has been found, a further regression modelling is applied for predicting a more precise duration of future incidents down to the minute level. 

Due to the main challenge of this task, we further propose an outlier removal approach (ORM) detailed in \cref{ORM} and our innovative Intra/Extra Joint Optimisation modelling coupled with several machine learning models trained via a hyper parameter tuning (we denote this approach as IEO-ML and is further detailed in \cref{IEO}). 

The boosted regression framework is finally applied under several regression scenarios (see section \cref{scenarios}), which are constructed to evaluate the framework capability to predict under all possible situations. For example, when we only have a subset A available (short-term incidents) but the TMC would like to predict long term incident (subset B) we denote this as a Scenario A-to-B (training the models on subset A and making predictions on subset B); all scenarios are constructed based on the assumptions that the framework needs to be robust in order to predict any type of incident durations, under all possible data shortage or lack of information availability. In the following subsection, we further provide the mathematical and theoretical modelling of each of the steps described above.

\subsection{Classification and regression definitions}

Using all available data sets and the incident information, we first denote the matrix of traffic incident features as:
\begin{equation}
    X = [x_{ij}]_{i=1..N_i}^{j=1..N_f}
\end{equation}
where $N_i$ is the total number of traffic incident records used in our modelling and $N_f$ is the total number of features characterising the incident (severity, number of lanes, type, neighbourhood, etc.) according to each specific data set (see examples provided in \cref{tab:my_label}). For the incident duration classification problem, we denote the incident duration classification vector as:
\begin{equation}
\left\{
\begin{split}
   & Y_c = [y_i^c]_{i \in 1..N} 			& & y_i^c \in \{0,1\}\\
    & Y_{mc} = [y_i^{mc}]_{i \in 1..N} 		& & y_i^{mc} \in \{0,1,2\}\\
\end{split}\right.
\end{equation}

where N is the duration of the traffic incident (in minutes), $Y_c$ is the vector of binary values taking values in $\{0,1\}$, and $Y_{mc}$ is the vector of integer values for the multi-class classification problem definition, taking values in $\{0,1,2\}$. More specifically, in the first stage we create a binary classification modelling with the purpose of identifying short versus long-term incident duration, split by the incident clearance threshold $T_c$. Thus our task is to predict $y_i^c$, where $Y_c$ takes one of the binary values:
\begin{equation}
\left\{\begin{split}
	y_i^c = 0 & & \text{if } y_i\leq T_c, 	& & \text{short-term incidents}\\    
    y_i^c = 1 & & \text{if } y_i > T_c, 	& & \text{long-term incidents} 
\end{split}\right.
\end{equation}
where the threshold is varied every 5min between $T_c \in \{20, 25, ... ,70\}$. 
Subsequently, the multi-class method identifies the best two thresholds to separate between short, mid and long-term incident duration. The main purpose of this approach is to test the limits of the class balance which would maintain good model performance, and is expressed as follows: 
\begin{equation}
\left\{\begin{split}
     y_i^{mc}= 0 & & \text{if }  y_i \leq T_c^1, 		& & \text{short-term incidents}\\
     y_i^{mc} = 1 & & \text{if }  y_i \in [T_c^1,T_c^2], & & \text{mid-term incidents}\\
     y_i^{mc} = 2 & & \text{if }  y_i \geq T_c^2, 		& & \text{long-term incidents}
\end{split}\right.
\end{equation}
where $T_c^1$ and $T_c^2$ take several values as further detailed in \cref{multi_class_classif}. The binary classification approach implemented with a computation time constraint for operational purposes (more details on computation time comparison can be found in \cref{appendix_B}).

The regression problem is further structured with a more fine-grained incident duration prediction in mind. The main objective motivating the regression modelling consists in more precise information regarding the duration of incidents which can fall into a wide class \revA{which contains mostly incident logs with a reported duration between and 0 minutes and 30 minutes} (for these cases, the traffic centres require more detailed precision to the minute level as a 5-min accident has different handling procedures than more severe accidents of 30min for example). The incident duration regression vector ($Y_r$) is represented as:
\begin{equation}
    Y_r = [y_i^r]_{i \in 1..N}, \text{ } y_i^r \in \mathbb{N}
\end{equation}
and the regression task is to predict the traffic incident duration $y_i^r$ based on the traffic incident features $x_{i,j}$. The regression models go via an extensive cross-validation procedure with hyper-parameter tuning, with the test of outlier removal using a joint optimisation approach as further detailed in the \cref{S3_tunning}-\cref{ORM}-\cref{IEO}.

\revA{
\subsection{Applicability of knowledge-based incident duration classification guidelines}

According to the The Manual on Uniform Traffic Control Devices (MUTCD) official guidelines \cite{mutcd} Section 6I, traffic incidents divided into three classes: a) Major - with expected duration more than 2 hours b) Intermediate — expected duration of 30 minutes to 2 hours c) Minor — expected duration under 30 minutes.

First, the MUTCD classification seems to be general knowledge-based system and does not consider specifics of each data set / country regulations / specifics of applied incident response guidelines. We approach the classification task from data analysis point of view in relation to application of ML models and try to infer these thresholds from the actual data sets. Shifting to MUTCD classification approach will also make incident duration classes imbalanced (see Figure \ref{fig:distmu}). Second, this classification may not be applicable due to road networks heterogeneity \cite{li2018analysis} and consequent differences in incident duration distribution. As can be seen from Figure \ref{fig:dataprofiling}, all three data sets have different distribution of incident durations and therefore such classification may be biased in each case.
Overall, in our study, we aim to provide insights from data analysis point of view.

\begin{figure}[pos=h,align=\centering]
\centering
\includegraphics[width=0.25\textwidth]{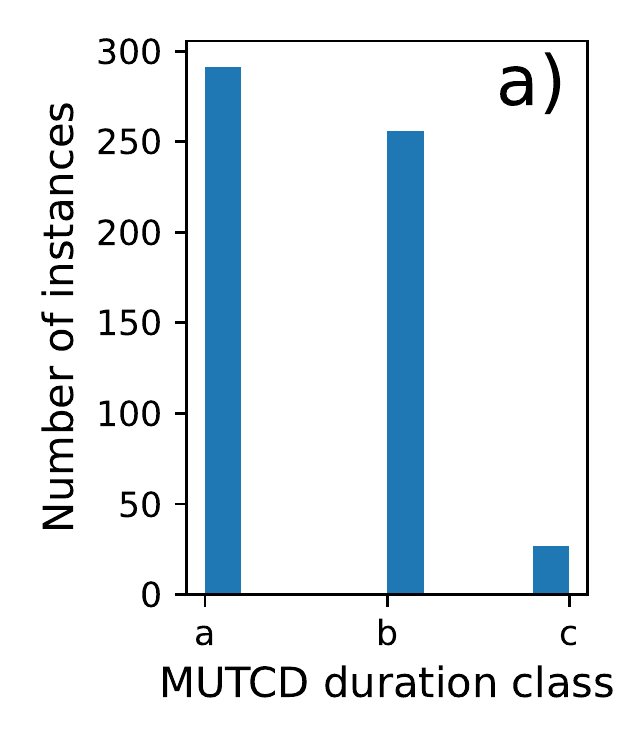}
\includegraphics[width=0.25\textwidth]{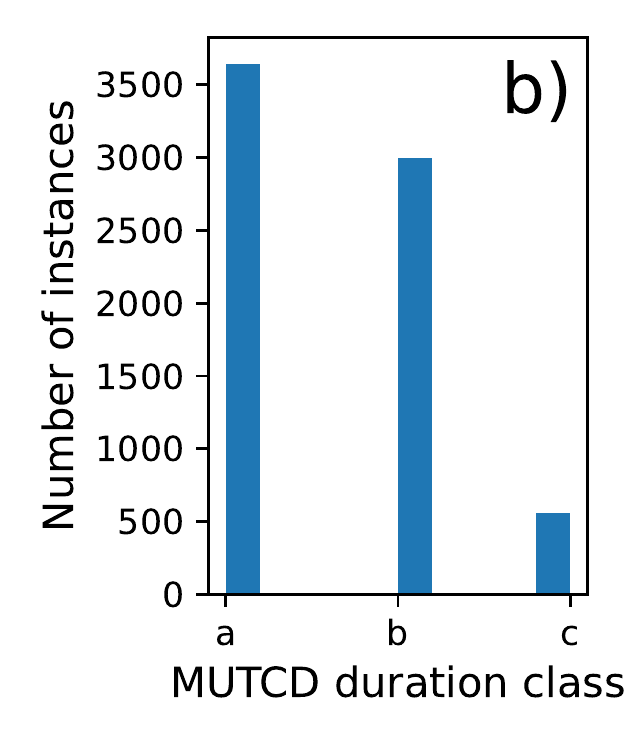}
\includegraphics[width=0.25\textwidth]{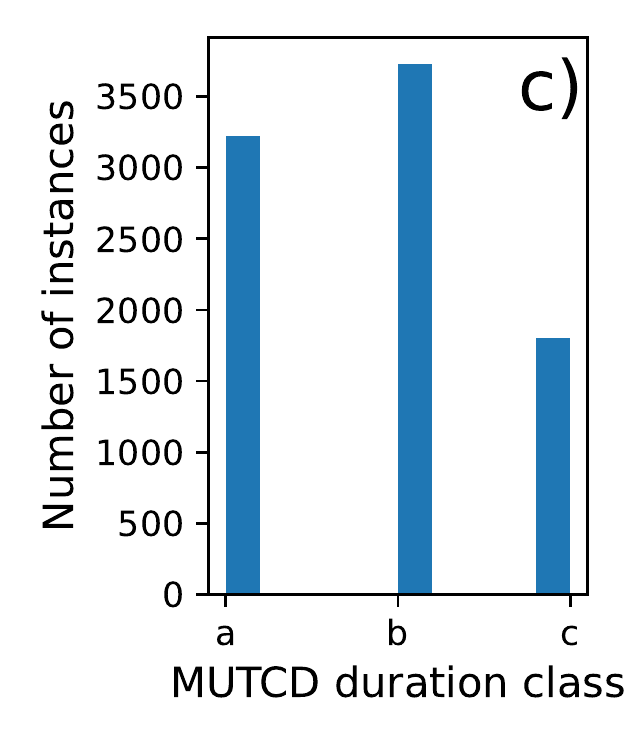}

\centering
\caption{Distribution of incident durations according to MUTCD duration classes: a) Arterial roads, Sydney, Australia b) M7 Motorway, Sydney, Australia c) San-Francisco, USA}
\label{fig:distmu}
\end{figure}
}

\subsection{Selection of baseline machine learning models}

We have tested and deployed several ML models for both the classification and regression problems for this current work, which have served as baseline models to compare our proposed optimisation approach. These are listed as follows: 
a) gradient boosting decision trees  - GBDT \citep{gbdt}  which rely on training a sequence of models, where each model is added consequently to reduce the residuals of prior models; 
b) extreme gradient decision trees - XGBoost \citep{xgboost} which finds the split values by enumerating over all the possible splits on all the features (exhaustive search) and contains a regularisation parameter in the objective function; 
c) random forests - RF \citep{randomforest} which applies a bootstrap aggregation (bagging, which consists of training models on randomly selected subsets of data) and uses the average (or majority of votes) of multiple decision trees in order to reduce the sensitivity of a single tree model to noise in the data; 
d) k-nearest neighbours - kNN \citep{fix1951discriminatory} which uses for the prediction on data points the majority of votes or the average from k closest neighbouring data points from the training set (based on a distance metric); 
e)  linear Regressions - LR - a standard predictor using linear equations to model the relation between the features and the regression variable;
f)  light gradient boosted machines - LGBM \citep{lightgbm} which applies gradient boosting to tree-based models; it also uses a Gradient-based One-Side Sampling (GOSS) and excludes data points with small residuals for finding split value.   
The models have been used for both classification and regression problems (except logistic regression applied to classification only and linear regression to regression problem only). They are the main base on which we further enhance and develop our outlier and joint optimisation prediction algorithm used in the current bi-level incident duration prediction framework.

\subsection{Hyper-parameter tuning through randomised search}\label{S3_tunning}

Most machine learning algorithms have a set of hyper-parameters related to the internal design of the algorithm that cannot be fitted from the training data. Both GBDT and XGBoost present dozens of hyper-parameters, out of which the most important ones are max\_ depth, learning\_rate, min\_ child\_weight, gamma, subsample, colsample\_ bytree and scale\_ pos\_ weight [24]. 
The hyper-parameters are usually tuned through randomised search and cross-validation. The most extensive search technique is the grid-search, in which several equally spaced points are chosen in the most credible interval for each parameter, and for each point combination, a model is fitted and tested through cross-validation. The grid-search parameter tuning is straightforward; however, the grid-search scales poorly as the number of hyper-parameters increases. In this work, we employ a Randomised-Search \citep{bergstra} which selects a (small) number of hyper-parameter configurations randomly to use through cross-validation.

\begin{figure}[h]
\centering
\includegraphics[width=0.32\textwidth]{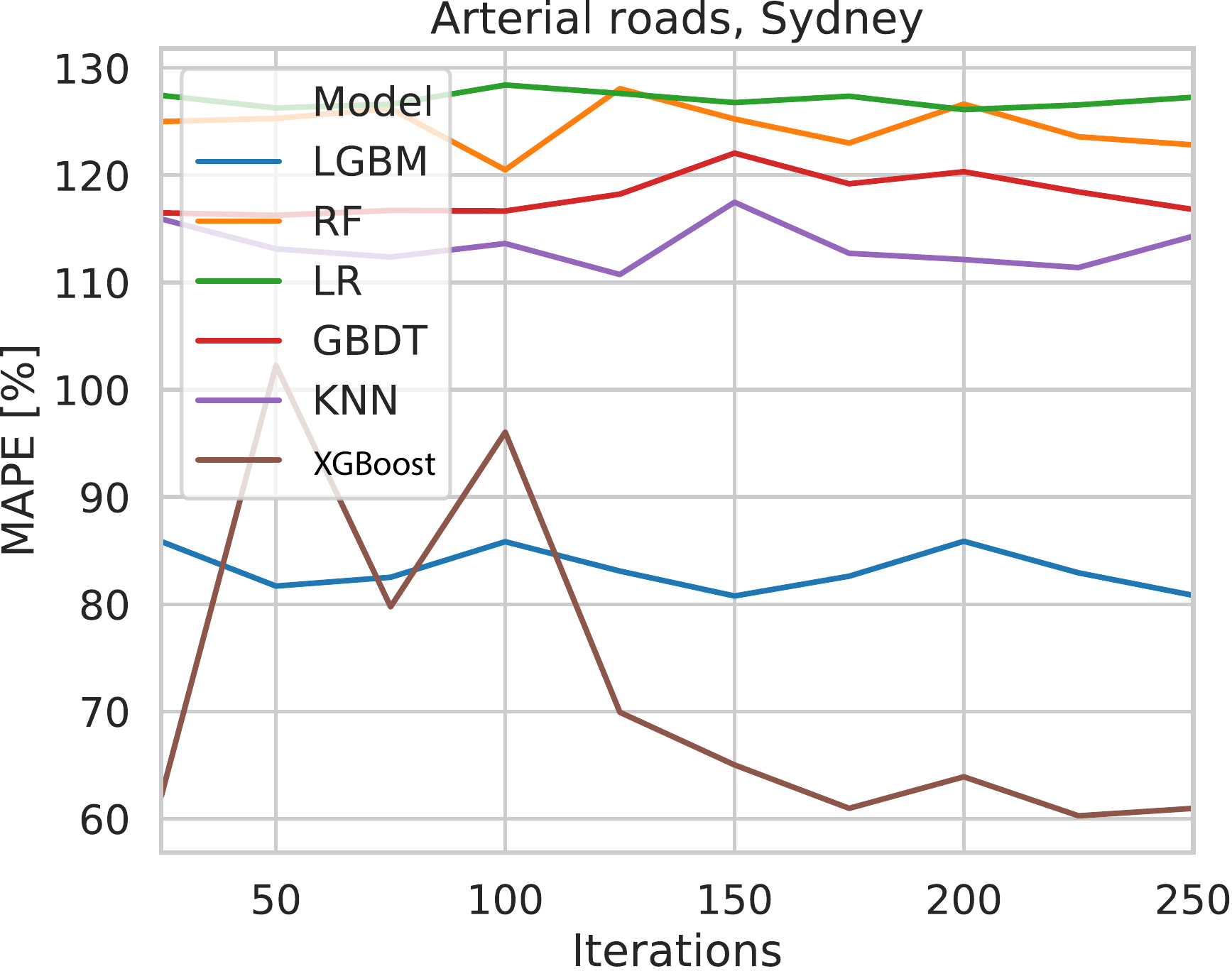}
\includegraphics[width=0.32\textwidth]{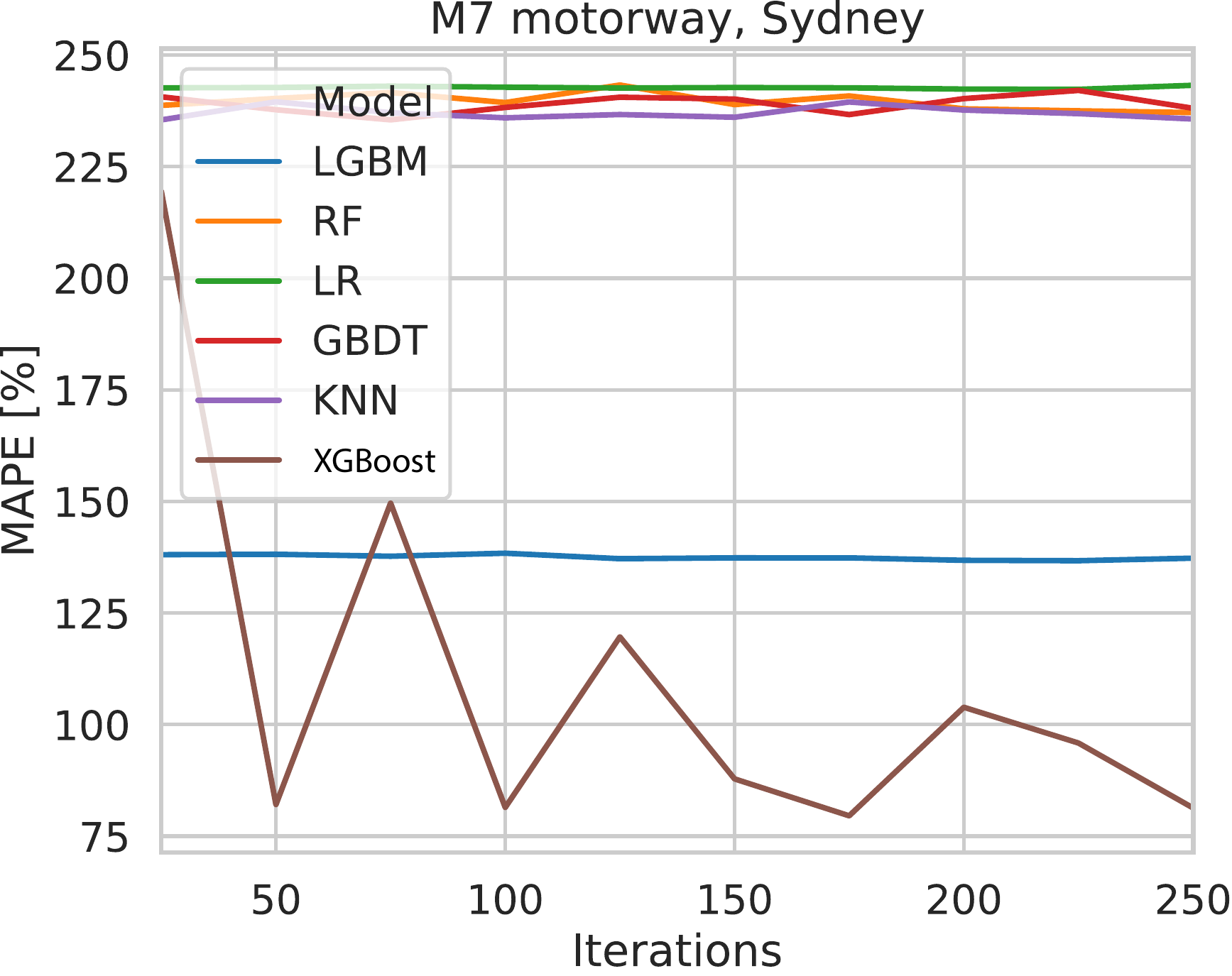}
\includegraphics[width=0.32\textwidth]{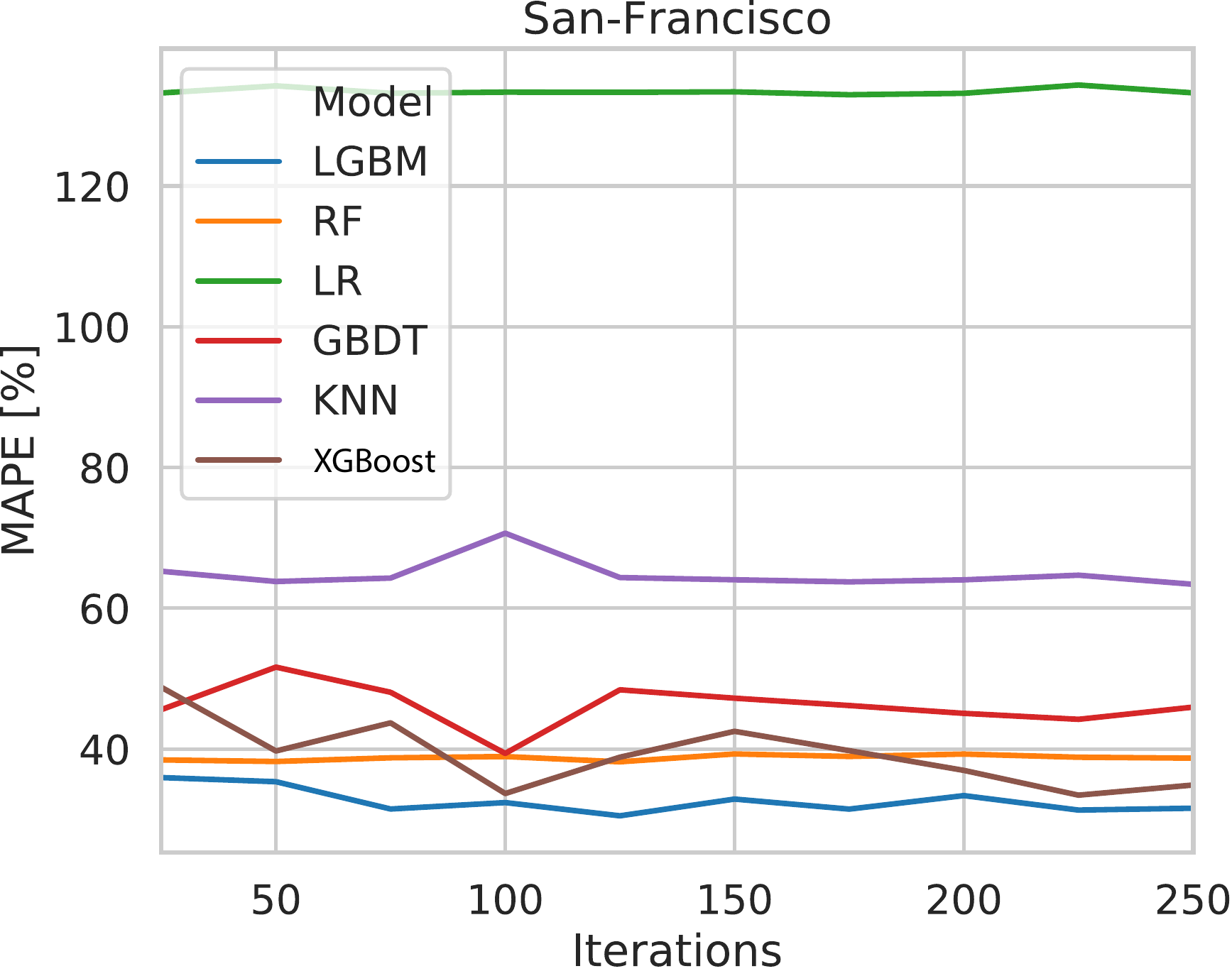}
\centering
\caption{Performance testing of ML models across three different data sets}
\label{fig:iters}
\end{figure}

To determine the optimal number of iterations for models and data sets, we perform iterative testing. The number of random-search iterations is from 25 to 250 with step 25. For example, on \cref{fig:iters}, (Arterial roads, Sydney), we see that XGBoost is the best performing model starting from 120 iterations, and it is already close to optimum starting from 175 iterations.
\revA{The second-best performing model is LGBM, but increasing the number of iterations does not seem to have a significant effect on the model performance which seems to be quite stable without many fluctuations across all evaluation metrics.}
Other methods perform significantly worse (more than 110\% MAPE). For San-Francisco, we see the superior performance of LGBM. The second best is XGBoost. Since there are no metric improvements across iterations for most models, the number of iterations is essential only for XGBoost.
According to the results, we decide to search for hyper-parameters for 250 random parameter combinations for each model. We evaluate each combination using a 5-fold cross-validation and then providing results using a 10-fold cross-validation using best combination.

\subsection{Model Performance Evaluation}

The performance of classification model is evaluated using the Precision, Recall, Accuracy and F1-score and defined as:
\begin{eqnarray}
 Precision &=& \frac{tp}{tp+fp},\\
  Recall &=& \frac{tp}{tp+fn},\\
 Accuracy &=&\frac{tp+tn}{tp+tn+fp+fn},\\
 F_1 &=& 2*\frac{precision*recall}{precision+recall}.
\end{eqnarray}
where $tn$ represents true negatives, $fn$ - false negatives, $tp$ - true positives, $fp$ - false positives. 
\revA{For example, We refer to true positives the incidents which have been predicted to be in a specific class (say short-term) and indeed they were short-term upon validation, false positives the incidents which were predicted to be short term but were not, etc. }

We use F1-score as a target metric for classification experiments as F1 represents the balance between Precision and Recall, and is in general a better performance metric to use when we are facing an uneven class distribution rather than interpreting the Accuracy results which take into consideration the total number of both false positive, false negative together with the true positives and true negatives; therefore for uneven class balances (especially the ones with fewer incident records), one should rely less on Precision and Accuracy metrics.

To evaluate the regression models we use the mean absolute percentage error defined as:
\begin{equation}
    MAPE = \frac{1}{n} \sum_{t=1}^{n} \left|\frac{A_t-F_t}{A_t} \right|
\end{equation}
where $A_t$ are the actual values and $F_t$ - the predicted values, $n$ - number of samples.

Other metrics have been calculated but we will keep them concise due to large amount of experiments to show.

\subsection{Regression scenarios definition}\label{scenarios}

The main objective of the bi-level framework is that the regression accuracy can benefit from different setups for different data subsets. For an even better accuracy compared to the classification problems, we are further developing more complex regression models that can provide incident duration prediction at minute-level accuracy. This is the second step of the bi-level prediction framework to be applied when more precision is needed at the minute level regarding the incident duration length. 
When training such regression models, a crucial step is the size of the data set and the distribution of the target variable (incident duration). Due to the long tail distribution of incident duration and the class imbalance problem previously identified, we need to design and construct various regression models capable of learning from various types of data sets to make accurate predictions. However, with limited information (small data set size), the prediction results can be skewed. This is the primary motivation that led to the construction of several scenarios of model training, validation and prediction that can be applied under both complete or incomplete data sets from traffic centres.
By using the classification thresholds identified previously, we split the traffic incident data set into two subsets: subset A (with duration below threshold $T_c$) and subset B (with duration above threshold $T_c$) as previously defined at the beginning of \cref{S3-Methodology}. We further contract several scenarios of subset combinations for training-validation-testing detailed with the aim of extrapolating the model performance: 

\textit{\textbf{Scenario All-All:}} we use the entire data set and apply several regression models using a 10-fold cross-validation approach and different hyper-parameter search methods. This approach will show us the general performance across various methods.

\textit{\textbf{Scenario A-to-B:}} we use subset A (short-term incidents) for training the regression models and evaluate the prediction on subset B (long-term incidents). In this scenario, we will analyse methods to extrapolate to higher values of the target variable.

\textit{\textbf{Scenario A-to-A:}} we use subset A for training the regression models and predict on subset A. In this scenario, we will analyse the prediction ability of methods with long-term incidents excluded (which includes values from the tail of the incident duration distribution).

\textit{\textbf{Scenario B-to-A:}} we use subset B for training the regression models and predict on subset A. In this scenario, we will analyse methods to extrapolate to lower values of the target variable.

\textit{\textbf{Scenario B-to-B:}} we use subset B for training the regression models and predict on subset B. In this scenario, we will analyse the prediction ability on long-term incidents.

\textit{\textbf{Scenario All-to-A:}} we use all the data for training the regression models and predict on each fold within subset A. \revA{In this scenario, we will analyse the effect of having access to all types of incident logs in the training phases, both long-term and short-term incidents and how their presence might affect or not, the prediction of short-term incidents duration only. This is to evaluate if using all types of records, including rare events, will help or not to predict better short incidents. }

\textit{\textbf{Scenario All-to-B:}} we use all the data for training the regression models and predict on each fold within subset B.  \revA{In this scenario, we will analyse the effect of having access to all types of incident logs in the training phases, both long-term and short-term incidents and how their presence might affect or not, the prediction of long-term incidents duration only. This is to evaluate if using all types of records, including short-term events, will help or not to predict better long incidents. }

\revA{Indeed, from an operational perspective the scenario All-to-All is the ideal situation when traffic management centres would have in their data base both long term and short-term incidents. However, from an operational perspective, several records of short incidents for example and not being kept all the time, while long incidents are often being transferred to various other division if they last more than one day, and they become more of a road infrastructure problem rather than an operational problem which requires constant intervention. Therefore, various incident logs can be imbalanced – some containing more short-term incidents, and others more long-term incidents. The main idea is to provide a good deep dive into the effects of data availability on the model training. For example, training any model only on short term incidents as these are the only ones available will most likely not provide good prediction results in case of long-term incidents and vice versa. }

%% file: sections/3B-ORM-Methodology.tex
\subsection{Outlier removal methods (ORM)}\label{ORM}
As previously discussed in \cref{2-Data-sources}-\cref{fig:scheme} during the data profiling, we observed that the traffic incident logs contain outliers appearing as either minor incidents, rare traffic incidents with highly long duration and/or as errors in incident reports. Therefore, to reduce the side-effect of outliers on all models, we deploy two commonly used outlier removal methods.
The IsolationForest (IF) \citep{isolation} is an outlier removal method, which uses forests of randomly split trees. For each tree, the method randomly selects a feature and a random feature value. The data set is divided into two parts in each step until each data point becomes ``isolated'' (split from the rest of the data). If the data point is an outlier, it will have a small tree depth (e.g. data point gets quickly separated from the rest by selecting values in just a few features).
Tree depth is then averaged between all the ``isolation'' trees and considered an anomaly score (e.g. if the average tree depth for a point is 1.3, the point is easily separable after a small number of splits).
LocalOutlierFactor (LOF) \citep{lof} is another outlier removal method, which estimates the anomaly score from local deviation of density within k-nearest neighbourhood. LOF relies on the calculation of a local reachability density (LRD), which represents the inverse of the average reachability distance (RD) of neighbouring data points from the selected data point. Reachability distance (RD) represents the distance to the most distant neighbour within a k-sized neighbourhood (k is also hyper-parameter). LOF of data point then represents the relation between LRDs of neighbours and its LRD and can take values: a) above 1 (higher LRD than its neighbours), b) below 1 (lower LRD than neighbours) and c) equal to 1 (data point has the same density as neighbours). According to the LOF score, we can sort data points and select specific per cent of data points, which have the highest LOF to be eliminated. LOF method relies on the fact that outliers belong to the area where the density of data points is low, while regular data points belong to the high-density area. To summarise, the above outlier removal procedures are applied in conjunction with the proposed optimisation framework and regression models and show a significant improvement in prediction accuracy as further detailed in \cref{IEO_results}.

\revA{
\subsection{Outliers from ORM point of view}

We would like to make the observation that all the incidents have scalar degree of anomaly when applying outlier removal method. herefore, there are no discrete categories of outliers and normal data points from an outlier method point of view. We simply remove a per cent of data points (e.g. $2\%$) with the highest degree of anomaly. These points are either easily separable using IF method (tree-based) or remain on a low local density area for the LOF method (distance-based). 

So does our outlier removal method actually remove long-term incidents failing to distinguish them from outliers? ML methods in our case, find outliers not only by the value of duration but by including all reported variables (e.g. 25 in the case of Arterial roads). Our aim in this work is to remove incident reports which have very rare characteristics overall, which are also known to negatively affect the ML method performance \cite{lu2021detecting}.

\begin{figure}[pos=h,align=\centering]
\centering
\includegraphics[width=0.32\textwidth]{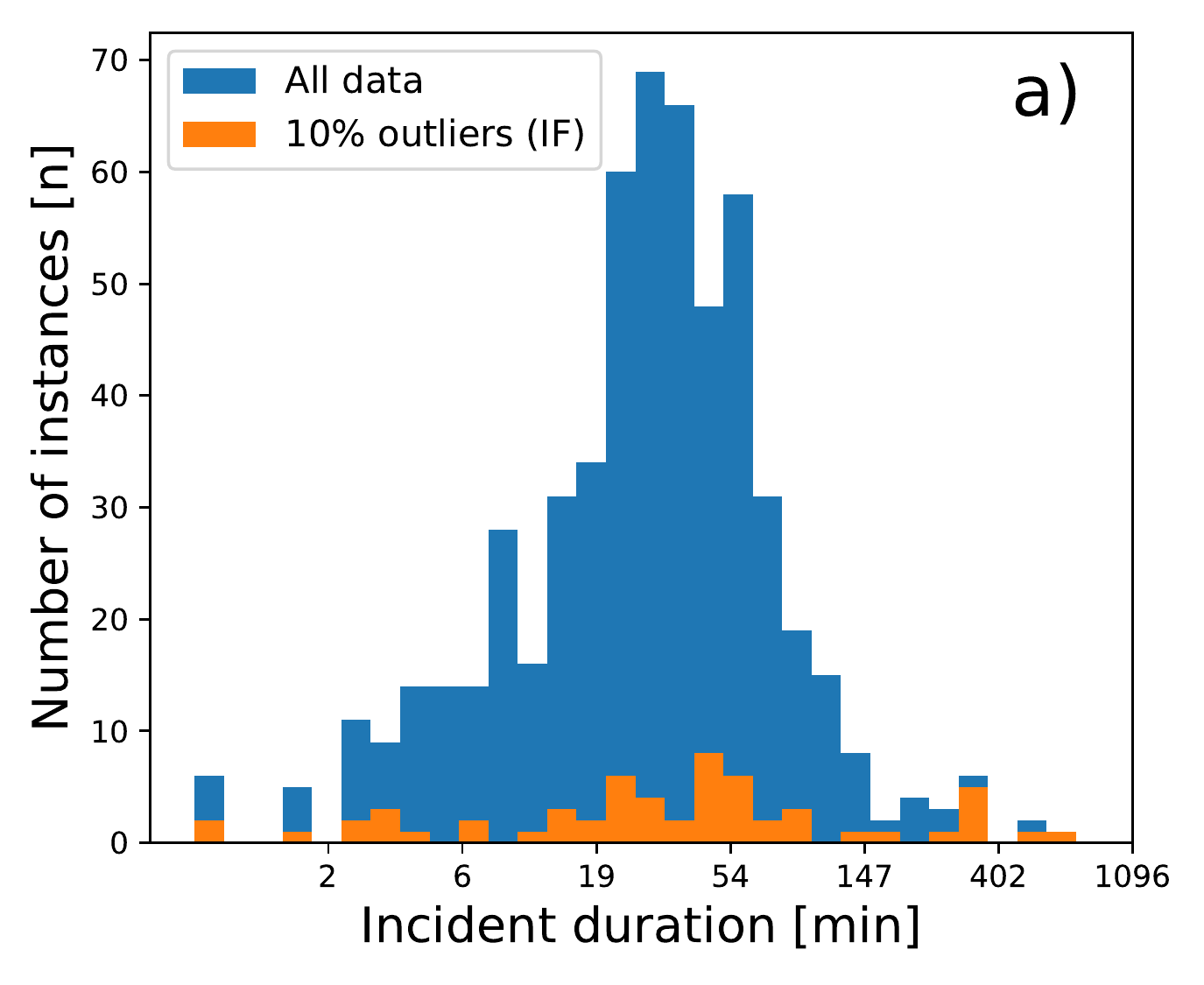}
\includegraphics[width=0.32\textwidth]{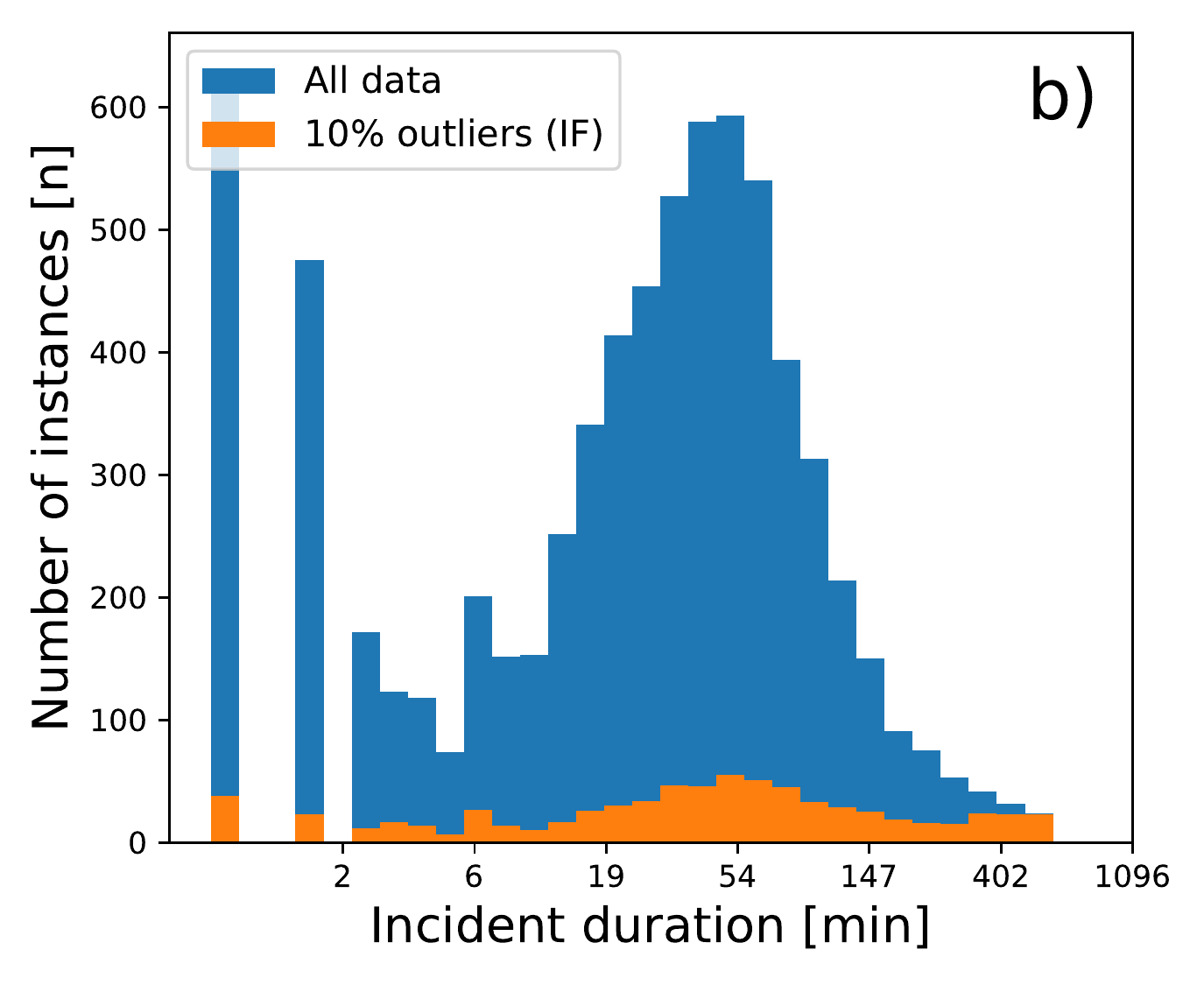}
\includegraphics[width=0.32\textwidth]{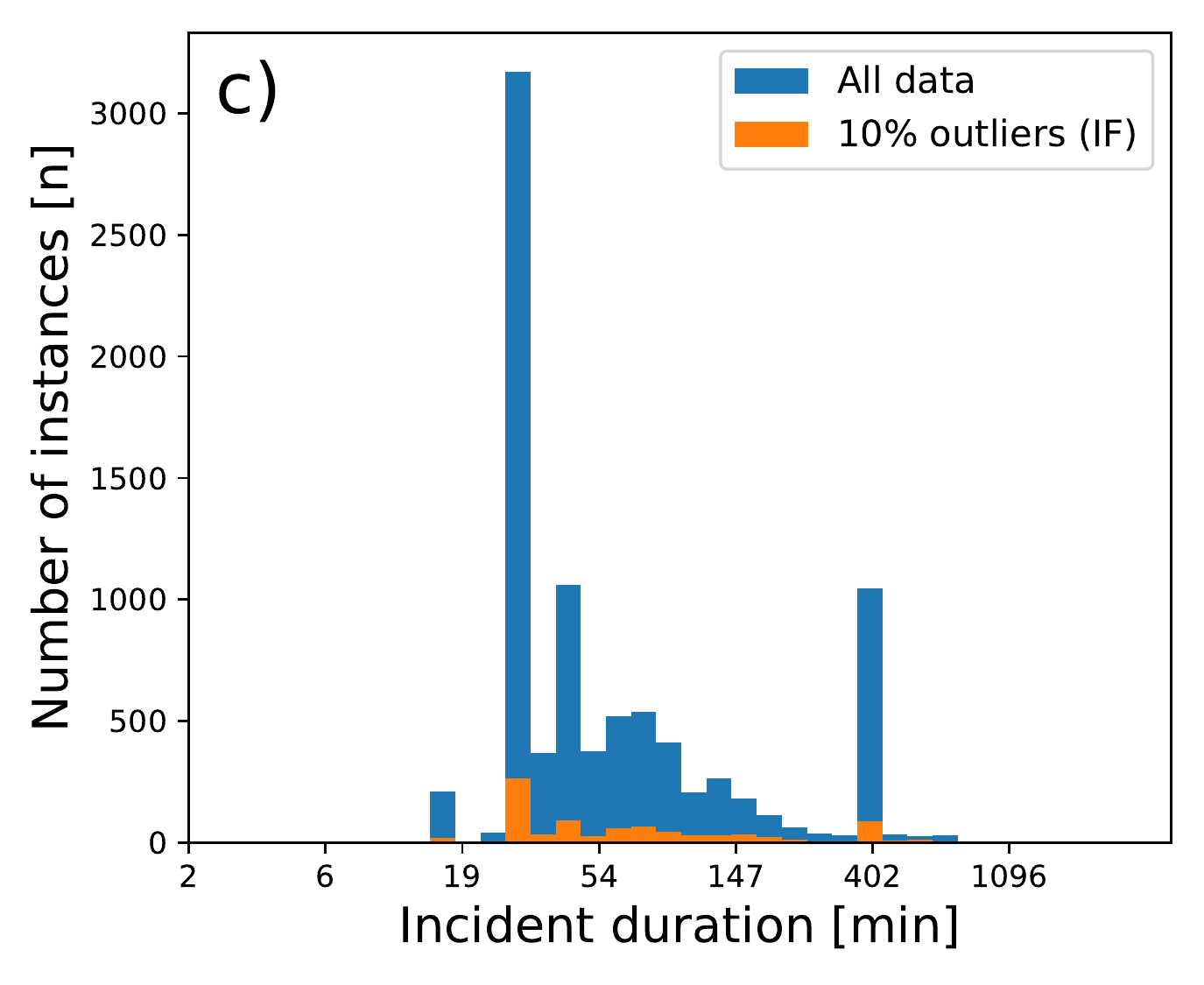}

\centering
\caption{Data sets with 10\% of points with the highest anomaly score removed using IsolationForest: a) Arterial roads, Sydney, Australia b) M7 Motorway, Sydney, Australia c) San-Francisco, USA}
\label{fig:orm3A}
\end{figure}

\cref{fig:orm3A} Showcases Data sets with $10\%$ of points with the highest anomaly score removed using IsolationForest: a) Arterial roads, Sydney, Australia b) M7 Motorway, Sydney, Australia c) San-Francisco, USA. By performing experiments with an outlier removal (isolation forest, $10\%$ of point with the highest anomaly rate removed), we see how many incidents were removed according to each duration interval. An important finding is that outliers do not reside in the area of long-term incidents but rather scattered among the general population of incidents. 
}

\subsection{Intra/Extra Joint Optimisation for ML regression prediction (IEO-ML)} \label{IEO}

\begin{figure}[pos=h,width=16cm,align=\centering]
\centering
\includegraphics[width=0.9\textwidth]{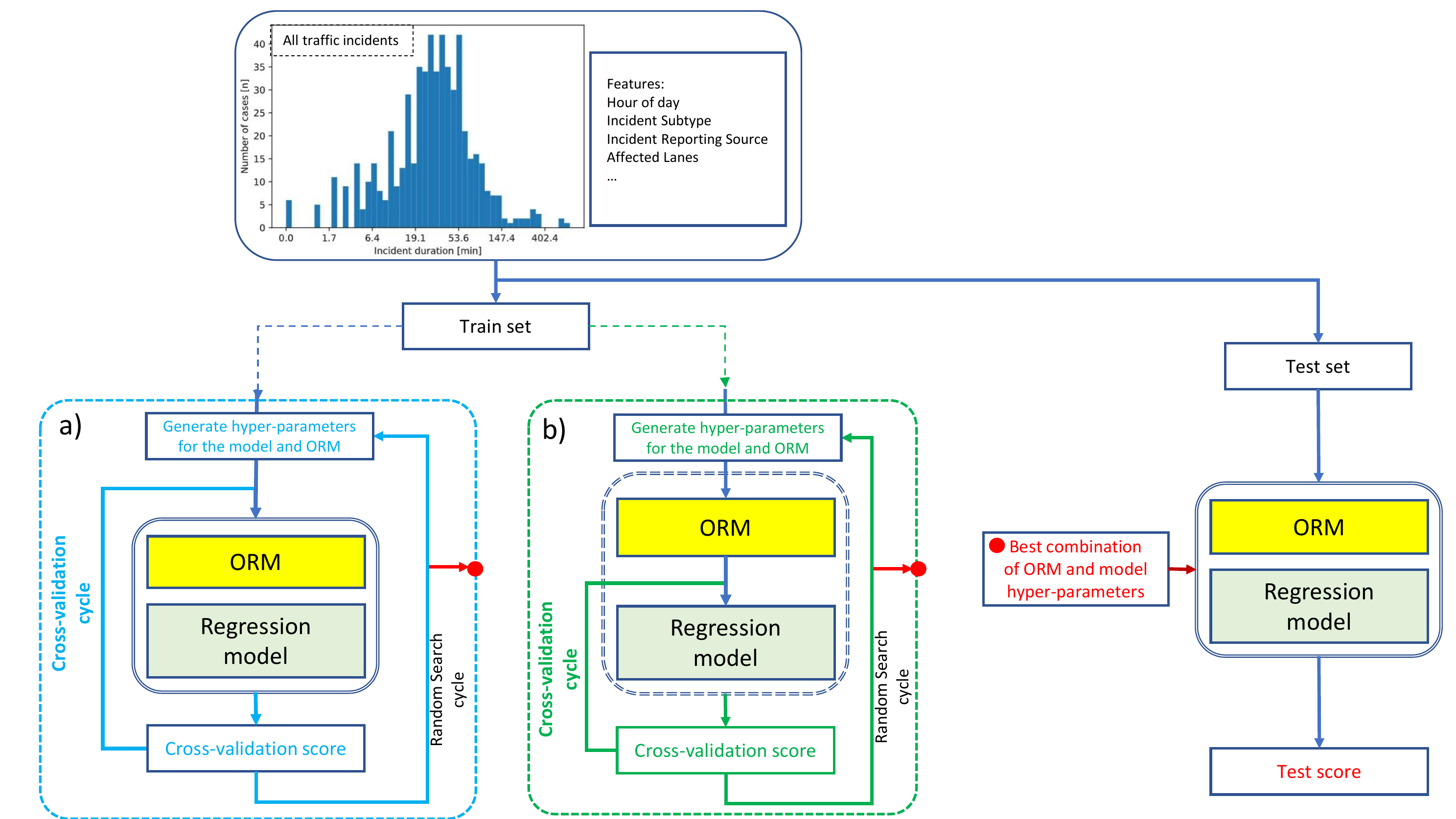}

\centering
\caption{IEO-ML algorithm with a) Intra joint optimisation schema for the EO-ML algorithm, b) Extra joint optimisation schema for the IO-ML algorithm. Red dot on schema blocks represents output in the form of the best combination of ORM and model hyper-parameters}
\label{fig:io-eo}
\end{figure}

This section presents our novel enhancements of ML regression models by constructing an \textbf{intra/extra optimisation technique} to jointly optimise the hyper-parameters of the regression models together with previous outlier optimisation methods. In the rest of the paper, we denote this approach as \textbf{IEO-ML}, where ML is one of the regression models previously described (GBDT, XGBoost, RF, kNN, LR, LGBM). We introduce this approach for multiple reasons:
1) the traffic incident data is prone to errors during the data collection, which is attributed to human factors (e.g. presence of incidents with 0 and 1-minute durations, for example),
2) an outlier removal performance cannot be assessed on the new dataset with no marking for outliers; thus, we can assess outlier removal performance by looking at model performance with outlier removal applied, use joint outlier removal and modelling to assess the outlier removal performance metrics,
3) both the outlier removal method and models have hyper-parameters forming a single hyper-parameters space,
4) we assume that the outlier removal can be performed either inside (Intra - see \cref{fig:io-eo}a)) or outside (Extra - see \cref{fig:io-eo}b)) of the cross-validation cycle, and we evaluate the effect of such an approach on the model performance, 5) Intra joint optimisation can provide a more effective outlier removal since common hyper-parameters will be found for different data subsets, which allows ORM to be adapted to different possible combinations of incidents in case of the model deployment and prediction on the newly acquired incident log.
Overall we want to compare and observe the impact of each technique on the accuracy of regression models and detect the best combination of Intra/Extra joint optimisation and various ML regression models.

Further, we present our proposed IEO-ML algorithm in conjunction with the two outlier removal methods IF and LOF, and several regressions models. Our approach explores the following combinations of ML models in selected working base (decimal or logarithm) with outlier removal and intra/extra joint optimisation; for example, we denote as \textit{iLOF-LT-MLmodel} a ``joint optimisation of any available baseline ML model with LOF in a log-transform base within a cross-validation cycle (an intra optimisation)''. As an observation, ORM has specific hyper-parameters but one parameter in common - the percentage of removed samples, which we assume to be outliers (ORperc). Thus, to solve the ORM problem, we assume that the amount of outliers in each data set (ORperc) can take values up to 5\%.
EJO is performed only once and before the cross-validation cycle, but IJO is performed within each fold in a number of times which is equal to the number of folds. Thus, ORperc has values in $\{0,1 \ldots 5\%\}$ for EJO, in $\{0, 1/5, \ldots , 5/5\}$ for IJO to ensure a comparable amount of removed samples from both approaches. 
Results for all combinations of the proposed approach inside the incident duration prediction framework are further provided in \cref{IEO_results} for eLOF-ML models, iLOF-ML, iIF-ML, eIF-ML (e.g. eIF-ML is a ``joint ML optimisation using IF optimised outside (e) of the cross-validation cycle'').

\newcommand\mycommfont[1]{\footnotesize\ttfamily\textcolor{blue}{#1}}
\SetCommentSty{mycommfont}

\vspace{0.1cm}

\begin{algorithm}[H]

\KwData{Traffic incident reports (feature vector  $X$, duration vector $Y_r$)}
\KwIn{HPSm (Hyper-Parameter Space for Model),

ORM: Outlier Removal Method,

HPSor: Hyper Parameter Space for ORM,

Model: ML regression model $\in\{GBDT, XGBoost, RF, kNN, LR, LGBM\}$,

Iters: Number Of Iterations (number of random search steps for hyper-parameter optimisation),

Folds: number of folds for cross-validation,

sample: function for random sampling from the hyper-parameter space,

FoldIndexes: function to get sample indexes for training folds and test fold,

extra: boolean variable stating the use of extra joint optimisation,

intra: boolean variable stating the use of intra joint optimisation,
}

split: function to split data set into two parts - train/test and validation parts

\KwOut{Predicted duration vector $Y_r$}

$x_{tr},y_{tr},x_{te},y_{te} = split(x,y);$

$P = \left[\right]$ \tcp*{temporary cross-validation prediction vector}

$results=\left[\right]$

 \For{$it \leftarrow 1 .. Iters$}{
    $HYPm \leftarrow sample(HPSm)$
    
    $HYPor \leftarrow sample(HYPor)$
    
    $idx_{train}=\left[\right]\;$ \tcp*{indexes of train samples}
    $idx_{valid}=\left[\right]\;$ \tcp*{indexes of validation samples}
    
    $res=0$ \tcp*{scoring results}
    
    \uIf{extra}{ 
          $x = \textbf{ORM}(x, HYPor)$ \tcp*{if EO then filter the outliers from the feature vector}
    }
    
    \For{$k \leftarrow 1 .. Folds$}{
  
      $idx_{train},idx_{valid} = FoldIndexes($x$,$k$)$;
      
      $x\_{train} \leftarrow x_{tr}[idx^{train}_0], ...,x_{tr}[idx^{train}_N]$ \tcp*{array of feature vector samples for training}
      
      $y\_{train} \leftarrow y_{tr}[idx^{train}_0], ...,y_{tr}[idx^{train}_N]$ \tcp*{array of duration vector samples for training}
      
      $x\_{valid} \leftarrow x_{tr}[idx^{valid}_0], ...,x_{tr}[idx^{valid}_N]$
      
      $y\_{valid} \leftarrow y_{tr}[idx^{valid}_0], ...,y_{tr}[idx^{valid}_N]$
    
      \uIf{intra}{ 
        $x^{train}_{filtered} = \mathbf{ORM}(x\_{train}, HYPor)$ \tcp*{if IO then filter outliers}
      }
      
      $initialize\_model(Model, HYPm)$ \tcp*{random hyper-parameter initialisation}
      
      $m \leftarrow fit\_model(Model,x^{train}_{filtered},y^{train}_{filtered})$ \tcp*{fitting the model to the filtered train set}
      
      $y\_{pred} \leftarrow predict(m, x_{valid})$ \tcp*{performing predictions}
      
      $P = \left[ P ; y_{pred} \right]$
     }
     
     $res \leftarrow Metric(y_{tr}, P)$ \tcp*{scoring the accuracy of predictions using performance metric}
     
     $r = \left[\right]$ \tcp*{Initializing hash-array}
     
     $r\left['metric'\right]=res$ \tcp*{populating hash-array with resulting metric}
     
     $r\left['HYPm'\right]=HYPm$ 
     
     $r\left['HYPor'\right]=HYPor$ 
     
     $results = \left[ results; r \right]$
     \tcp*{collecting results for sampled hyper-parameters into array}
     
     
 }
 
 $best = sort(results,by= 'metric')\left[0\right]$ \tcp*{selecting the best combination of hyper-parameters}
 
 $initialize\_model(Model, best\left['HYPm'\right])$
 
 $x^{tr}_{filtered},y^{tr}_{filtered} = \mathbf{ORM}(x_{tr}, best\left['HYPor'\right])$ \tcp*{applying ORM to the training set}
 
 $m \leftarrow fit\_model(Model,x^{tr}_{filtered},y^{tr}_{filtered})$ 
 
 $Y_r \leftarrow predict(m, x_{te})$ \tcp*{performing predictions}

 \caption{Intra and extra joint optimisation algorithm with outlier removal and ML regression modelling.}
 \label{IEO_ML_alg}
\end{algorithm}

\vspace{1cm}

The algorithm represents the modified cross-validation cycle within the randomised hyper-parameter tuning procedure. We use multiple iterations (in fact, attempts) to find optimal parameters both for the selected model (HYPm) and the outlier removal method (HYPor). On every iteration, we sample hyper-parameter sets from hyper-parameter spaces. Then, if extra joint optimisation selected, an outlier removal procedure performed using all the data before the fold-rotation cycle. Then we perform an n-fold cross-validation procedure, where we split data set into training and testing parts (by preserving ratio between them at F-1:1, where F is the number of folds) according to sequentially generated indexes (e.g. in case of 500 data points, fold 0 will represent indexes from 0 to 100 for the testing set, rest of the folds - indexes from 100 to 500 for the training set, fold 1 - 100-200 for the testing set, rest - 0-100 and 200-500 for the training set, etc). Then, if intra joint optimisation is selected within the cross-validation cycle, we perform outlier removal with sampled hyper-parameters using only the train subset within each train-test split. Hyper-parameters for ORM include the percentage of samples to be removed. After removing outliers, we train a model using a train set and make predictions on the test set. 

All arrays with actual and predicted samples collected to be used after the fold-rotation cycle for the model accuracy estimation using specified metric. Since we are selecting test folds in order and making predictions on them, the predicted duration vector will be composed of prediction results composed of these folds. So, first, we collect the resulting metric together with hyper-parameters, actual and predicted labels. To collect data we use hash-array, which is represented as an array, where each element can be addressed by name and not by index as for conventional array. Then we perform the sorting procedure, which will order solutions according to the resulting metric, where we select the best combination of hyper-parameters. Furthermore, finally, we obtain the predicted duration vector by filtering data using the ORM method, training model on the train/test part and making predictions on the validation part. 

%% file: sections/5A-Classification-Results.tex
\section{Incident classification results}\label{S5-AResults}

This section details the results of the first layer of the bi-level prediction framework related to the classification prediction findings, either via a standard binary classification with varying threshold analysis or via a multi-class classification enhanced by outlier removal procedures.

\subsection{Binary incident classification results using varying split thresholds}\label{binary_var_Tc}

The first classification problem that we address is to predict whether an incident duration will be lower or greater than a selected threshold (we classify short-term versus long-term traffic incidents), which can then be used to supply the initial assessment needs of the traffic management centre (TMC) under fast decision times. For example, an operational clearance threshold for the Sydney TMC has been currently established at 45min based on previous operational field experience; however, choosing a fixed threshold for classification can have a significant impact on the results of any prediction algorithm and is highly dependent on the incident duration distribution chart (as represented in Fig. \cref{fig:dataprofiling}-g, h, i). 
\cref{fig:scheme} showcases the data split for the binary classification problem where the threshold $T_c$  (dashed red line) is varying according to the two set-ups mentioned above: every 5 minutes ($T_c \in \{20, 25,\ldots,70\}$). We name as Subset A all incident duration records which are lower or equal to $T_c$, $(\text{if } y_i\leq T_c)$, and as Subset B all the incident duration records which are higher than $T_c$ $(\text{if } y_i > T_c)$. Based on the variation of $T_c$, the size of Subsets A and B will have an impact on the prediction algorithms and this impact is further quantified.

\begin{figure}[h]
\centering
\includegraphics[width=1\textwidth]{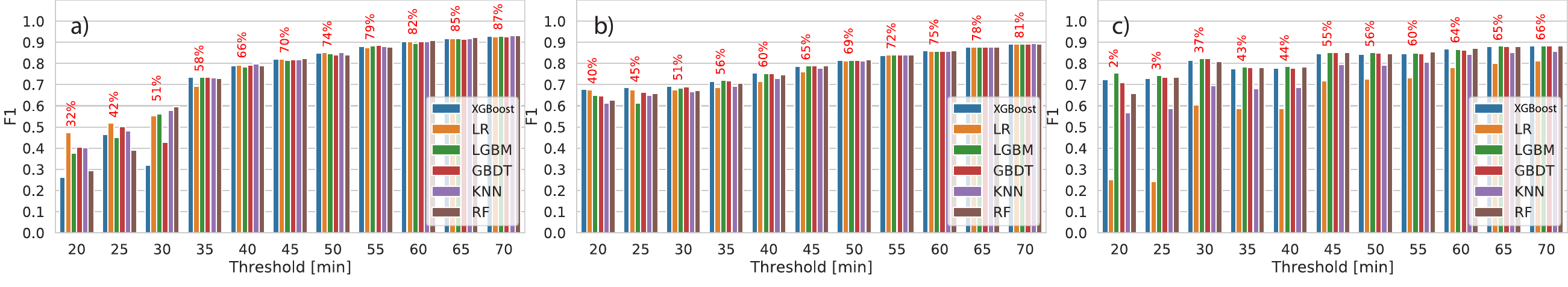}
\caption{Incident duration classification using varying thresholds for a) data set AR b) data set M c) data set SF. The red percentage above each set of ML results indicate the percentage split of Subset A and B for that particular $T_c$. }
\label{fig:rate5}
\end{figure}

The results of the binary classification approach of incident durations using a varying split threshold are detailed in \cref{fig:rate5} (for a 5-minutes frequency split) across all data sets. More specifically, \cref{fig:rate5} presents the F1 results obtained for each ML model that we have developed (XGBoost, LR, LGBM, GBDT, kNN, RF); we observe that other performance metrics have been calculated such as Accuracy, Precision and Recall and these are provided in the \cref{appendix_A}).

For example, \cref{fig:rate5}a) showcases the classification results for data set AR in which the blue bar represents the F1-result of the XGBoost classifier (F1=0.28) when the data set has been split in Subset A containing incidents with a duration less than 20min (32\% of all incident records fall in this subset) and Subset B containing incidents with duration higher than 20min (the rest of 68\% of incident records). Therefore, the percentage numbers written in red above each ML result represent the percentage of records lower than the $T_c$ threshold chosen for this experiment.

The split around $T_c=20min$ is not ideal given the data imbalance ($32\%$ versus $68\%$) and the low F1 score; therefore further variations have been undertaken which have reported an increased $F1=0.8$ for $T_c=45min$. According to these results, if we use the best performing binary classifier, we need to select a threshold between 35 and 50 minutes because: 
a) it will reduce the imbalance between classes (and thus reduce the effects of imbalanced classification, which is vital for modelling when using a small data set); 
b) there is only a tiny improvement in F1-score after $T_c>40$min; 
c) it will be a reasonable split for short incidents lower in terms of field operation management.
An exciting finding is revealed for $T_c\in\{20,25\}$min: we record an overall lousy performance across all ML models in all data sets (F1-score less than 0.5) while some did not even take effect, such as GBDT; for this reason, we exclude from consideration any thresholds which provide an F1-score of less than 0.5. Furthermore, we set our minimum acceptable F1-score to 0.75, and any model performing lower than this threshold will not be considered for further optimisation.
By analysing all sub-figures in \cref{fig:rate5} which provide both a good F1 score and class balance, we conclude that the optimal thresholds for the binary classification problem are the following: a) $T_c=40$min for the arterial road network in Sydney (\cref{fig:rate5}a: $F1=0.79$ and a class balance of 66\% for small incident duration), b) $T_c=45$min  for the motorway network in Sydney, (\cref{fig:rate5}b: $F1=0.75$, class balance = 65\%) and c) $T_c=45$min for the San Francisco network (\cref{fig:rate5}c: $F1=0.83$, class balance=55\%).

The other important finding is the cases when $T_c>45$min which present a significant improvement across all models on all performance metrics, with the best result being the one when Subset A incorporates all incidents lower than 70min (which represents the majority of incidents); this is easily explained by the fact that we use almost all the entire data set for training of the models. However, the binary classification can be a rough estimate. If TMCs need a higher prediction precision instead of incidents less than 45min or higher (which can last up to several days), then several regression and multi-class classification models are needed to provide more precise predictions. These will be further detailed in Sections 6 and 7.
We will further use the detected optimal thresholds for each data set to perform the split between subset A and B in various scenarios of the incident duration regression problem.

Tree-based models yield similar results. However, in multiple cases (e.g. 35, 45, 50, and 60-minute thresholds for data set AR, 25, 30, 40, 60-minute thresholds for data set M), XGBoost produces a slightly better result than other tree-based models. Thus, we are selecting XGBoost as the best model for the incident duration classification.

\subsection{Classification with outlier removal}\label{ORM_class}

\begin{figure}[pos=h,width=16cm,align=\centering]
\centering
\includegraphics[width=0.96\textwidth]{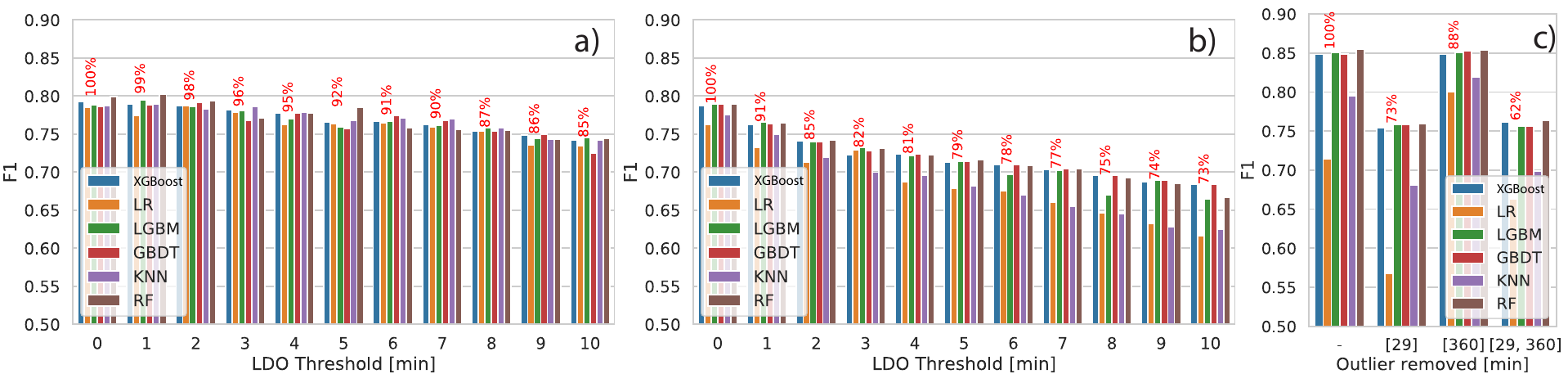}
\caption{Outlier removal for a) data set AR b) data set M c) data set SF}
\label{fig:or}
\end{figure}

After selecting the optimal thresholds for binary classification, we further assess the effect of: a) \textbf{low-duration outliers} (LDO) (which we define as reports of incidents with zero or less than a few minutes duration) and b) \textbf{high-duration outliers (HDO)} as in the San-Francisco data set, by trying different outlier removal procedures, as depicted in Fig \cref{fig:or}. 

For example, an LDO Threshold of 1min represents removing outliers below 1 minute (e.g. 0min) and the percentage above each removal test. For example, 99\% indicates the number of samples remaining after such removal.
Removing these outliers is essential since it represents errors in the incident reporting and may affect the accuracy of prediction. For example, \cref{fig:or}a represents the LDO removal from the data set AR, up until 10min reported incident durations; by removing these outliers, we observe that the F1-score does not fall below the acceptable threshold of 0.75 until 5min (this indicates that removing all accidents reported with a duration of 0 or lower than 5min does not reduce the model performance. Therefore, we applied an LDO removal for all traffic incidents for this data set with a duration below 5min.
For the data set M, the effect of LDO outlier removal is more significant, as depicted in \cref{fig:or}b. This data set contains a lot of incidents with duration of 0 and 1 minute (which represents almost 15\% of the entire data set); by removing these, we observe that the highest F1-score drops down to 0.74 across all ML models, which falls below the acceptable threshold for a good prediction accuracy). Therefore, we decide to remove only incidents with duration of 0min or 1min from this data set. Lastly, in the case of the San-Francisco data set, we have a completely different range of outliers since there are no incidents reported with a duration of fewer than 17 minutes (see \cref{fig:or}c). There are multiple incidents cleared off at around 29min and 360min (as represented as well in \cref{2-Data-sources}, which can be identified as HDO. However, by removing these HDO data points from the ML model training (representing almost 38\% of all incident records), we observe a depreciation of the F1 score from 0.85 to 0.76 for XGBoost, while some models dropped to lower values below 0.7). Therefore, the removal of HDO for the San Francisco data set can not be adopted due to several reasons: 1) we cannot separate ``rounded'' duration from actually reported duration, 2) the amount of these values is almost half of the data, which becomes property of the data set, 3) these outliers still convey information related to the separation between short-term and long-term traffic accidents and 4) all models perform better when using the entire data set than with outlier removal, which makes the ORM procedure in this case non-necessary. Finally, we observe that the outlier procedure is highly related to the specificity of the data set and the incident area location, not by making default assumptions on either LDO or HDO.

\subsection{Multi-class classification} \label{multi_class_classif}

While binary classification can provide fast insights in the overall incident duration, traffic incidents can have more precise duration definition and can be split (based on the histogram profiling) into short-term, mid-term, long-term. In this case one needs to solve a multi-class classification problem. \revA{We have split this problem in two subsection in which we analyse the impact of choosing three equally sized classes, versus quantile varying split thresholds and analyse the best approach.

\subsubsection{Equally split multi-class classification}
}

Firstly, we analyse the impact of using equally-sized classes (based on duration percentiles of almost $33\%$ from each data set). We use F1-macro to assess the performance of a multi-class classification, defined as the unweighted average of class-wise F1-scores:
\begin{equation}
    \begin{array}{rcl} \text{F1-macro} & = & \frac{1}{N} \sum_{i=0}^{N} {\text{F1-score}_i} \\ \end{array}
\end{equation}
where i is the class index and N is the number of classes.
\cref{F1_score_multi_class} contains the F1-macro scores across all three data sets for a 3-class prediction problem which can be calculated across each data set independently. For example, $C1$ for data set AR in Sydney contains incidents between $0-24$min, while $C1$ for the SF data set contains incidents between $0-30$min; similarly, the $C3$ class for the SF data set contains substantial incidents which can reach up to 2,715min (45h) (this is consistently larger than 710min or 595min in Australia). The F1-macro score is aggregated across all classes, and a low value (below 0.5) indicates that we cannot use a 3-class split for the data set AR (F1-macro=0.35) and M (F1-macro=0.46), but we can do so for the data set SF (F1-macro=0.72). The significant difference between these data sets is the number of records (584 incident records for the data set AR versus 8,754 records for the data set SF), which may affect model performance.
The precision of predictions on the data set indicates how many classes we can have to distinguish traffic incidents by duration. However, each data set's specificity seems to dictate the best classification approach to be done and further justifies the need for a more refined regression prediction approach.

\begin{table}[pos=h,width=20cm,align=\centering]
	\centering
\begin{tabular}{llllll}
\toprule
Dataset &      $[0-33\%]_{C_1}$ & $[33-66\%]_{C_2}$ & $[66-100\%]_{C_3}$ &  F1-macro(3-class) & F1 (2-class)\\
\midrule
Data set AR &  0-24 min & 25-44 min & 44-710 min & 0.35 & 0.79\\
Data set M &   0-24 min  & 25-54 min  & 54-598 min & 0.46 & 0.74\\
Data set SF &  0-30 min & 31-71 min & 72-2,715 min & 0.72 & 0.85\\
\bottomrule
\end{tabular}
\caption{Multi-class classification results for equally-sized 3-class split}
\label{F1_score_multi_class}
\end{table}

\subsubsection{Varying multi-class classification via quantile split}
\revA{
To analyse the effect of splitting data into more varying groups we performed a multi-class classification procedure using quantiles and the F1 results are provided in Figure \ref{fig:mc} for three data sets: Figure \ref{fig:mc}a) when using the Arterial Roads in Sydney, Australia and b) when using M7 Motorway data set and c) when using the San Francisco data set. The result metric represents an average of F1-scores across classes, where multi-class classification performed as 3 one-vs-all classifications. }

\revA{
The low/high threshold matrix represented in Figure \ref{fig:mc} indicates a 3-class split performance and allows for the modelling of different size groups separated by quantile thresholds. As an example, the Ox axis in Figure \ref{fig:mc}a) represents the first threshold split ranging from [10\% to 80\%], while Oy represents the second threshold split percentage ranging from 20\% to 90\%. The coloured dots represent the F1 scores obtained when splitting the data according to the two thresholds; for example, the combination quantile pair of [10\%;20\%] gives an F1 score of 0.34, meaning a multi-class split of data logs between the following three classes $\{C1=[0-10\%]$, $C2=[10\%-20\%]$ and $C3 =[20\%-100\%]\}$ does not provide good accuracy. Instead, when using the first quantile threshold of 0.3 and the second quantile threshold of 0.6 (meaning $\{C1=[0-30\%], C2=[30\%-60\%]\ and\ C3 =[60\%-100\%]\}$), we obtain the highest F1-macro score, F1 = 0.44.

\begin{figure}[pos=h,width=16cm,align=\centering]
\centering
\includegraphics[width=0.43\textwidth]{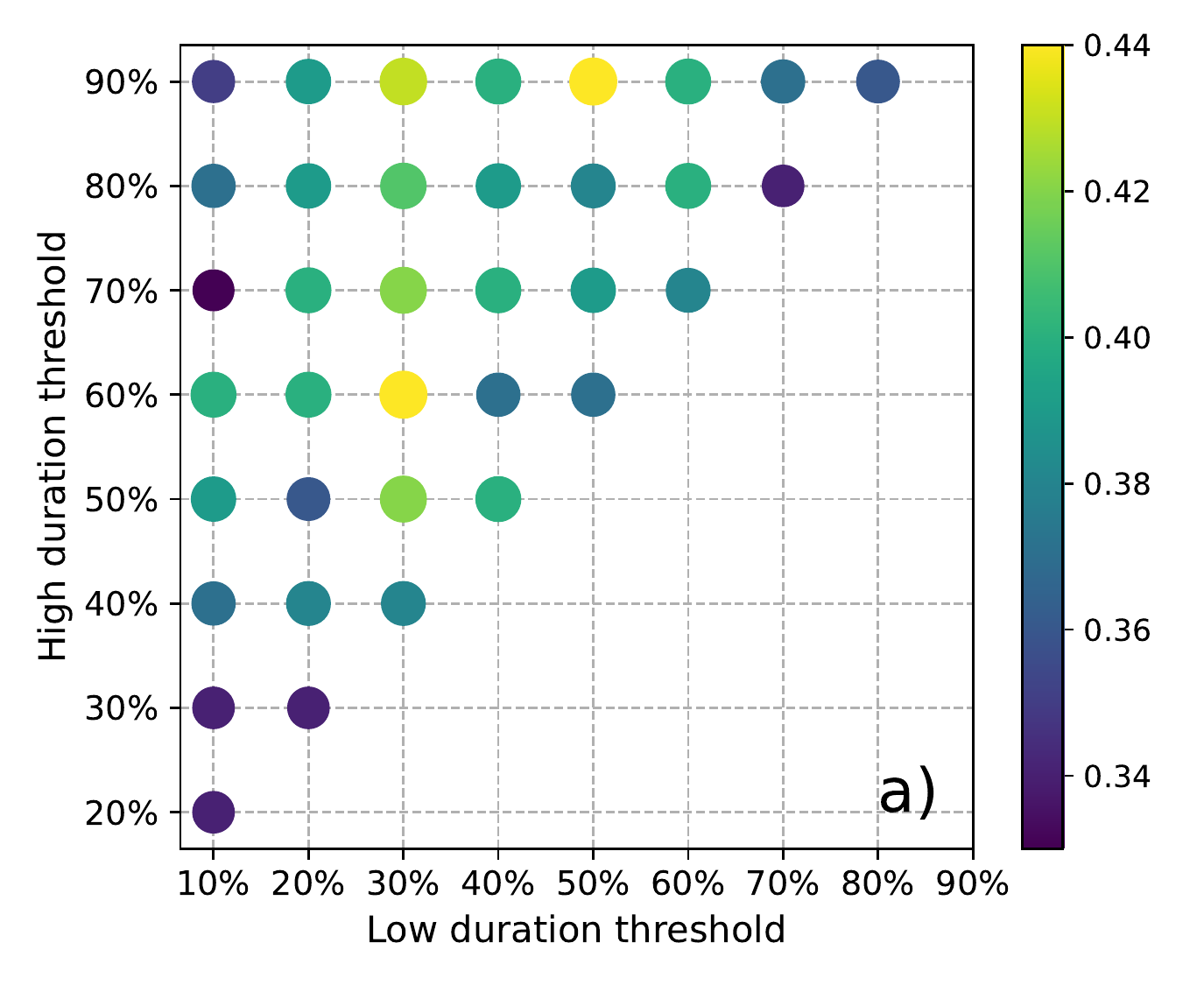}
\includegraphics[width=0.43\textwidth]{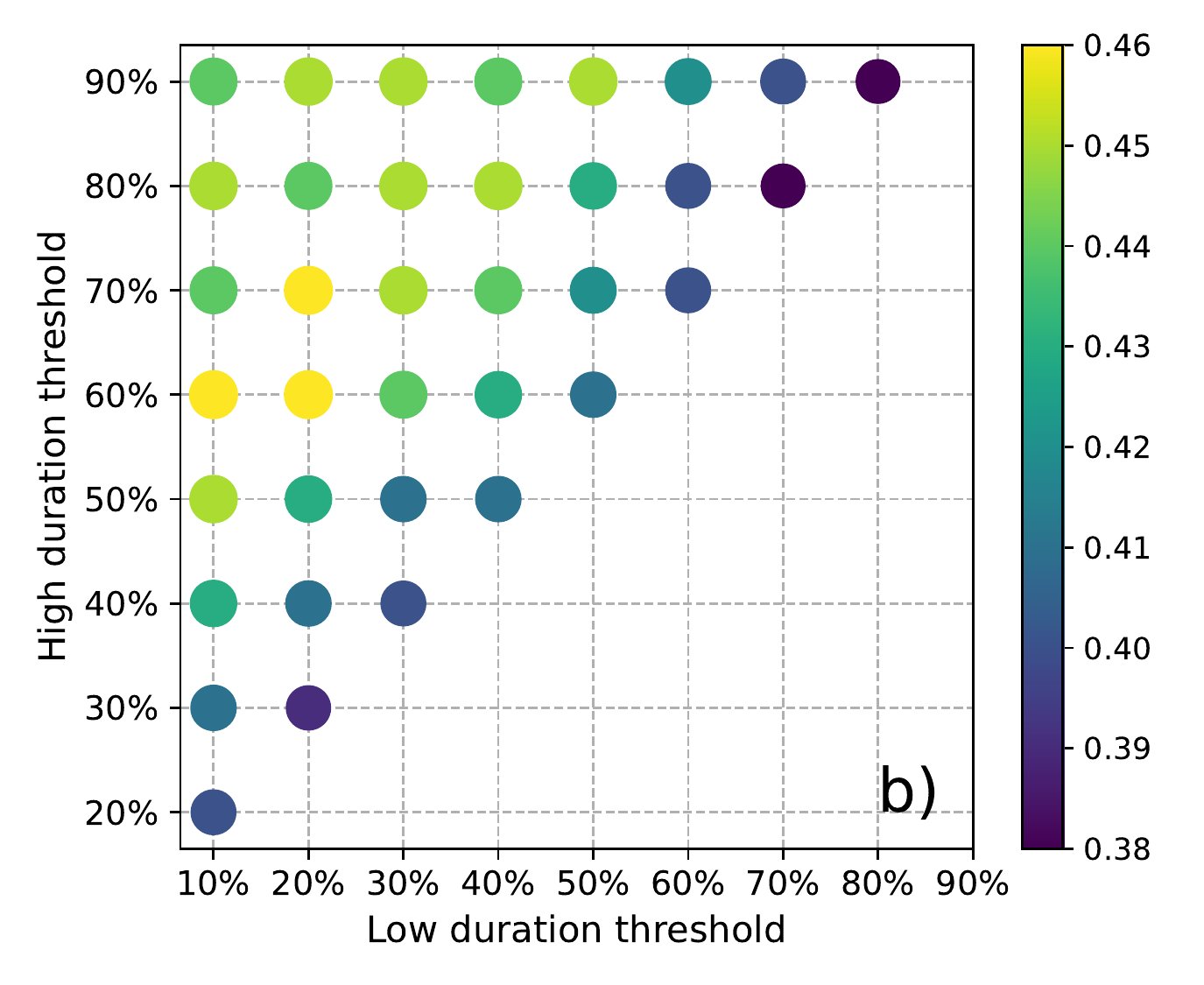}
\includegraphics[width=0.43\textwidth]{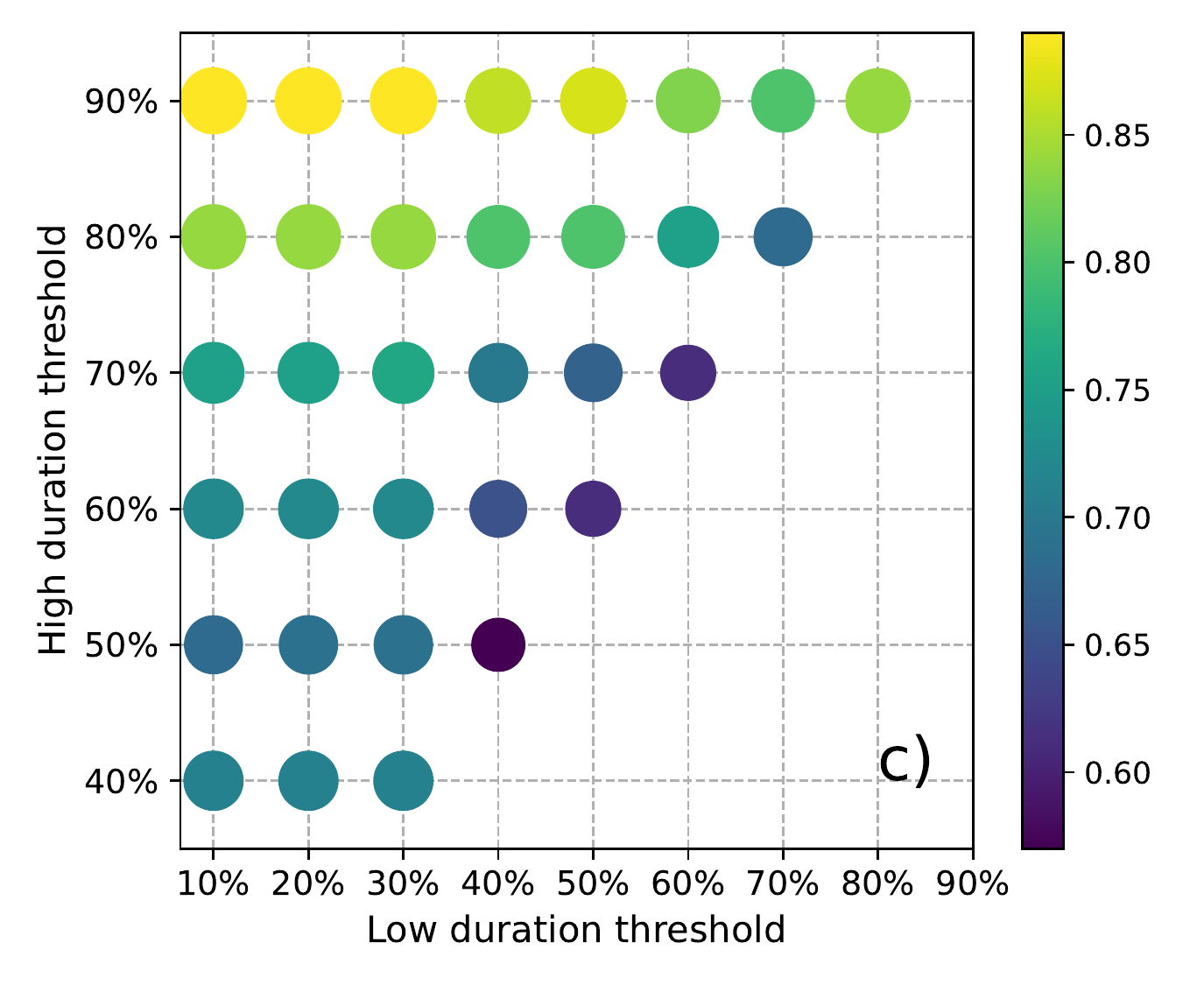}
\caption{Multi-class (3-class) classification using quantile splits for a) data set AR b) data set M c) data set SF}
\label{fig:mc}
\end{figure}

}

\revA{

In the case of M7 Motorway (see Figure \ref{fig:mc}b), we obtain the best performance for 20\% and 60\% quantile thresholds (meaning  $\{C1=[0-20\%], C2=[20\%-60\%] and C3 =[60\%-100\%]\}$; ${20\%, 40\%, 40\%}$ size grouping. Other options include $\{20\%, 70\%\}$ and $\{10\%, 60\%\}$ duration thresholds.

In the case of San-Francisco (see Figure \ref{fig:mc}c), we obtain the best performance for 10\% and 90\% quantile thresholds (meaning  $\{C1=[0-10\%], C2=[10\%-90\%]$ and $C3 =[90\%-100\%]\}$; this means that the best data split when using quantile thresholds for San Francisco is a $\{10\%, 80\%, 10\%\}$ size grouping. This is highly explained by the incident distribution plots for the San Francisco area which is different than the rest of data sets.

\begin{figure}[pos=h,width=16cm,align=\centering]
\centering
\includegraphics[width=0.32\textwidth]{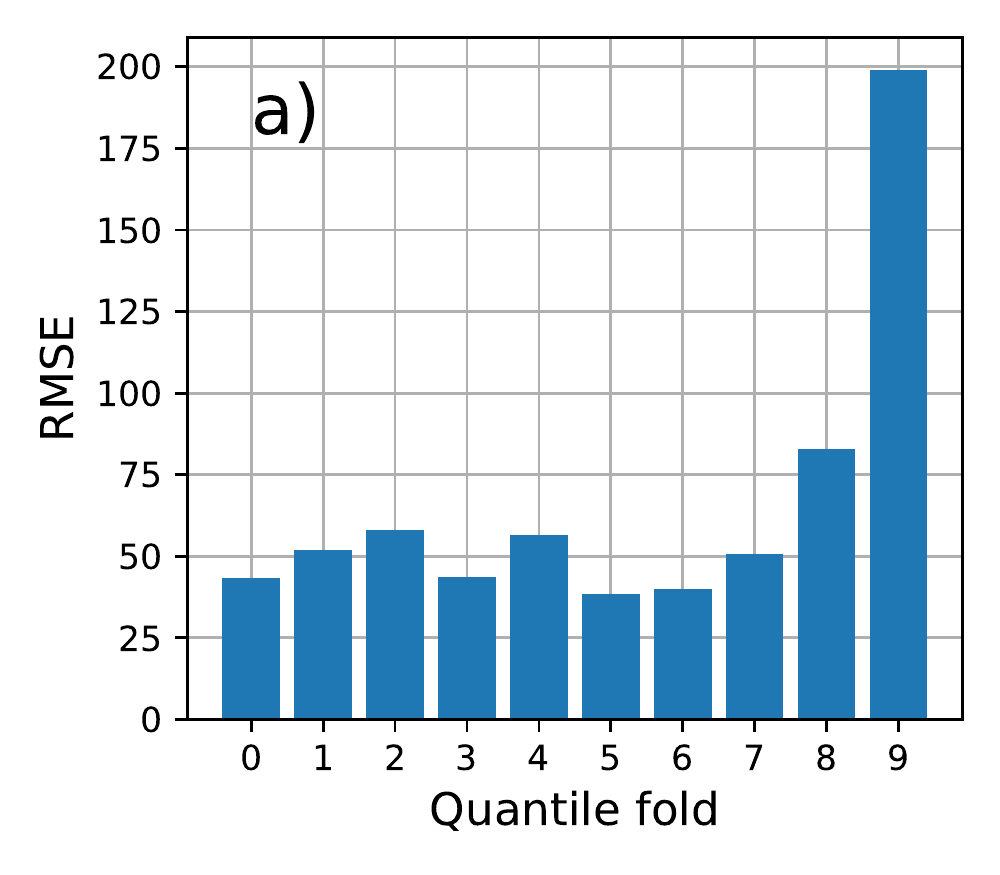}
\includegraphics[width=0.32\textwidth]{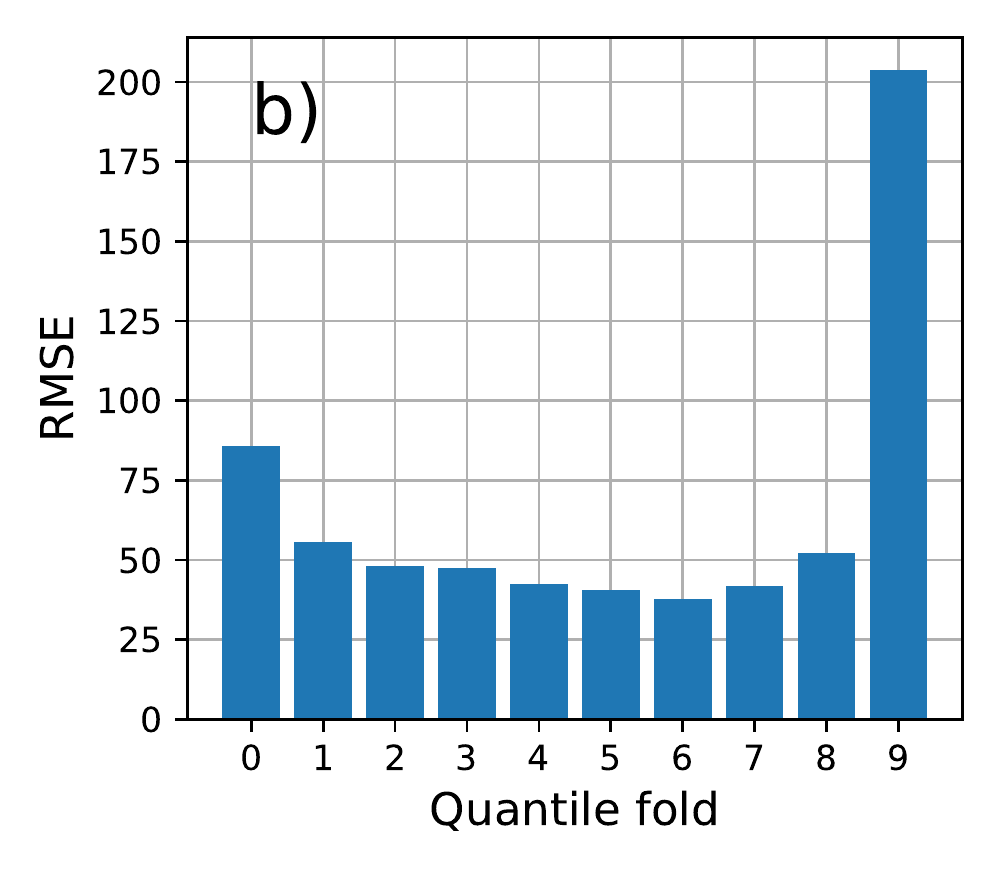}
\includegraphics[width=0.32\textwidth]{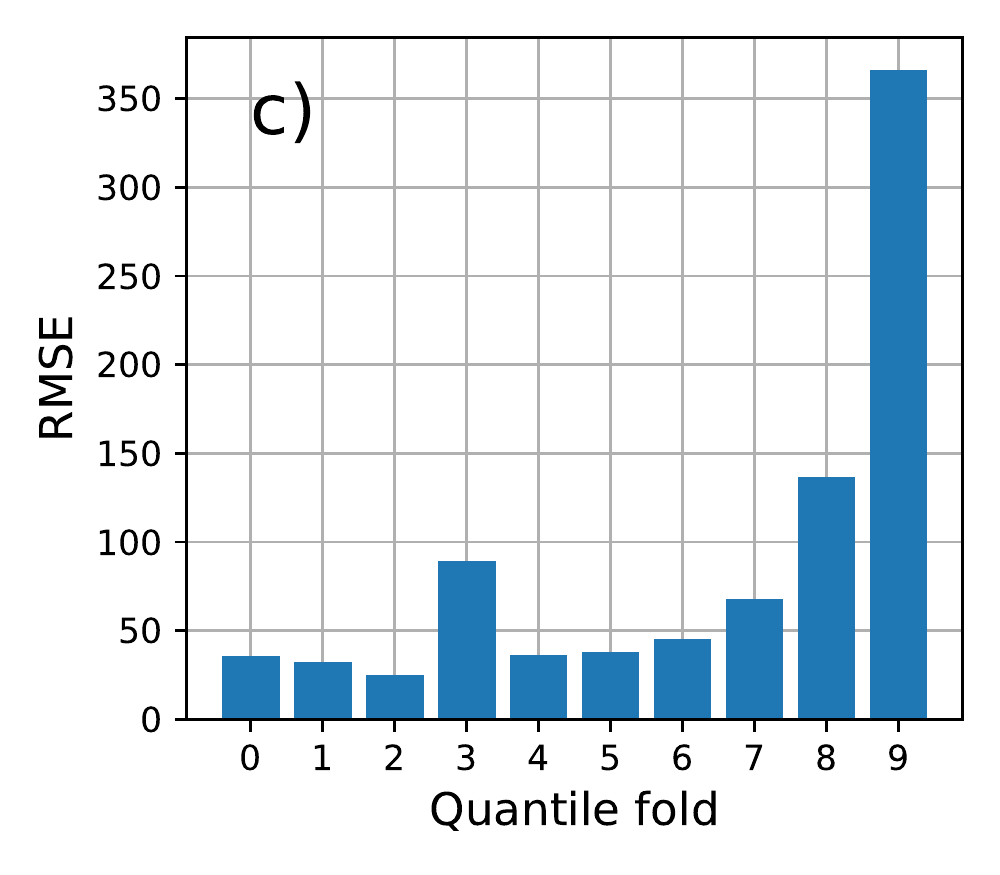}
\caption{Regression using Quantiled Time Folding for a) data set AR b) data set M c) data set SF}
\label{fig:qf}
\end{figure}

\begin{figure}[pos=h,width=16cm,align=\centering]
\centering
\includegraphics[width=0.32\textwidth]{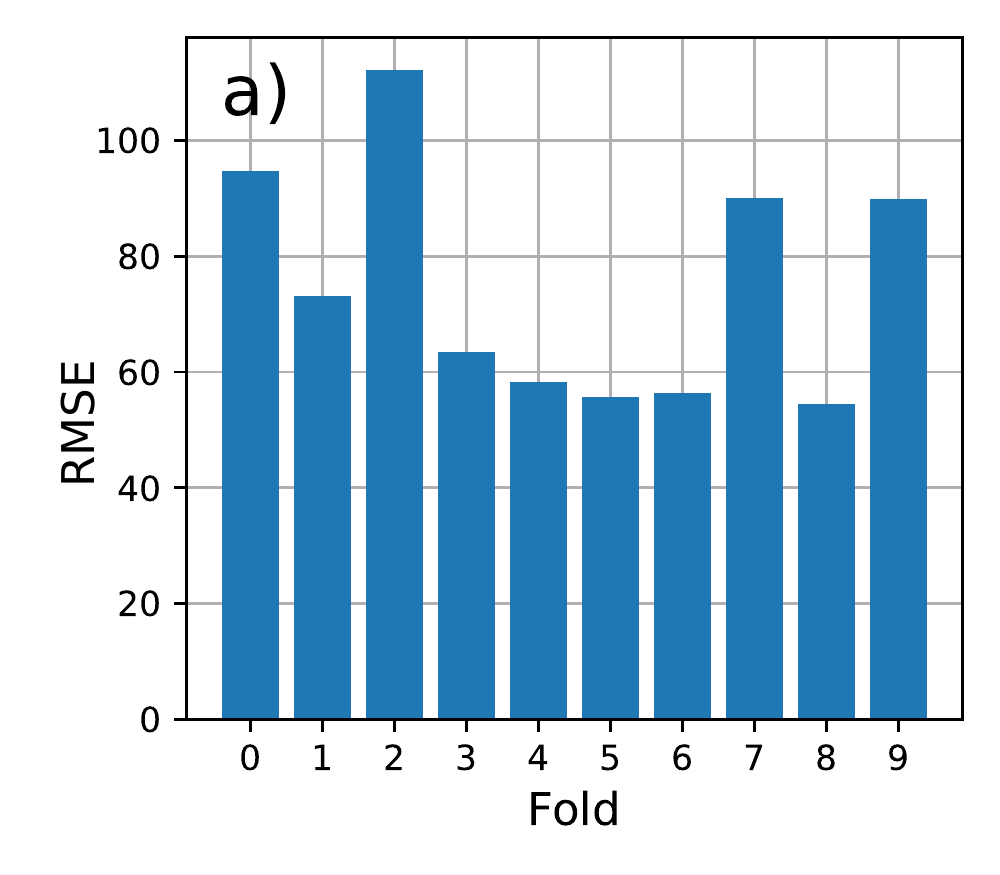}
\includegraphics[width=0.32\textwidth]{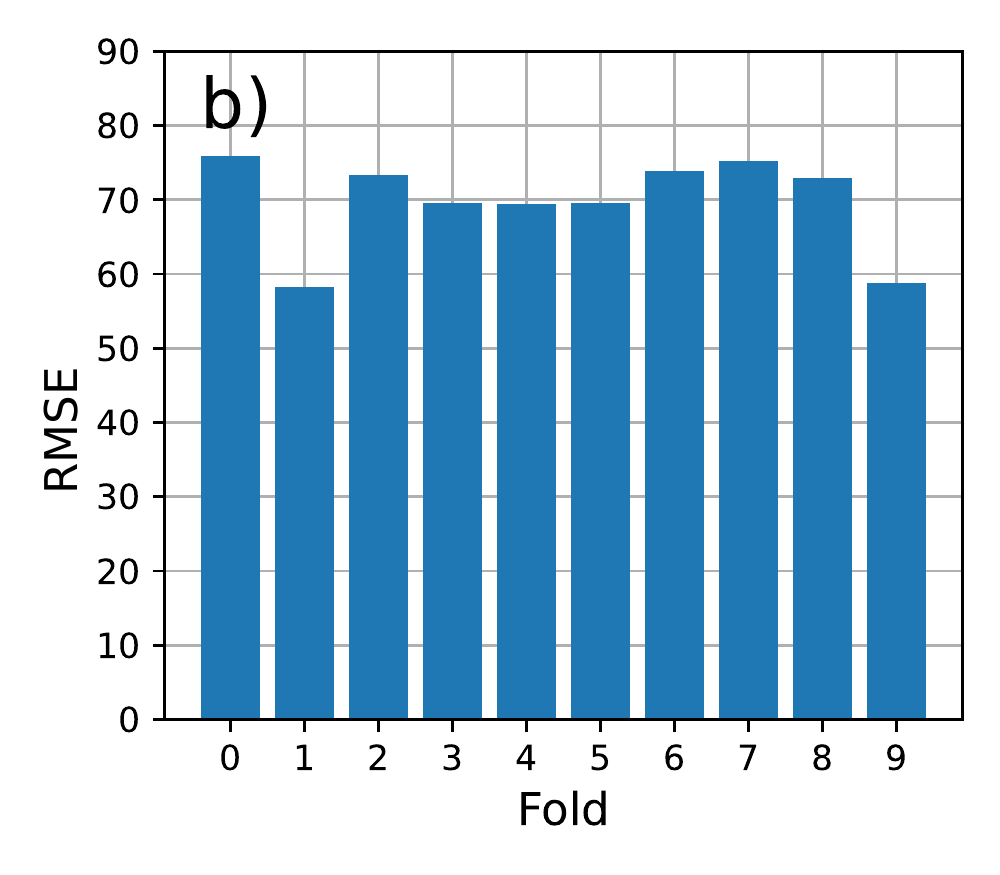}
\includegraphics[width=0.32\textwidth]{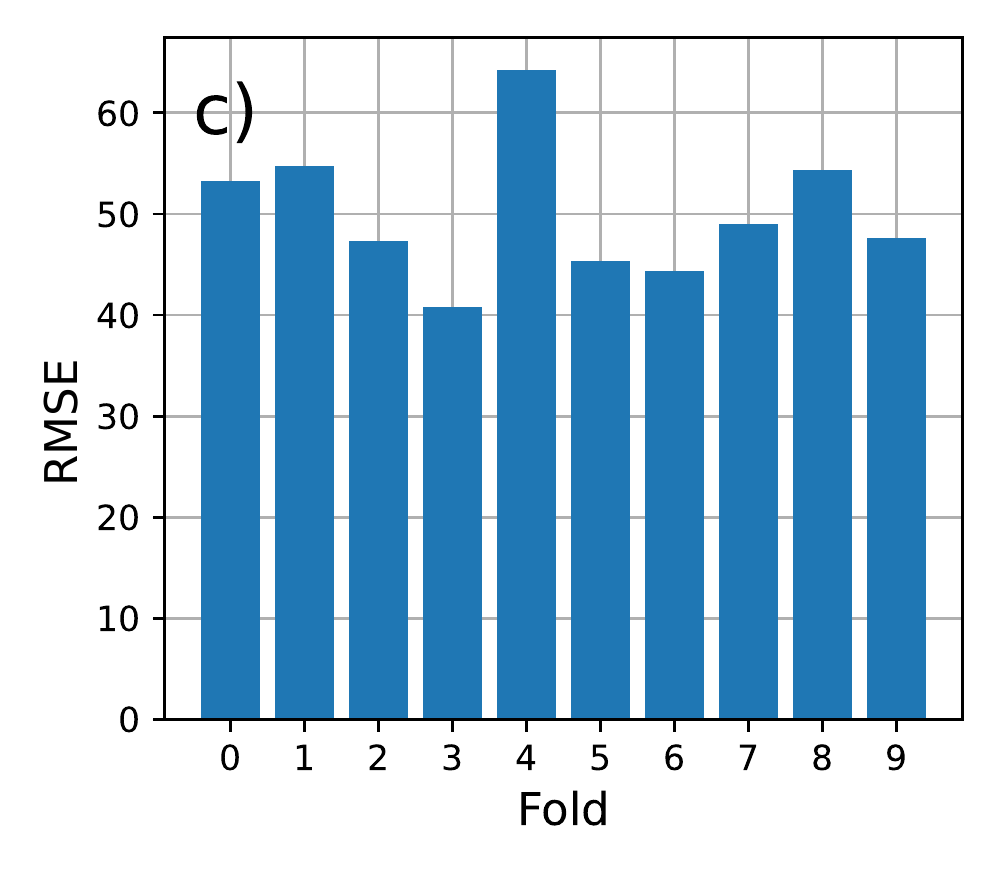}
\caption{Regression using randomised 10-folds for a) data set AR b) data set M c) data set SF}
\label{fig:kf}
\end{figure}

To further see the impact on error by various incident duration groups we introduce the Quantiled Time-Folding and present the results in Figure \ref{fig:qf}. Incident reports are separated into equally-sized duration groups to perform the procedure of cross-validation (each 9 folds evaluated against 1 excluded fold, repeated 10 times). For all three data sets, incidents with the longest duration have the highest contribution to error, even though they represent only 10\% of the data set. Considering this error, we may choose to use the hybrid classification-regression framework, where we perform regression only for intervals with acceptable prediction error. Qunatiled Time-Folding can also be useful to see the contribution to error of every duration group and possible extrapolation error towards incidents with unobserved duration groups. Also, the RMSE metric showcased in Figure \ref{fig:qf} is related to the scale of duration observed in the fold (e.g. high durations can easily translate in high errors), whereas if we adopt a regular 10-fold cross validation (see Figure \ref{fig:kf}), the RMSE error remains below $125.0$ for most of the folds.
}

%% file: sections/5B-Regression-Results.tex
\section{Incident duration prediction using regression: results}

The final objective of the bi-level framework is to predict with an accuracy at the minute level the length of a freshly reported incident, regardless of its previous classification as either short, medium or large. Therefore, the second step of the bi-level prediction framework is to develop more advanced regression models that can adjust to each data set independently and over-perform baseline ML models previously used to solve classification problems. When training such regression models, a significant step is the size of the data set and the distribution of the target variable (incident duration). Due to the long tail distribution of incident duration and the class imbalance problem previously identified, we need to design and construct various regression models capable of learning from various types of data sets to make accurate predictions. However, with limited information (small data set size), the prediction results can be skewed (this effect of prediction skewing will be further discussed). This section first presents the regression results obtained across several scenarios of model training, validation and testing, followed by results of our proposed Intra-Extra Optimisation algorithm applied over all baseline ML models.

\subsection{Regression scenarios results and comparison}

In order to find the best set-up that works for traffic incident prediction in TMCs, we test various regression scenarios (detailed previously in \cref{scenarios}), which show the extrapolation performance for different ML methods. The outlier removal procedures (LDO, HDO) together with the classification thresholds (which separate short-term and long-term duration of incidents) are selected as described in \cref{binary_var_Tc}-\cref{ORM_class}. The primary purpose of this section is to recommend the best scenario set-up for model training and validation when parts of the data set might be hidden. 
\cref{tab:reg1}, \cref{tab:reg2} and \cref{tab:reg3} present the MAPE results for all 7 scenarios (All-to-All, AtoA, AtoB, BtoB, BtoA, AlltoA AlltoB) using all the Baseline ML models across all three data sets (and a dedicated winning regression model across each scenario - last column). Overall, XGBoost seems to be the best regression model in a majority of scenarios across data set AR and M (\cref{tab:reg1},\cref{tab:reg2}): 1) the improvement from using XGBoost shows the lowest MAPE for scenario AtoA of 49.11 and 67.92 correspondingly (predicting short term incidents only using only short term training information), 2) XGBoost also the best performing model for All-to-All regression (59.36\% and 85.98\% MAPE correspondingly). The main difference between LGBM and XGBoost results is that LGBM struggles with extrapolation to lower values as seen in scenario B-to-A for all data sets: 292.68\% vs 77.66\% MAPE for data set A, 663.12\% vs 180.77\% MAPE for data set M, 166.06\% vs 32.62\% MAPE for data set SF for LGBM and XGBoost correspondingly.

In the SF data set, the LGBM is the best performer reaching a MAPE of 9.34\% for the AtoA scenario (which is almost 10 times better than the same scenario for the M data set) and 33.16\% MAPE for All-to-All scenario. This is a significant improvement that reveals what model is adapting to what data set, but most importantly, that each data set reacts differently to the seven scenarios. In the following, we provide a summarised comparison across a selection of few scenarios and their performance.

\begin{table}[pos=h,width=20cm,align=\centering]
	\centering
\begin{tabular}{llllllll}
\toprule
Model &    LGBM &      RF &      LR &    GBDT &     KNN &    XGBoost & Best model \\
\midrule
AlltoAll &   82.76 &  117.28 &  110.99 &  113.41 &  107.79 &  \textbf{59.36} &          XGBoost \\
AtoA     &   60.17 &   59.49 &   59.92 &   62.08 &   58.35 &  \textbf{49.11} &          XGBoost \\
AtoB     &   64.46 &   64.39 &   64.34 &   \textbf{63.82} &   64.68 &  64.39 &          GBDT \\
BtoA     &  292.68 &  381.61 &  367.16 &  348.09 &  349.62 &  \textbf{77.66} &          XGBoost \\
BtoB     &   29.52 &   \textbf{25.03} &   45.14 &   46.26 &   43.82 &  27.55 &          RF \\
AlltoA   &  117.78 &  121.82 &  175.48 &  176.71 &     120 &  \textbf{51.18} &          XGBoost \\
AlltoB   &   34.39 &   37.47 &   32.11 &   \textbf{31.67} &   35.57 &  37.46 &          GBDT \\
\bottomrule
\end{tabular}
\caption{MAPE results for all 7 scenarios on data set AR}
\label{tab:reg1}
\end{table}

\begin{table}
	\centering
\begin{tabular}{llllllll}
\toprule
Model &    LGBM &      RF &      LR &    GBDT &     KNN &     XGB & Best model \\
\midrule
AlltoAll &  135.59 &   226.6 &  229.53 &  229.46 &  229.82 &   \textbf{85.98} &          XGBoost \\
AtoA     &   95.89 &   95.38 &  107.29 &  104.87 &  105.26 &   \textbf{67.92} &          XGBoost \\
AtoB     &   68.78 &   69.01 &   69.49 &   \textbf{68.62} &   69.79 &   68.69 &          GBDT \\
BtoA     &  663.12 &  939.59 &  818.08 &  878.47 &  854.81 &  \textbf{180.77} &          XGBoost \\
BtoB     &   34.14 &   51.02 &   52.33 &   50.99 &   48.68 &   \textbf{31.18} &          XGBoost \\
AlltoA   &  233.48 &  406.43 &  387.25 &  398.13 &  402.02 &   \textbf{76.71} &          XGBoost \\
AlltoB   &   34.38 &   34.34 &   \textbf{34.21} &   34.48 &   36.89 &   34.98 &          LR \\
\bottomrule
\end{tabular}
\caption{MAPE results for all 7 scenarios on data set M}
\label{tab:reg2}
\end{table}

\begin{table}
	\centering
\begin{tabular}{llllllll}
\toprule
Model &    LGBM &      RF &      LR &    GBDT &     KNN &    XGBoost & Best model \\
\midrule
AlltoAll &   \textbf{33.16} &   36.88 &  128.42 &   41.85 &   64.24 &  37.03 &          LGBM \\
AtoA     &    \textbf{9.34} &   11.91 &   16.07 &   12.56 &   14.05 &  11.44 &          LGBM \\
AtoB     &   68.08 &   65.77 &   67.21 &   \textbf{65.53} &   66.26 &  65.84 &          GBDT \\
BtoA     &  166.06 &  191.55 &  389.07 &  211.61 &  302.46 &  \textbf{32.62} &          XGBoost \\
BtoB     &   \textbf{23.69} &   28.76 &   70.18 &   31.08 &    37.6 &  27.61 &          LGBM \\
AlltoA   &   45.35 &   50.74 &  218.49 &   60.03 &   99.06 &  \textbf{35.49} &          XGBoost \\
AlltoB   &   24.28 &   \textbf{23.97} &   45.08 &   25.49 &   30.82 &  24.78 &          RF \\
\bottomrule
\end{tabular}
\caption{MAPE results for all 7 scenarios on data set SF}
\label{tab:reg3}
\end{table}

\textbf{Scenario AtoA} uses short-term traffic accidents (below $T_c$) for both training and the prediction. XGBoost shows a significant performance for AR and M data sets compared with other scenarios; more specifically, they outperform by 10\% all models in data set AR (MAPE=51.2) and 30\% all models in dataset M (MAPE=68.4). For the SF data set, the improvement is even larger (MAPE=12.7), but XGboost loses ground over LGBM, which reaches a MAPE=11.0. The comparison of scenarios AtoA and AlltoA shows that adding incidents with a longer duration can severely affect the prediction performance across all data sets, regardless of the size or location of the incident logs. For the best prediction performance on data sets AR, M and SF, we need to split the data and use separate models for the short-term incidents as predictions become skewed towards longer incident duration. Thus, if we predict short-term incidents using only short-term incidents data logs, we obtain a higher accuracy across all data sets.

\textbf{Scenario AtoB} is unique because regression models are trained on Subset A, which contains short-term incident duration logs while they are trying to predict long-term incidents; therefore, the performance is much worse than for AtoA scenario since incidents with long duration are much scarcer and have unique traffic conditions. BtoB scenario shows lower error than AtoB across all three data sets (e.g. BtoB provides 23.69\% MAPE and AtoB provides 65.53\% MAPE for best models for data set SF). Vice-versa, \textbf{Scenario BtoA} shows very high extrapolation errors across all methods to lower values. Adding short-term incidents into the training set of long-term incidents (when we move from BtoA to AlltoA scenario) significantly reduces the error (76.71\% MAPE for scenario AlltoA, data set M using XGBoost), but it is still significantly higher than for AtoA scenario (67.92\% MAPE for M data set using XGBoost). \textbf{Scenario BtoB} shows better performance (e.g. MAPE=31.18\% for data set M using XGBoost) than using data addition (such as the case of AlltoB, where MAPE=34.21\% using best model) or any extrapolation (as in the case of AtoB, where MAPE=68.62\% using best model). By comparing scenarios AtoB and AlltoB we observe a significant performance improvement when adding data for long-term incidents and predicting subset B (from 63.82\% to 31.67\% MAPE for dataset AR using best model), where error is still higher than for BtoB (25.03\%, AR, best model). \textbf{Scenario BtoA} shows high prediction errors across all scenarios highlighting a bad extrapolation accuracy when predicting short-term incidents duration using long-term traffic incident data. It means that prediction of the duration of short-term incidents should be performed separately from long-term incidents. Thus, we can't use long-term incidents to predict the duration of short-term incidents and vice versa if we are looking at maximising model performance with limited data set; the second reason lies mainly in different traffic behaviour along with severe accidents that can last for several hours which are harder to clear off - these require similar previous events in order to be predicted for their duration.

\verify{\subsubsection{Fusion framework for the incident duration prediction}}
\verify{
In comparison to the above proposed framework, we also present a fusion framework approach, which can be applied when the incident duration category is unknown.
When an incident occurs, the incident duration category is not known, but we have a historical data on traffic incidents which allows us to predict the incident duration category and apply specialised regression models (oriented towards the prediction on subsets of short-term and long-term incidents). 
We further propose two possible approaches to this problem:
\begin{itemize}
    \item \textbf{the pipeline approach} (see \cref{fig:pipe}a ): we train a classification model using a historical data available to predict the incident duration category. Then we predict the incident duration category using the available incident reports. This prediction decides which model we need to use for a further regression (either specialised on short-term or long-term incident duration prediction). The prediction result of the specialised model is then considered to be the final prediction. Specialised regression models are trained on their corresponding subsets. In this case, the decision about the incident duration class is made by the classification model only, which becomes the most important part of the model that is highlighted by significantly improved results (see \cref{tab:reg1,tab:reg2,tab:reg3}).
     \item \textbf{the fusion approach} (see \cref{fig:pipe}b ): instead of relying on the classification model to decide on the incident duration subset, we place a decision-making function on the additional "fusion model", which is the global regression model; it now receives the prediction results from the classification model, the regression models specialised on short-term and long-term incidents (subsets A and B) and from the regression model trained on historical data of traffic incidents regardless of the incident duration group. After training all these models on historical data, we perform the incident duration prediction on this historical data. We then use these predictions (such as the predicted incident class, the incident duration predicted by short-term incident duration regression model, the incident duration predicted by the short-term incident duration regression model, and the incident duration predicted by the regression model) in order to train the global fusion model to make a final prediction of the incident duration; we call this the global fusion model and the predicted duration is a result of multiple models fused in a centralised architecture.
\end{itemize}
}

\verify{

The fusion approach can be perceived as the ensemble model, which allows to solve the computational problem of model training. Ensemble models may perform better than single models due to three main reasons: \citep{Dietterich2000}:
a) statistical: without sufficient data, a model can find multiple hypothesis about the data approximation which has the same accuracy. Each of these hypotheses can lean towards its local optima. By averaging hypotheses, we may find a better approximation of the data;
b) computational: many machine learning models may get stuck in a local optima (e.g. stochastic gradient descent in the case of neural networks or the greedy split finding in the case of decision trees). An ensemble constructed by models performing local search from many different starting points may provide a better prediction performance than the individual models  \citep{Dietterich2000,BALLINGS20157046},
c) representational: each model forms an approximation (representation) of the data, which forms a local representation hypothesis. By combining models it is possible to extend the space of representable functions.

\begin{center}
\vspace{-0.5cm}
\begin{figure}[pos=h,width=16cm,align=\centering]
\centering
\includegraphics[width=0.70\textwidth]{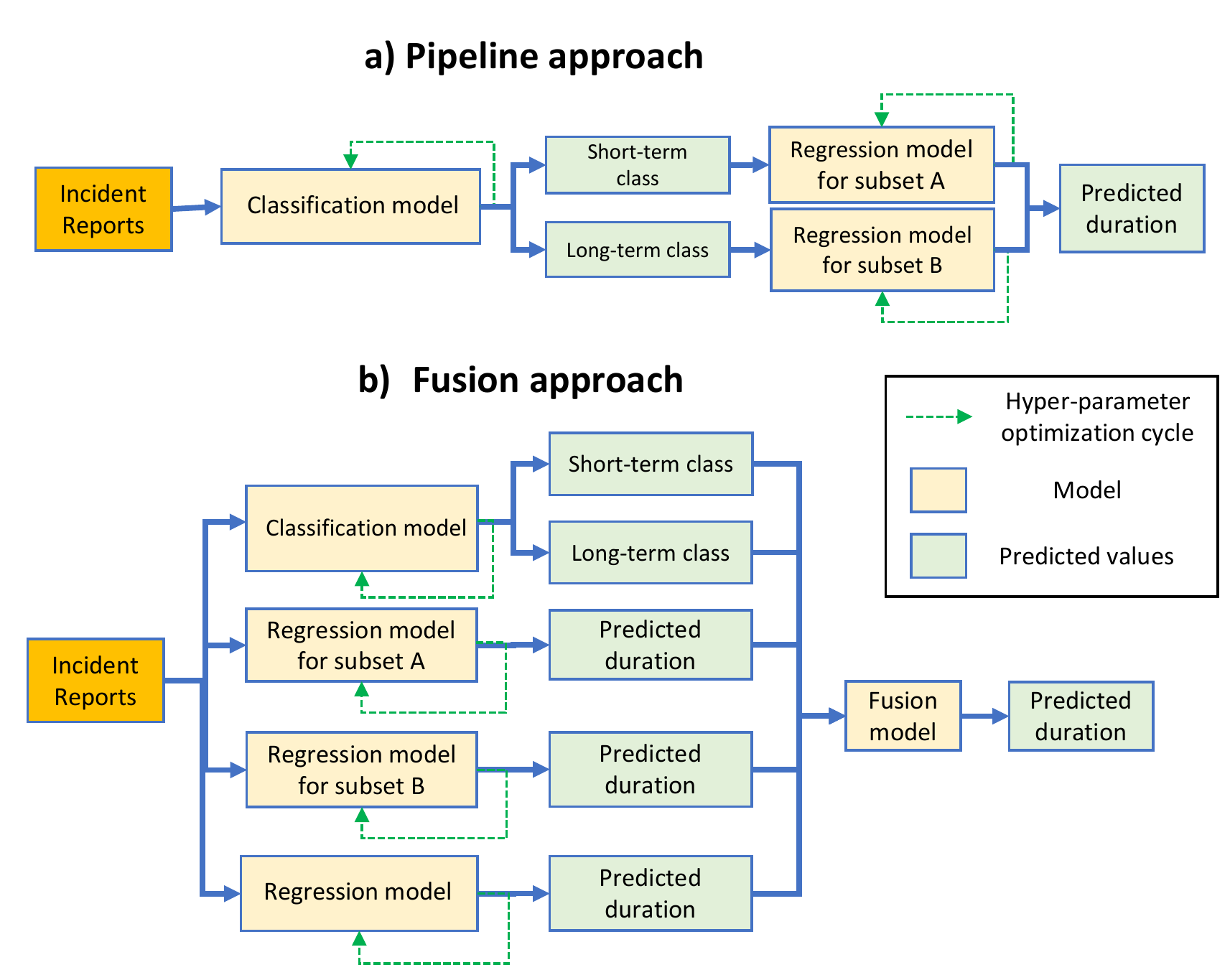}
\caption{Pipeline (a) and fusion (b) approaches for the bi-level framework structure}
\label{fig:pipe}
\end{figure}
\vspace{-1cm}
\end{center}

In the case of a bi-level framework we have statistical, computational and representational reasons to expect a better performance from using an ensemble model rather than a single model, since we use different kinds of models on different subsets (in our case a simple regression model, a classification model, a regression model for subset A, a regression model for subset B). In other words, by splitting the data and by using multiple models we obtain models that are having different local optima (subset A, subset B models) and a different representation of the data (classification and regression models); in this way we can obtain a better prediction performance using model ensemble than using individual models.

Finally, we compare the fusion model, single regression model (e.g. for the data set SF it is the model with the best performance for the task of All-to-All regression) and the pipeline model (where the choice between the regression models depends on the predictions from the classification model) performance on all three data sets in \cref{fig:fusion}. We evaluate all model performance on each fold in a randomised 10 fold cross-validation. We observe that the fusion model performs at least not worse than a single model on all three data sets. We use XGBoost as a fusion model. We also use the corresponding best models for each subset of each data set (see \cref{tab:reg1,tab:reg2,tab:reg3}) with hyper-parameter optimisation (e.g. LightGBM as a single model, performing All-to-All regression task for the data set SF, RandomForest as a best classification model for data set SF according to \cref{fig:rate5}). There is a subtle difference in the average RMSE score among the folds for data set A ( see \cref{fig:fusion}a) where the average RMSE for the fusion model is 59, for the single regression model is 59.8, for the pipeline model is 62.2). The same is for the data set M (see \cref{fig:fusion}b) where 68.9, 68.4 and 70.8 are the average RMSEs for the fusion, single and pipeline models correspondingly). There is a significant improvement in the average RMSE score for data set SF (see \cref{fig:fusion}c) with an improvement from 73.6 to 58 of the average RMSE when using of the fusion model instead of the single model); the pipeline model didn't show any improvement in the model performance. Overall, results show that data availability (the amount of information available about the incidents, which is high for the data set SF) can significantly affect the performance of the fusion model.

Given the observed performance (from a subtle difference to significant improvement) we recommend to use the fusion approach within bi-level framework for the task of the incident duration prediction.

\begin{figure}[pos=h,width=16cm,align=\centering]
\centering
\includegraphics[width=0.32\textwidth]{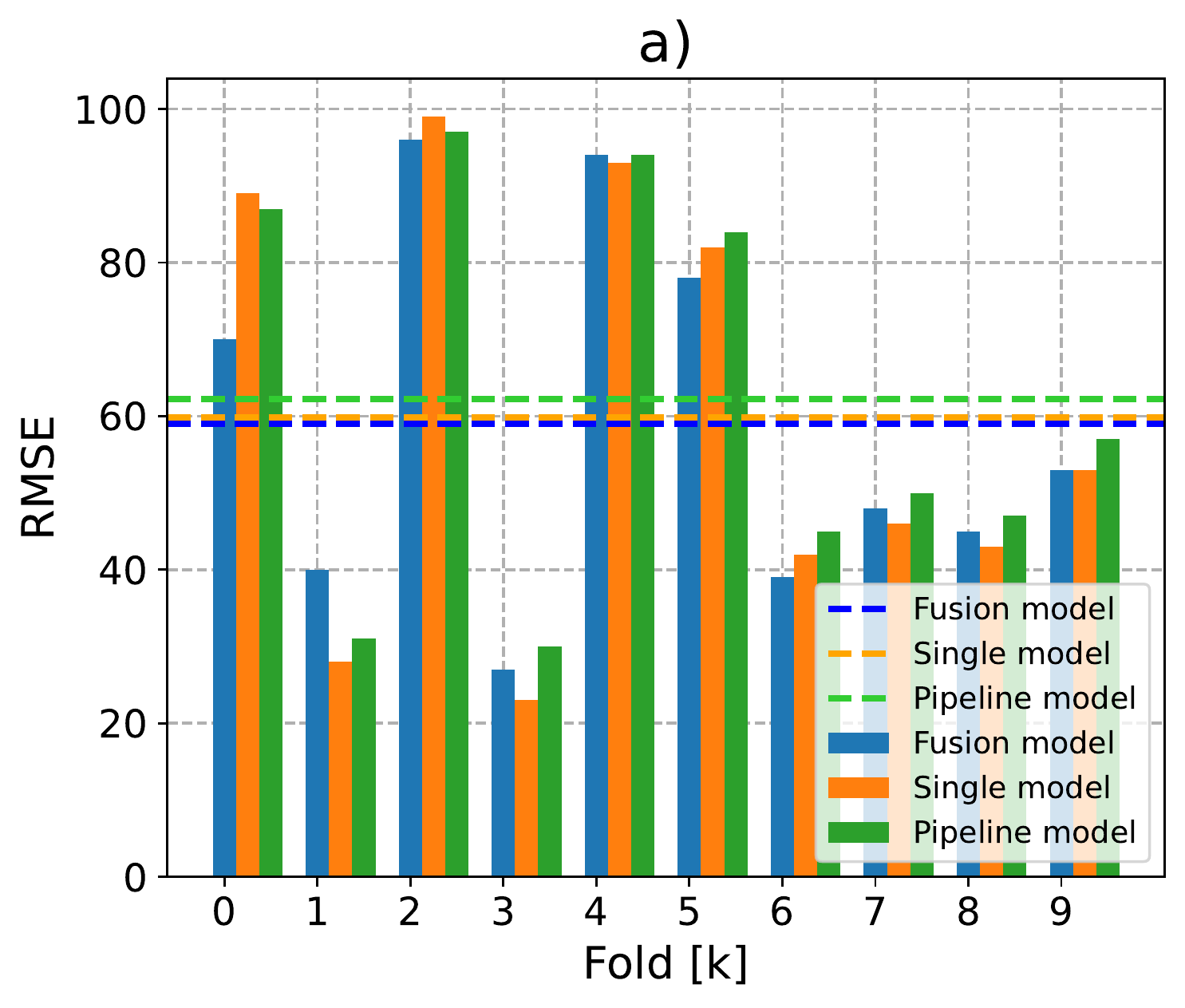}
\includegraphics[width=0.31\textwidth]{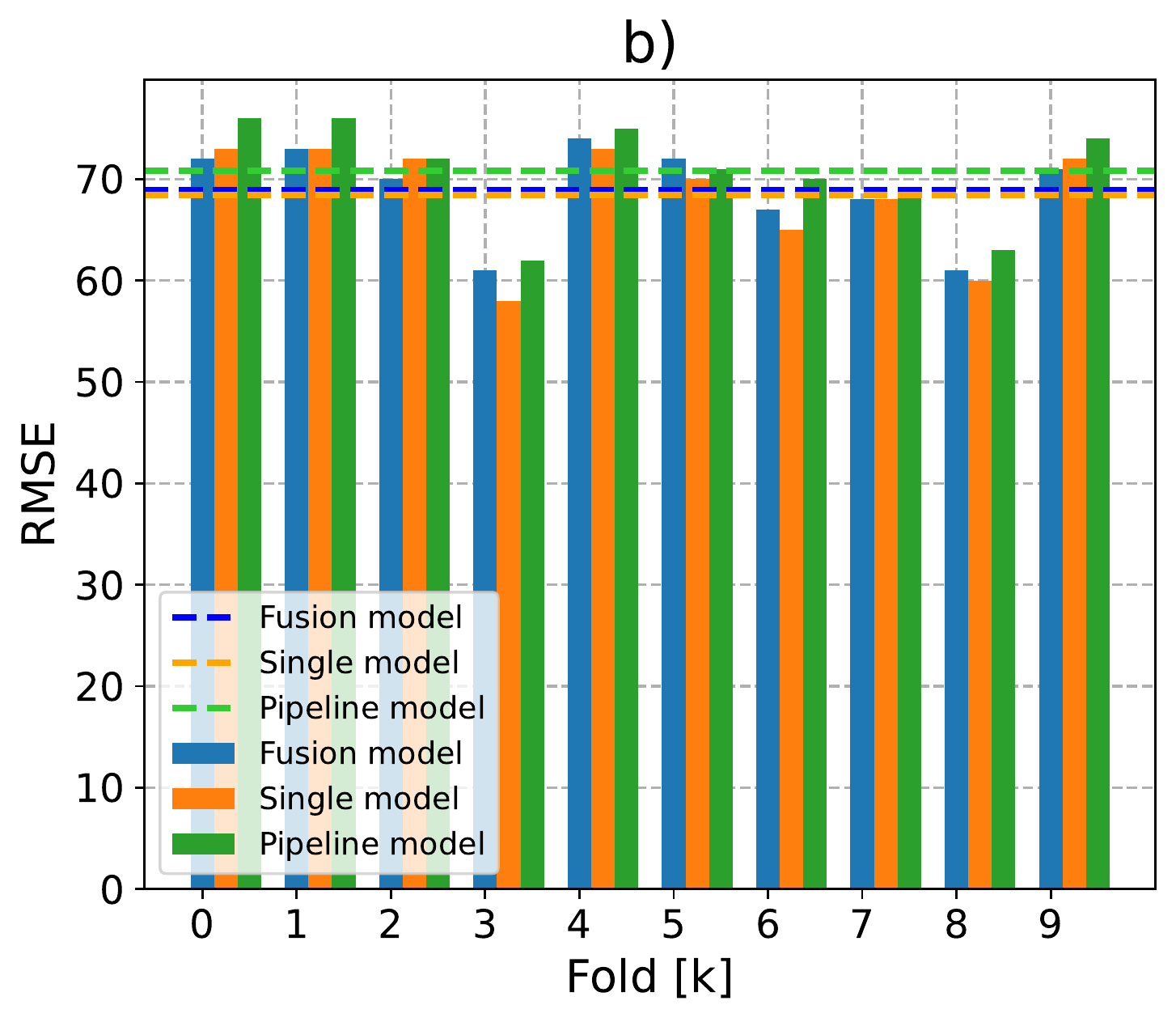}
\includegraphics[width=0.32\textwidth]{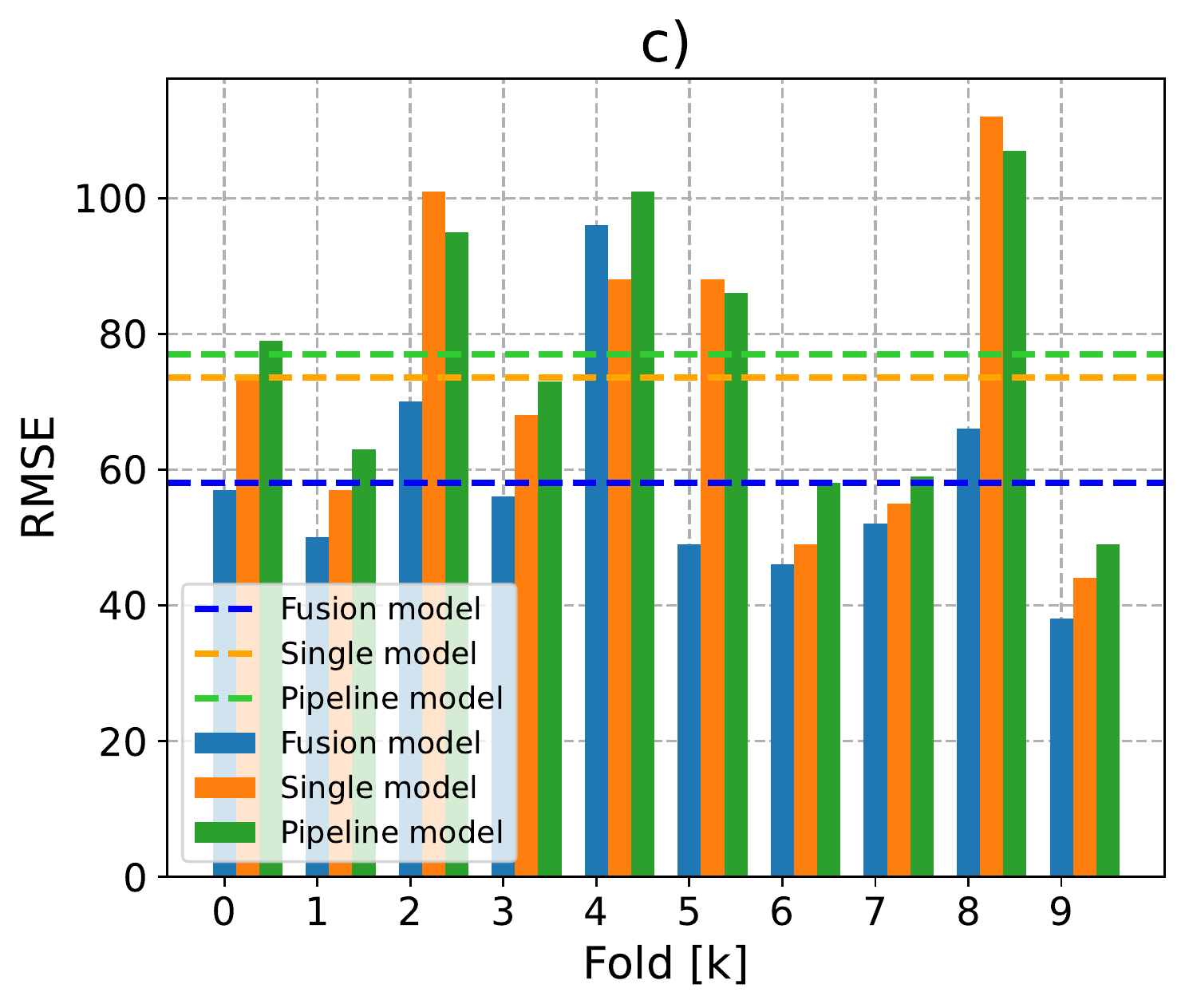}
\caption{Comparison of the fusion and single model performance for a) data set AR b) data set M c) data set SF. Dashed lines represent the average RMSE score across all folds for each corresponding model}
\label{fig:fusion}
\end{figure}
}

\subsection{Outcomes and recommendations}

\revA{From an operational perspective the scenario All-to-All is the ideal situation when traffic management centres would have in their data base both long term and short-term incidents. However, from an operational perspective, several records of short incidents for example and not being kept all the time, while long incidents are often being transferred to various other division if they last more than one day, and they become more of a road infrastructure problem rather than an operational problem which requires constant intervention. }

Scenario modelling shows that the baseline ML models are not improving when facing incident duration extrapolation or data addition (e.g. AlltoA versus AlltoA, BtoB versus AlltoB); these two training set-ups badly affect the model performance extrapolating in any direction. 

\revA{
By evaluating regression scenarios, we highlight the importance that incidents from different duration groups need to be modelled separately in order to significantly improve the accuracy of duration predictions (see (see \cref{tab:reg3}) for SF data set: if we use all available data, to predict the incident duration, we will have $MAPE=33.16\%$ (lower is better), but if we managed to categorise incidents into the short-term group, we could model these incidents with only $9.34\%$ error, which is a significant improvement. Also, the classification may point us to which data we need to include in the modelling because if we use all data to predict the duration of the short-term incident (scenario AlltoA), we will have a much higher error $45.35\%$ MAPE than just using the short-term incident for modelling. From the comparison regression of extrapolation scenarios (e.g. scenarios AlltoA versus AtoA), we see how significant can be the impact of having incidents with long duration in the training set when we need to predict the duration of short term incidents, and therefore ML methods become biased towards long-term incidents, which significantly reduces their performance. If we can perform incident duration regression, then we are able to perform incident duration classification as well. We can do this before performing the regression in each group. In other words, our scenario modelling shows modelling advantages of classifying incidents into duration groups.

}

Therefore, it is essential for the bi-level framework and traffic incident duration prediction to use separate models for short-term and long-term traffic incidents. Moreover, tree-based methods significantly outperforming LR demonstrates that traffic incident regression is a complex non-linear problem that requires more advanced investigations. This aspect was the one that motivated our research to further improve and build a better ML framework for any type of incoming data set, and the results of this novel IEO-ML framework are further detailed in the following section.

%% file: sections/5C-Regression-ORM.tex
\subsection{Regression results for proposed IEO-ML model}\label{IEO_results}

In this section, we employ our proposed Intra-extra joint optimisation approach previously presented in \cref{IEO} and we further present the results of the All-to-All regression scenario, with a log-transformation of incident duration and several outlier removal techniques such as the LocalOutlierFactor (LOF) and the IsolationForest (IF), previously described in \cref{ORM}. All results across the three data sets are presented in \cref{IEO_A}-\cref{IEO_M}-\cref{IEO_SF}.

\begin{small}
\begin{table}[pos=h!,width=16cm,align=\centering]
	\centering
\begin{tabular}{llllllllllll}

$ML_j$	& Log	& Unprocessed &	iIF-Log	 & eIF-Log &	eLOF-Log	& iLOF-Log	& Best approach\\
\toprule
LGBM &	80.4	& 81.1	& 79.9	& 82	& \textbf{78.4} &	80.8	& \textbf{eLOF-Log-LGBM}\\
RF	& 80.3	& 121.9	& 79.5	& 80.7	& \textbf{78.5}	& 79.1	& \textbf{eLOF-Log-RF}\\
LR	& \textbf{80.0}	& 128.4	& 80.4	& 81.6	& 80.5	& 80.5 &	Log-LR\\
GBDT	& \textbf{79.4} &	128.2 &	82.0	& 81.3 &	81.4	& 83.4	& \textbf{Log-GBDT}\\
KNN	& 82.9	& 127.4	& 82.3	& 86.2	& 81.7	& \textbf{81.3} &	iLOF-Log-kNN\\
XGBoost	& \textbf{59.4}	& 61.1	& 60.8	& 59.8	& 60.9 &	59.9	& \textbf{Log-XGboost}\\
\midrule
Best $ML_j$ &	XGBoost &	XGBoost &	XGBoost &	XGBoost	& XGBoost &	XGBoost & \\	

\bottomrule
\end{tabular} 
\caption{MAPE results for All-to-All scenario of data set A, using different ORM approaches and incident duration transformation, via the proposed IEO-ML approach.}
\label{IEO_A} 
\end{table}
\end{small}

For the data set A (\cref{IEO_A}), we observe a significant impact of using the log-transformation of the incident duration vector via the resulting MAPE (see Unprocessed versus Log columns). Since the log-transformation provides a significant improvement among majority of ML models, we decide to use it in our outlier removal scenarios. When comparing results across all models, both regular and re-enforced by our IEO approach (column comparison - see Best $ML_j$ results), we observe that XGBoost is the best performing baseline model for this data set reaching a 59.4 MAPE. Furthermore, when comparing results across regular ML models versus our proposed IEO-ML enhancements (row comparison), then the extra optimisation approaches seem to outperform the intra optimisation approaches (see iIF-Log versus eIF-Log and eLOF-Log versus iLOF-Log columns). The last column indicates the best approach that won across all proposed IEO approaches where for example, eLOF-Log-RF model is read as the extra optimisation method applied together with the Local Outlier Factor and Random Forest over the log scale data transformation; for this data set A results indicate a similar performance between using baseline ML models with log transformation versus enhanced IEO-ML - for example the joint optimisation provides an improvement (eLOF-log-LightBGM, eLOF-log-RF) versus the cases cases when only the baseline ML with the log-transformation was used (e.g. Log-LR, Log-BDT). However, the A data set is very small and has a special behaviour when compared to the others as further results revealed.

\begin{small}
\begin{table}[pos=h!,width=16cm,align=\centering]
	\centering
\begin{tabular}{llllllllllll}
\toprule
$ML_j$	& Log	& Unprocessed &	iIF-Log	 & eIF-Log &	eLOF-Log	& iLOF-Log	& Best approach\\
\midrule
LGBM	& 124.6	& 138.0	& \textbf{123.6}	& 126.8	& 125.1	& 124.1	& \textbf{iIF-Log-LGBM}\\
RF      	& 126.3	& 238.6	& 126.6	& \textbf{125.7}	& 127.1	& 126.6	& \textbf{eIF-Log-RF}\\
LR	        & 130.7	& 245.9	& \textbf{129.8}	& 129.9	& 131.1	& 131 &	\textbf{iIF-Log-LR}\\
GBDT	    & \textbf{126.7}	& 240.1	& 126.9	& 126.7	& 127.2 & 126.9	& \textbf{Log-GBDT}\\
KNN	        & 139	& 248.2	& \textbf{135.1}	& 137	& 139.4	& 138.2	& \textbf{iIF-Log-KNN}\\
XGBoost	    & 78.6	& 113.2	& \textbf{77.5}	& 80.6	& 78.3  & 79.6	& \textbf{iIF-Log-XGBoost}\\
\bottomrule
Best $ML_j$	& XGBoost	& XGBoost	& XGBoost &	XGBoost	& XGBoost	& XGBoost\\	
\end{tabular}
\caption{MAPE results for All-to-All scenario of data set M, using different ORM approaches and incident duration transformation, via the proposed IEO-ML approach.}
\label{IEO_M}
\end{table}
\end{small}

For the data set M (\cref{IEO_M}), when we use Log-transformation, we observe very high MAPE scores (100\% and higher), except for XGBoost, which provides a MAPE of 78.6\%. When comparing the models with each other against the IEO enhancements as well (column comparison), using XGboost as a baseline seems to over-perform all the other approaches, with the best results being a MAPE=77.5 for iIF-Log-XGBoost. When comparing against the proposed approaches (row comparison), the Intra joint optimisation using Isolation Forest in log-transform shows the best performance on this data set for four models (iIF-Log-LGBM, iIF-Log-LR, iIF-Log-kNN, iIF-Log-XGBoost), which can be attributed to data set data structure - outliers can be better analysed using tree-based outlier removal methods rather than distance-based LOF. For the majority of models (4 out of 6), our proposed joint optimisation algorithm obtains the best results for this data set.

\begin{small}
\begin{table}[pos=h!,width=16cm,align=\centering]
	\centering
\begin{tabular}{llllllllllll}
\toprule
$ML_j$	& Log	& Unprocessed &	iIF-Log	 & eIF-Log &	eLOF-Log	& iLOF-Log	& Best approach\\
\midrule
LGBM	& 29.9	& 32.6	& 29.7	& \textbf{29.5}	& 30.2 &	29.9	& \textbf{eIF-Log-LGBM}\\
RF	& 28.9	& 38.7	& \textbf{28.7}	& 28.9	& 28.8	& 28.9 &	\textbf{iIF-Log-RF}\\
LR	& 72.6	& 140.5	& 72.8	& 73.1	& 73.3	& \textbf{72.4} &	\textbf{iLOF-Log-LR}\\
GBDT	& \textbf{31.2}	& 46.3	& 31.5	& 31.4	& 32.4 &	32.2	& \textbf{Log-GBDT}\\
KNN	& \textbf{61.5}	& 108.6	& 61.7	& 62.5	& 62.2	& 61.8 &	\textbf{Log-KNN}\\
XGBoost	& 31.7	& 35.1	& 31.9	& 31.6	& 32.7 &	\textbf{31.0}	& \textbf{iLOF-Log-XGBoost}\\
\bottomrule
Best $ML_j$	& RF	& LGBM	& RF	& RF	& RF	& RF\\	
\end{tabular}
\caption{MAPE results for All-to-All scenario of data set SF, using different approaches for ORM and incident duration transformation, via the proposed IEO-ML approach.}
\label{IEO_SF}
\end{table}
\end{small}

For the data set SF (\cref{IEO_SF}), we observe two competing models - LGBM and Random Forests with a prevalence for Random Forests (column comparison - see Best $ML_j$ results). Also, we observe a considerably lower MAPE score for the best performing models which reached the lowest threshold of 28.7 across all the data sets used in this study. This reveals the power of more complete and larger data sets which can significantly improve the model performance. When comparing the IEO approaches (row comparison), the intra joint optimisation shows improvement across three models and more specifically for the best performing model on this data set, RF. One consistent finding across all results is the fact that the log-transformation of the incident duration vector should be used at all times for incident duration prediction since it significantly improves predictions accuracy; this is mostly related to the long tail distribution and extreme outliers which can affect the final errors in the model performance evaluation. Overall, the best performing models are considered to be XGBoost and Random Forests.

\textbf{To summarise}, every data set has its specifics in the data structure, which make some models and outlier removal methods performing better than others. Thus, it is necessary to deploy different models and outlier removal approaches on every data set. Conventional models (KNN and Linear Regressions) show the highest error which is almsot twice in comparison to tree-based models. Thus, tree-based models are preferred options for solving the incident duration prediction together with adapted optimisation and outlier techniques. Overall, we proved that our proposed intra joint optimisation is improving the regression results across multiple data sets (especially data sets M and SF in 7 out of 12 cases). The joint optimisation of the model together with the outlier removal method shows a significant improvement in majority of cases (12 out of 18) across all three data sets.

\subsection{Bi-level framework implementation}

The code for the bi-level framework exploring previously described scenarios can be found by the link:\\ \url{https://github.com/Future-Mobility-Lab/bi-level-framework}

%% file: sections/6-Feature-importance.tex
\section{Feature importance impact and evaluation}\label{S6-Feature-importance}

\begin{figure}
\centering
\includegraphics[width=0.48\textwidth]{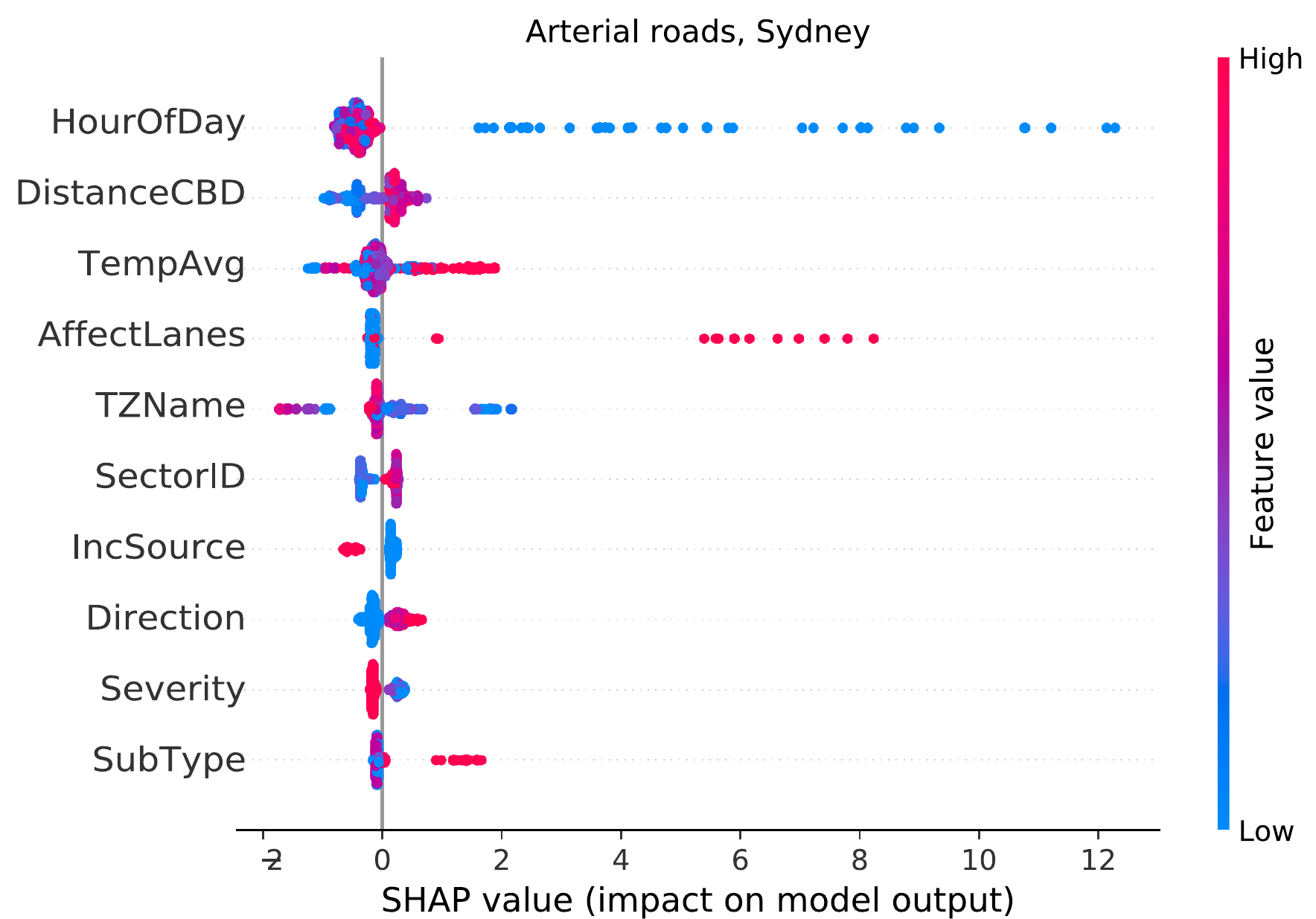}
\includegraphics[width=0.48\textwidth]{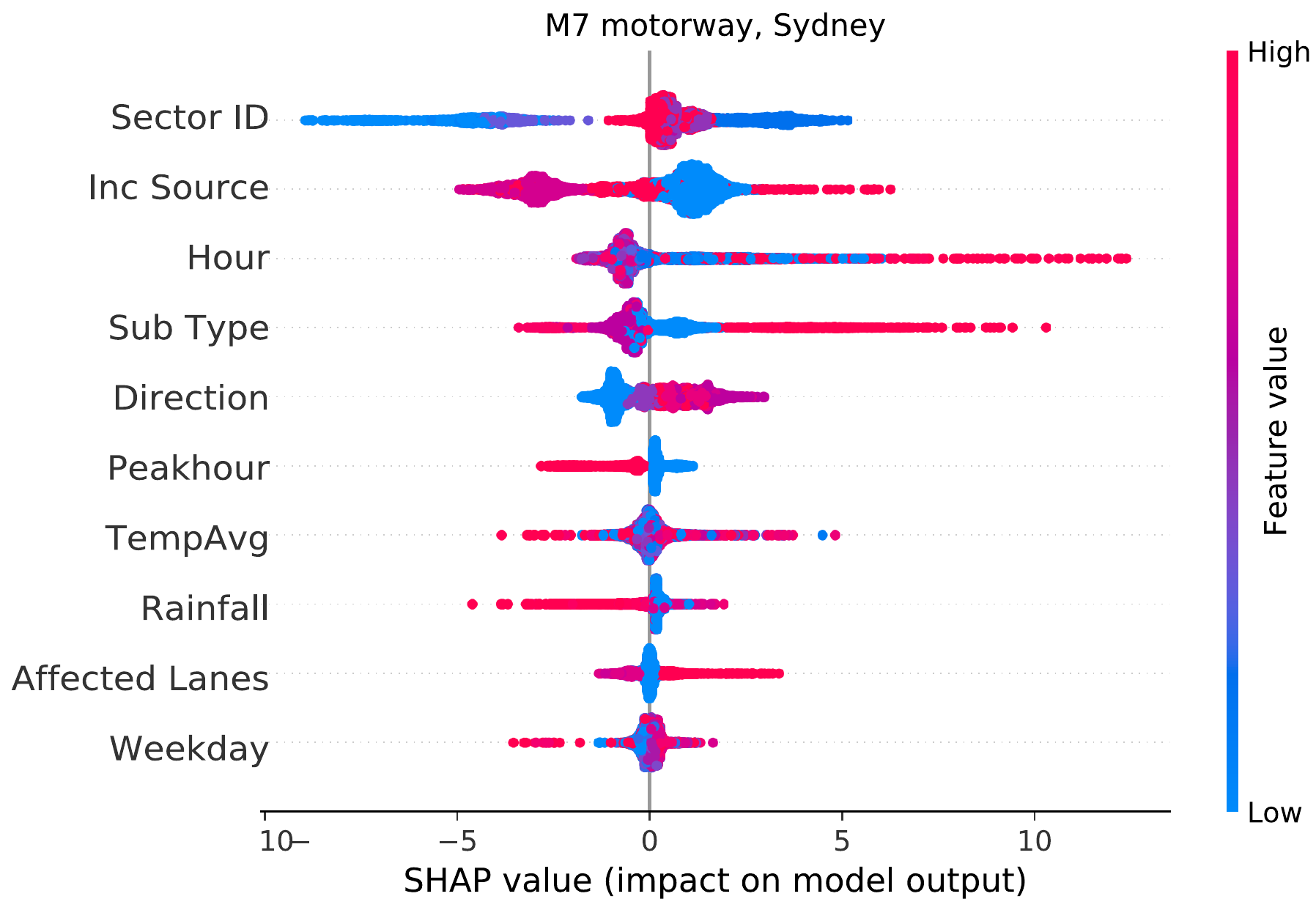}\\
\includegraphics[width=0.48\textwidth]{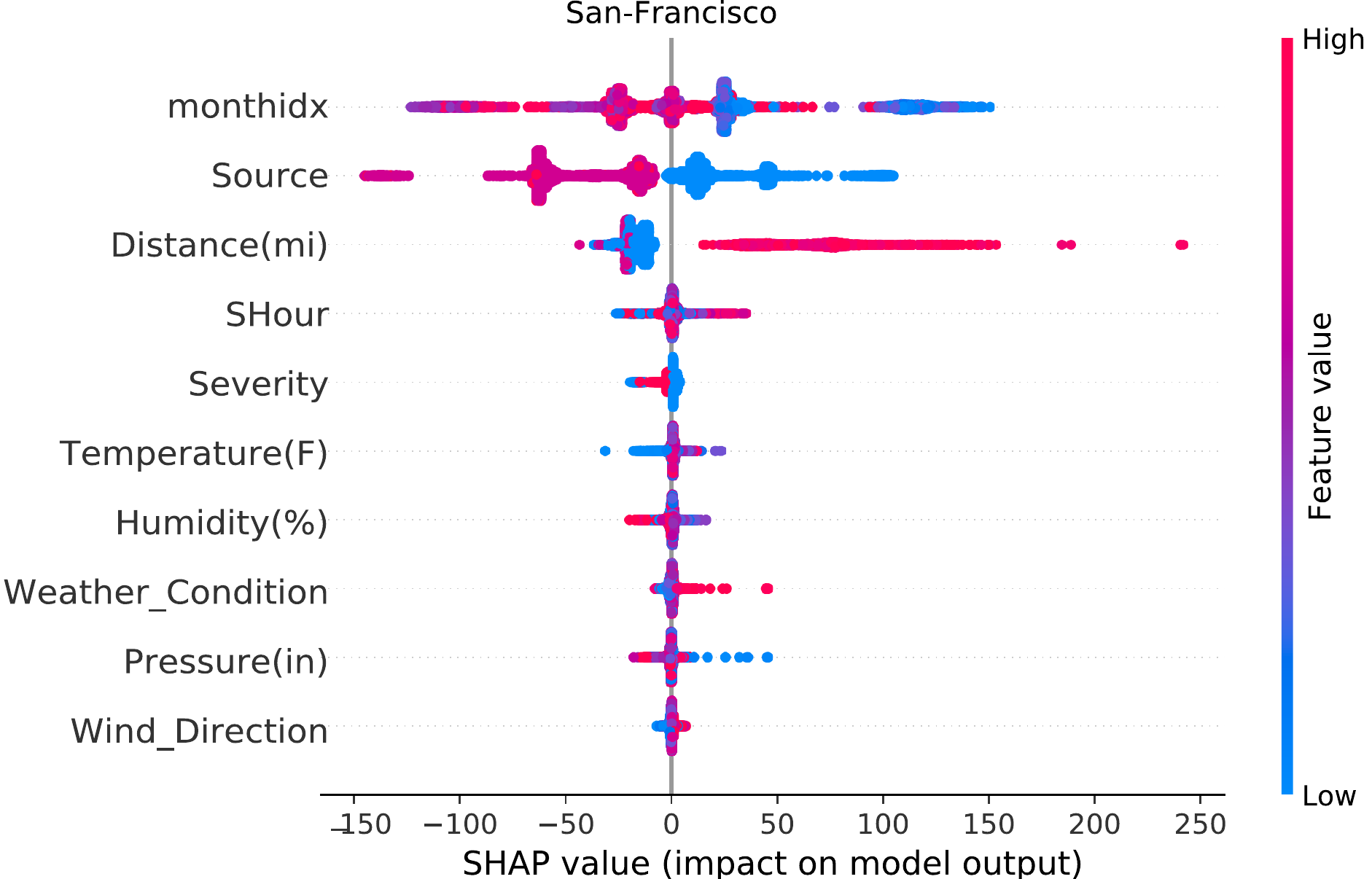}
\caption{Feature importance for All-to-All regression using XGBoost for a) Arterial roads, Sydney, Australia b) M7 motorway, Sydney, Australia c) San-Francisco, USA}
\label{fig:shap}
\end{figure}

Finally, we evaluate the feature importance using a Shapley value calculation in order to estimate the contribution of each feature to the final prediction score. Each point related to a feature is shown in \cref{fig:shap} and represents the SHAP value score (Oy-axis), coloured by its value (from low to high),while the Ox-axis shows the impact of that feature information on the entire prediction output. The used models for this feature importance analysis are the winning models of each data set (A, M, or SF) as previously discussed. 

The hour-of-the-day when the incident started is among the top 5 features sorted by importance (ranked on the $1^{st}$ place for data set A, $3^{rd}$ for M and $4^{th}$ for SF). For example, \cref{fig:shap}a) showcases that as the hour of the day increases (getting closer to midnight) the traffic durations are lower as the congestion is lower and rescue teams arrive faster to the accident location; this is the opposite on the motorways as \cref{fig:shap}b) reflects that rescue teams havea a harder time reaching the incident location in the evening, which is mostly explained by the high distance of the motorway from the local incident management centre. The incident reporting source also has a high significance ( ranked as $7^{th}$ most important for A, $2^{nd}$ for M, $2^{nd}$ for SF). The Ox-axis on SHAP plots represents the impact on model output (e.g. the effect on the predicted duration value). Even though the average temperature is considered significant, its effect on the regression model output is very small $\left[-5min;+5min\right]$ for data set AR, $\left[-5min;+5min\right]$ for data set M, $\left[-25min;+25min\right]$ for data set SF. The distance from CBD (DistanceCBD) is important in the data set A, as it can point at some problematic areas, therefore causing a higher incident duration. The number of affected lanes is also an important feature for incident duration prediction on arterial roads in Sydney. The model outputs for the M7 motorway revealed that is highly dependent on the sector ID (similar to the traffic zones in the data set A), which may be linked to the nature of the location or to the distance from incident management agencies. The average daily temperature also affects predictions ($3^{rd}$ place in A, $7^{th}$ in M and $6^{th}$ in SF). Weather factors (rainfall) are found to play a significant role in the M and SF data sets (humidity and barometric pressure may be predictors of rainfall). Different incident sub-types in the M data set (e.g. car, motorcycle, truck, multi-vehicle) contribute to the difference in the accident duration. Severity is weakly connected to the incident duration in the A and SF data sets. It is important to note that the SF data set contains 49 features, but 39 are of very low importance for the incident duration prediction. The length of the affected road segment (Distance in SF) may also be an essential feature which is not found in Sydney data sets. Overall, the specificity of each data set is reflected once again not only in the models that may be more successful than others but also in the way that the same model can provide various feature importance due to each country, their unique landscape and different way of dealing with the disruptions. 

\revA{
\subsection{Short-term vs long-term incident duration prediction feature importance}

We further perform a comparison of feature importance for the duration prediction of short-term vs long-term traffic incidents, across all data sets. 

\subsubsection{Arterial Roads Feature Importance, Sydney Australia.}

\cref{fig:impA} showcases the Feature importance for All-to-All regression using XGBoost for a) short-term incidents b) long-term incidents of Arterial roads, Sydney, Australia. When analysing long-term incidents, one important observation is the direct influence of the number of affected lanes on the severity and duration of disruptions. However, this feature is found to have low importance for short-term incidents. The farther short-term incidents happen from the CBD, the longer it takes to clear them off. The location of the incident is extremely important for both long and short term incidents, most likely due to the easiness to reach the affected location by the intervention teams. Another important factor affecting the short term incidents in Sydney seems to be the travel patterns for commuting [month of the year, day of week, sectionID, section capacity]. Also,  the DayOfWeek (value ranges from 0 to 6), we see that the higher the value (closer to the end of the week), the longer it takes for the incident to clear. Also, some sectors reflected by the SectionID feature demonstrate a lower incident duration, which may highlight that some specific areas of the city are less affected by traffic incidents.

\begin{figure}
\centering
\includegraphics[width=0.48\textwidth]{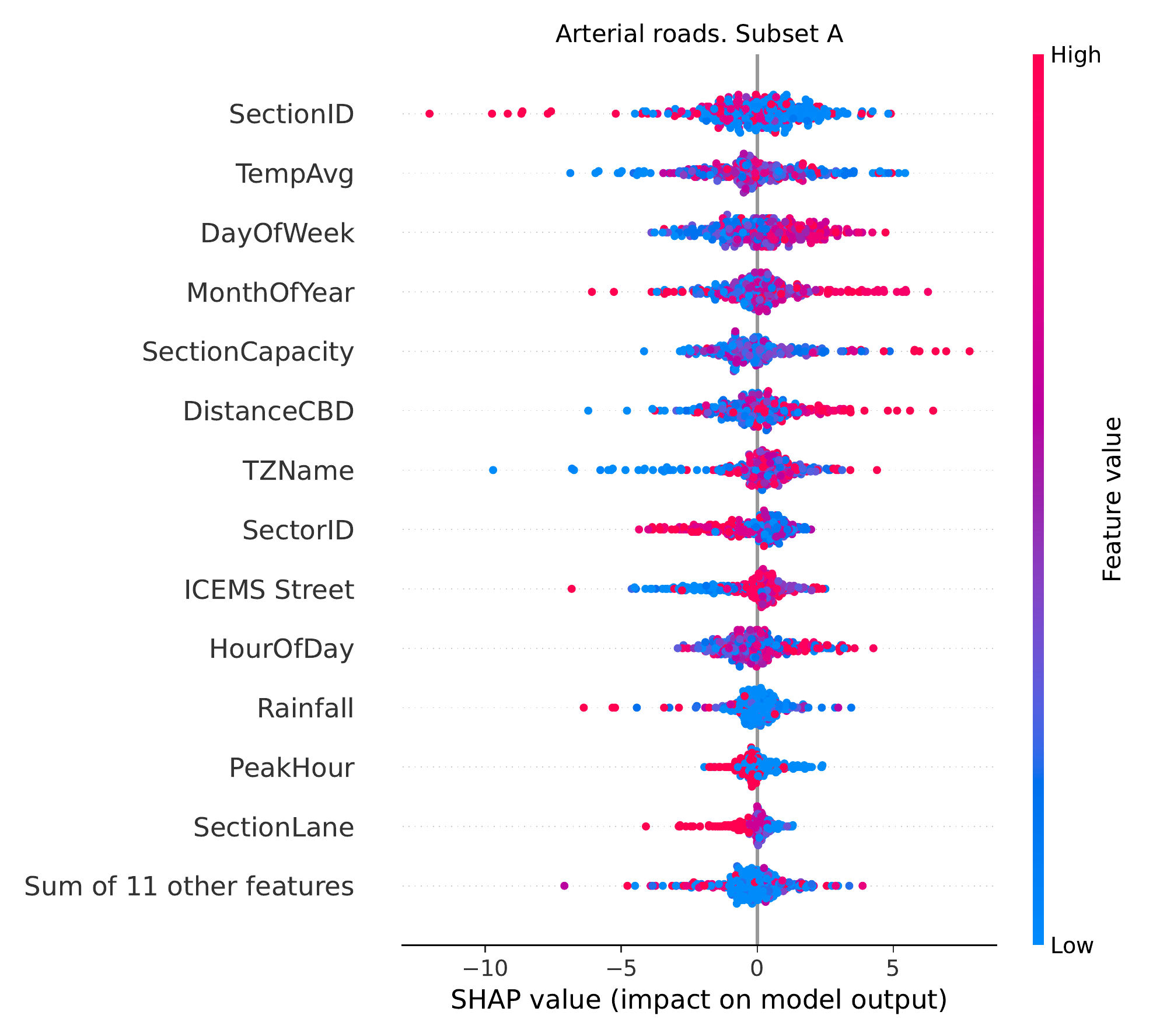}
\includegraphics[width=0.48\textwidth]{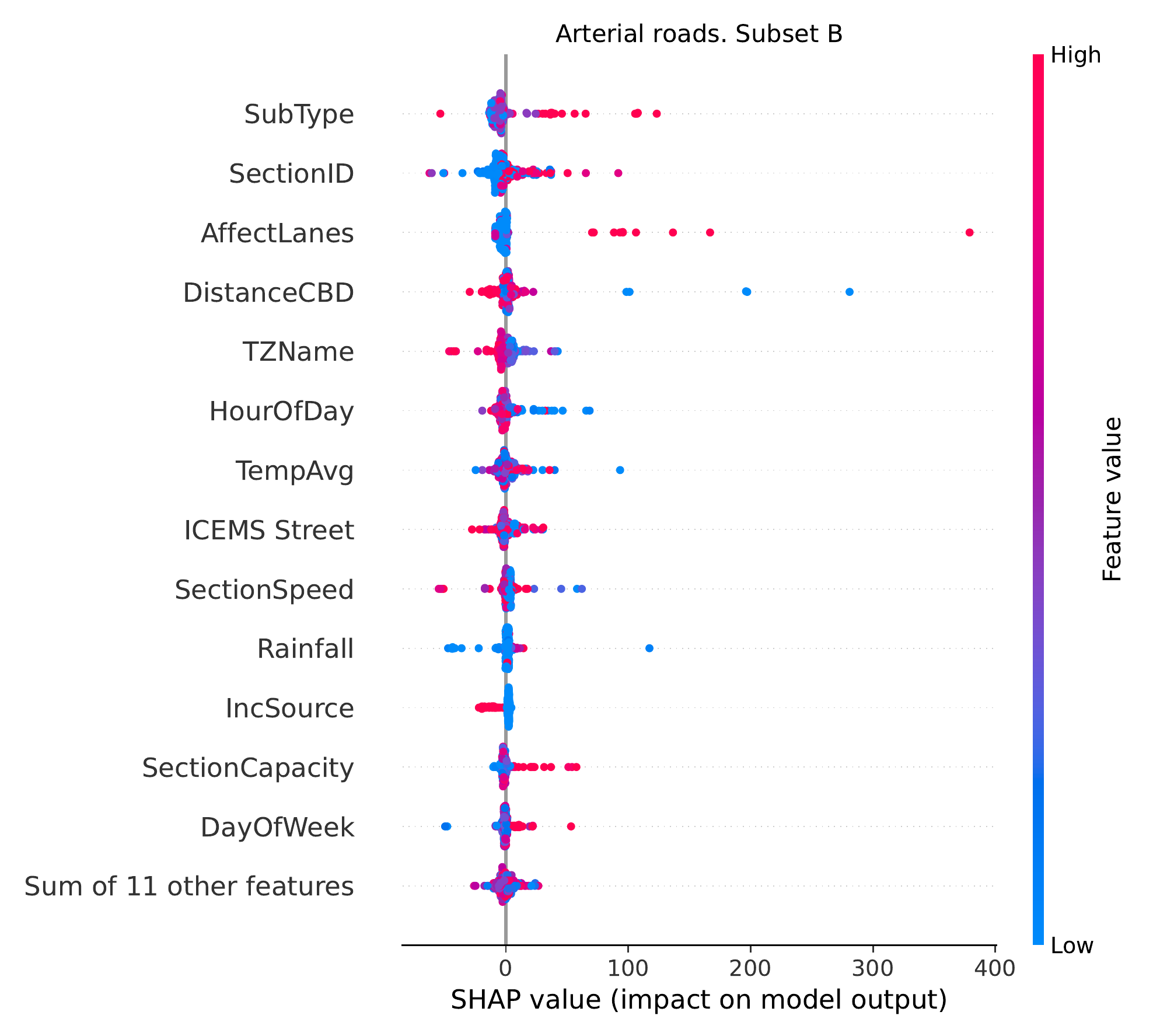}
\caption{Feature importance for All-to-All regression using XGBoost for a) short-term incidents b) long-term incidents of Arterial roads, Sydney, Australia}
\label{fig:impA}
\end{figure}

\subsubsection{Motorway Feature Importance, Sydney Australia.}

\begin{figure}
\centering
\includegraphics[width=0.48\textwidth]{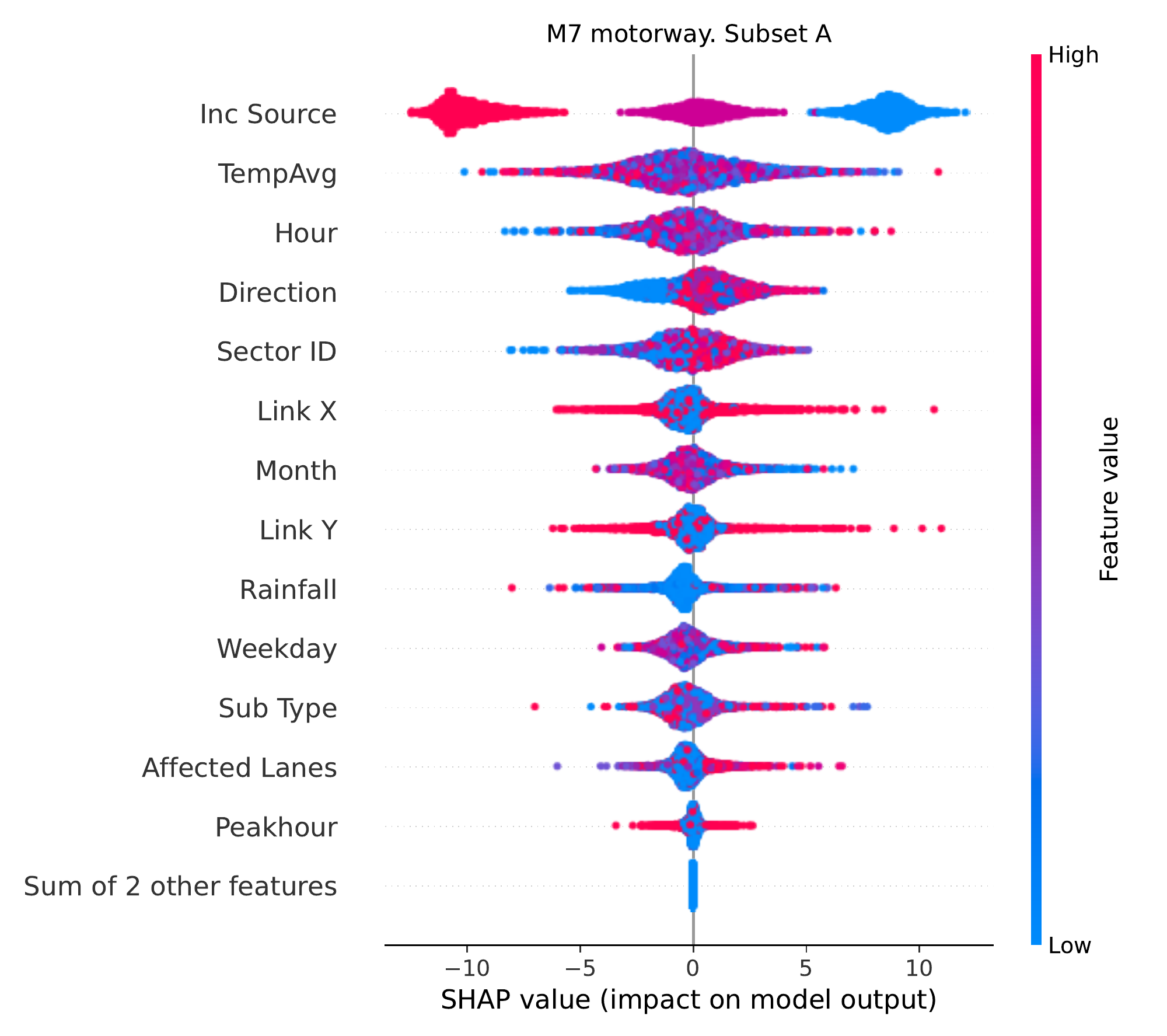}
\includegraphics[width=0.48\textwidth]{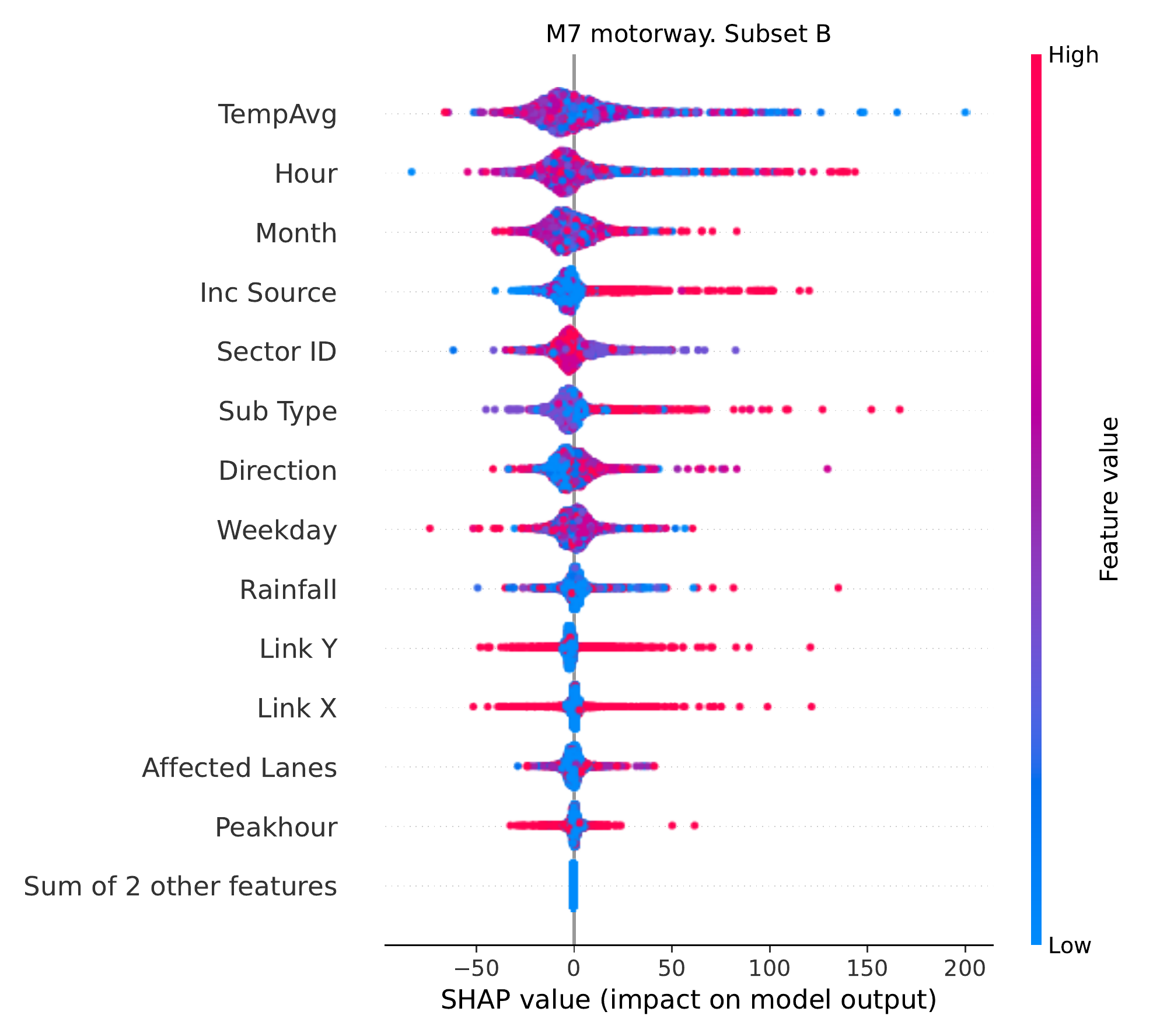}
\caption{Feature importance for All-to-All regression using XGBoost for a) short-term incidents b) long-term incidents of M7 Motorway, Sydney, Australia}
\label{fig:impM}
\end{figure}

\cref{fig:impM}showcases the Feature importance for All-to-All regression using XGBoost for a) short-term incidents b) long-term incidents of M7 Motorway, Sydney, Australia. One immediate observation is the fact that the data has 3 sources of reporting, and this can be seen as three different distributions in the top 1 most important feature ranked in Figure 4a). The source reporting the incidents seems to be the one factor which influence the most the incident duration. When comparing the top features for both short versus long term incidents, these are almost the same in both subsets: average temperature, the hour when the incident happened, the Sector ID, the direction of travel and the source of information that reported the incidents. Overall, for this data set, same features can be collected for both types of incidents. 

\subsubsection{San Francisco Feature Importance, U.S.A.}

\begin{figure}
\centering
\includegraphics[width=0.48\textwidth]{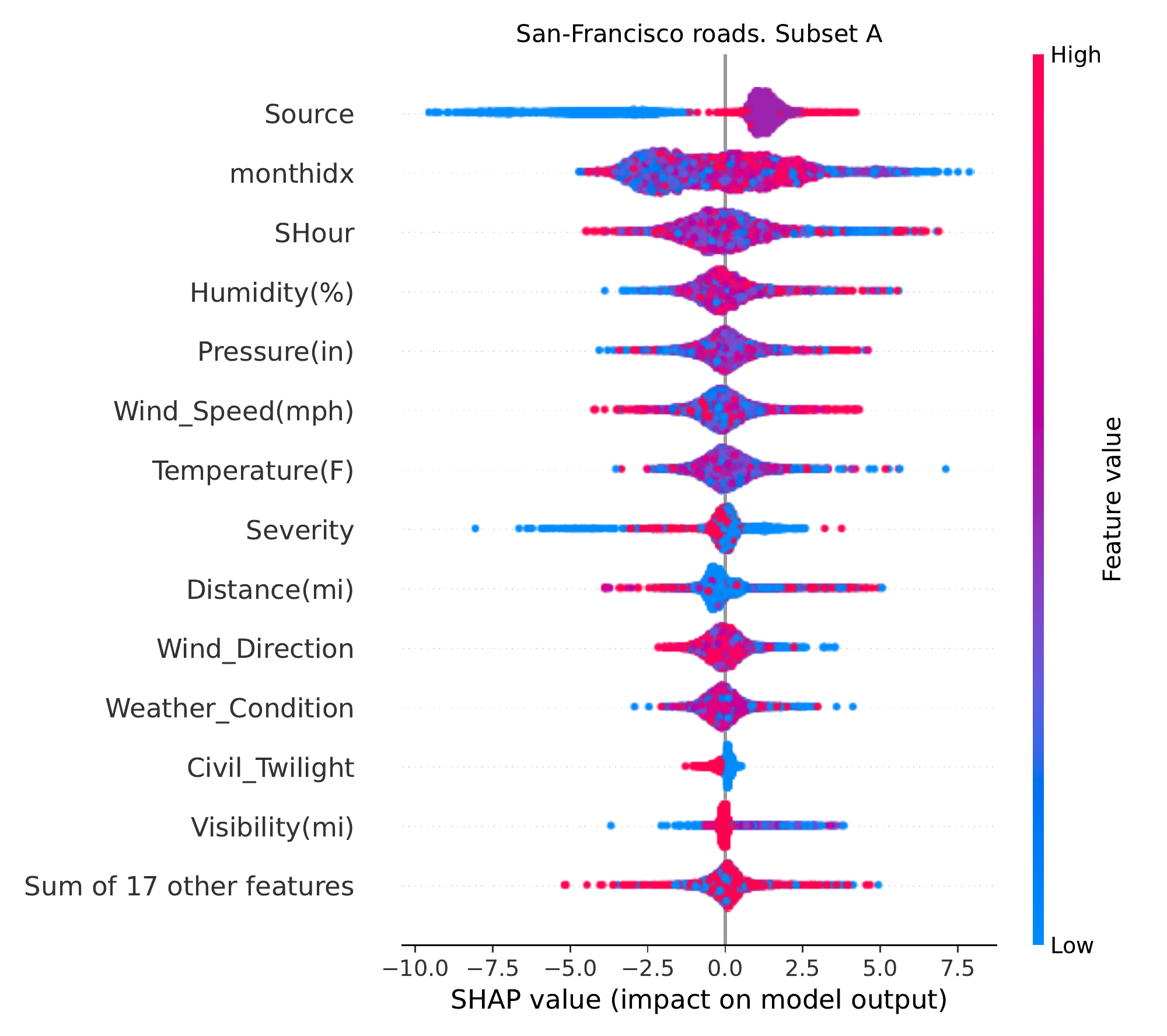}
\includegraphics[width=0.48\textwidth]{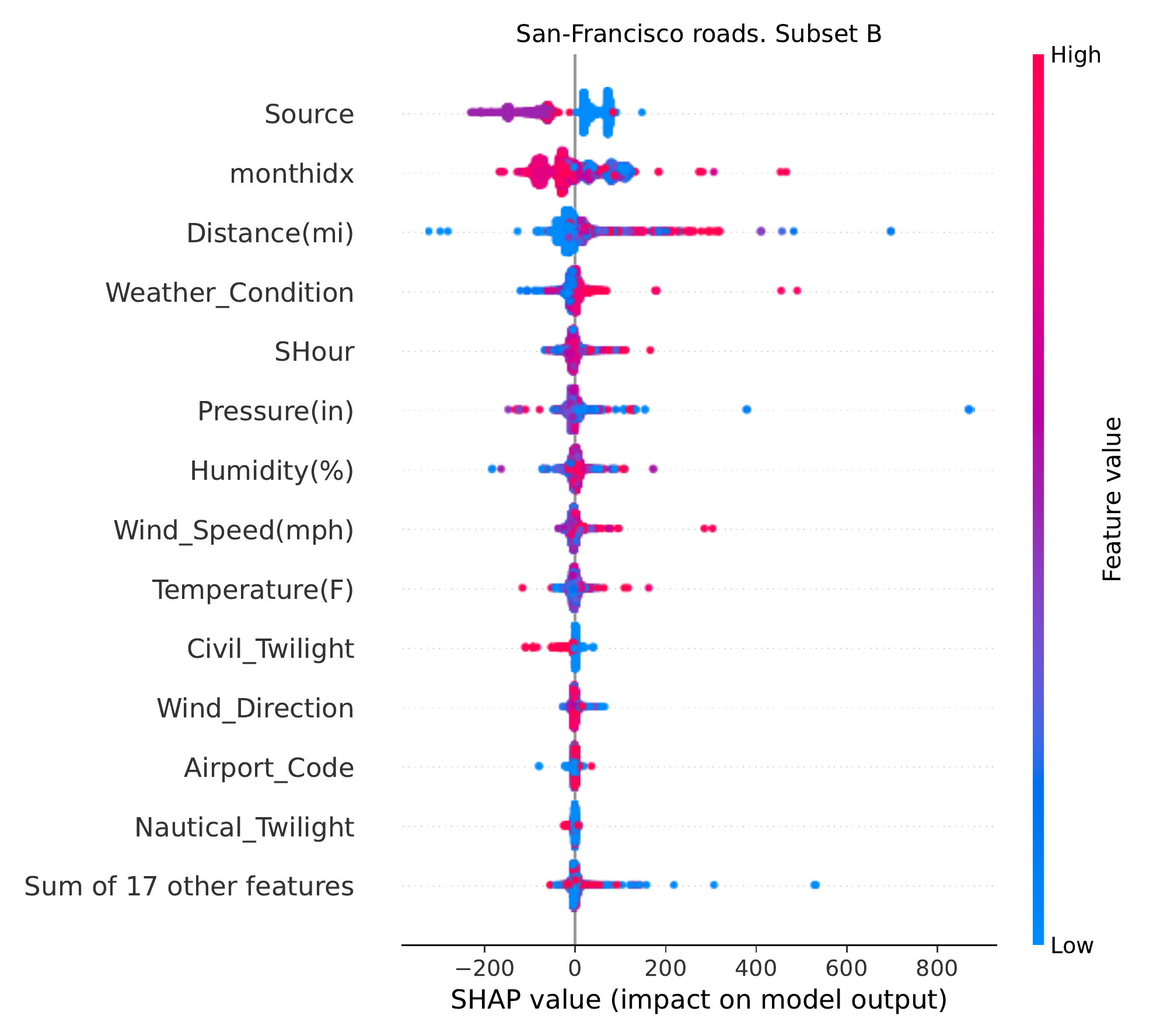}
\caption{Feature importance for All-to-All regression using XGBoost for a) short-term incidents b) long-term incidents of San-Francisco, USA}
\label{fig:impSF}
\end{figure}

Finally, \cref{fig:impSF} showcases the feature importance for All-to-All regression using XGBoost for a) short-term incidents b) long-term incidents of San-Francisco, USA. This data set is very different than the rest, but as in the case of M7 motorway, the source reporting the incident seems to be most important factor affecting the duration – this is mostly related to the way the information is received to the centre (from road users, from local traffic agents, from video camera surveillance, etc.). We observe that short-term and long-term incident are very different in their nature and incident characteristics found to have different importance in the prediction of the incident duration. For the SF data set, the most important features are Source, monthidx, Shour, regardless of the incident duration. In terms of large accidents however, the distance from the CBD is very important while for small accident the humidity plays an important factor ranking 4th (which might indicate that weather in San Francisco can cause small traffic accidents to happen often). 
Overall, despite all data sets being different, their specificity and feature important is highly related to their setup, the location of the network and the way the management centre received and handle the disruption. In order to help improve the prevention techniques more effort should be invested in understand which source of incident reporting causes the most errors overall and why.

}

%% file: sections/7-Conclusions.tex
\section{CONCLUSIONS}\label{S7-Conclusions}

This paper proposed a novel bi-level framework for predicting the incident durations via a unique combination of baseline machine learning models (for both classification and regression), together with an outlier removal procedure and a novel intra-extra joint optimisation technique. The accuracy and importance of the proposed approach have been proved via three different data sets from 2 countries (Australia and the United States of America) under several scenarios for testing and validation.

\textbf{Major contributions:}
Firstly, regarding the classification prediction of incidents into short versus long-term: we found that the optimal duration classification thresholds are similar among the three different data sets: 40min for data set AR, 45min for M, 45min for SF. Sydney TIMS also found 45 minutes to be the threshold for incident removal performance evaluation via their on-the-field expertise; this represented a confirmation that our threshold split is in coherence with realistic operational rescue times. Secondly, the best performing and robust models in the classification and regression experiments were the tree-based models (XGBoost, RandomForest, etc.). Thirdly, our extensive regression scenarios demonstrate that the short-term and long-term traffic accidents should be modelled separately. Otherwise, we will observe a drop in performance due to the adverse effect of different scale values in the training set on the model output. Fourthly, our proposed IEO-ML approach outperformed baseline ML models in 12 out of 18 cases ($66\%$), showcasing its strong value to the incident duration prediction problem. Finally, when evaluating the feature importance, we showed that features related to time, location, type of accident, reporting source and weather are among the top 10 critical features in all three data sets. By improving the precision of the most important and removing non-important features from the incident reports, TIMS can significantly improve the quality of data acquisition.

\revA{
\textbf{Limitations of this study:}
One of the biggest challenges when studying the problem of incident duration prediction represents data availability. In most cases, the privacy around traffic incidents represents the main reason why data sets are not released publicly. For example, the two data sets from Australia are private and have only been released for the purpose of this study, whereas only the San Francisco data set is made open publicly. Many other countries around the world have not yet fully released their incident logs, and this represents a challenge for this topic. However, if more incident data logs become available, they can represent a good test best for our approach.

Regarding the model performance, we make the observation that the performance of ML methods is highly affected by the data sets and the used methodology. Our approach shows a better performance for 4 of 6 methods in the case of San-Francisco, but if looked more precisely into details, KNN (where there is no improvement) produces an error that is twice as large as the best performing model (GBDT). The same is for data set A when using the LR method. And, with only GBDT left with no improvement may point to the fact that GBDT is robust to outliers and does not need outlier removal (as observed on all three data sets). As can be seen for the data set A, where MAPE is high ($80\%$), there is a very weak connection between features and the labelled data, and thus the performance for all methods is poor. Therefore, there is not much effect from the outlier removal approach on poor data sets or for methods that are weaker by design.
}

\textbf{Future research} can be related to the usage of traffic simulation with information on predicted traffic incident duration included in the decision making process during route planning. For example, the vehicle can consider that a traffic incident is short-term and assume that it will be cleared before arriving at the incident location and therefore reduce its travel time by not planning a route around the incident site. Furthermore, the cost of prediction error and the benefit of traffic accident duration estimation can be estimated from the simulation model, where occasional traffic accidents happen within traffic flow. Also, the benefit of this approach can be estimated for online route planning and not only at the time of the departure.

%% file: TRC2022_Incident_duration_pred.bbl
\begin{thebibliography}{53}
\expandafter\ifx\csname natexlab\endcsname\relax\def\natexlab#1{#1}\fi
\providecommand{\url}[1]{\texttt{#1}}
\providecommand{\href}[2]{#2}
\providecommand{\path}[1]{#1}
\providecommand{\DOIprefix}{doi:}
\providecommand{\ArXivprefix}{arXiv:}
\providecommand{\URLprefix}{URL: }
\providecommand{\Pubmedprefix}{pmid:}
\providecommand{\doi}[1]{\href{http://dx.doi.org/#1}{\path{#1}}}
\providecommand{\Pubmed}[1]{\href{pmid:#1}{\path{#1}}}
\providecommand{\bibinfo}[2]{#2}
\ifx\xfnm\relax \def\xfnm[#1]{\unskip,\space#1}\fi
\bibitem[{Abou~Elassad et~al.(2020)Abou~Elassad, Mousannif and
  Al~Moatassime}]{abou2020real}
\bibinfo{author}{Abou~Elassad, Z.E.}, \bibinfo{author}{Mousannif, H.},
  \bibinfo{author}{Al~Moatassime, H.}, \bibinfo{year}{2020}.
\newblock \bibinfo{title}{A real-time crash prediction fusion framework: An
  imbalance-aware strategy for collision avoidance systems}.
\newblock \bibinfo{journal}{Transportation research part C: emerging
  technologies} \bibinfo{volume}{118}, \bibinfo{pages}{102708}.
\bibitem[{Alkaabi et~al.(2011)Alkaabi, Dissanayake and Bird}]{Clearance2011}
\bibinfo{author}{Alkaabi, A.M.S.}, \bibinfo{author}{Dissanayake, D.},
  \bibinfo{author}{Bird, R.}, \bibinfo{year}{2011}.
\newblock \bibinfo{title}{Analyzing clearance time of urban traffic accidents
  in abu dhabi, united arab emirates, with hazard-based duration modeling
  method}.
\newblock \bibinfo{journal}{Transportation Research Record}
  \bibinfo{volume}{2229}, \bibinfo{pages}{46--54}.
\bibitem[{Ballings et~al.(2015)Ballings, {Van den Poel}, Hespeels and
  Gryp}]{BALLINGS20157046}
\bibinfo{author}{Ballings, M.}, \bibinfo{author}{{Van den Poel}, D.},
  \bibinfo{author}{Hespeels, N.}, \bibinfo{author}{Gryp, R.},
  \bibinfo{year}{2015}.
\newblock \bibinfo{title}{Evaluating multiple classifiers for stock price
  direction prediction}.
\newblock \bibinfo{journal}{Expert Systems with Applications}
  \bibinfo{volume}{42}, \bibinfo{pages}{7046--7056}.
\newblock \URLprefix
  \url{https://www.sciencedirect.com/science/article/pii/S0957417415003334},
  \DOIprefix\doi{https://doi.org/10.1016/j.eswa.2015.05.013}.
\bibitem[{Bekkerman(2015)}]{bekkerman2015present}
\bibinfo{author}{Bekkerman, R.}, \bibinfo{year}{2015}.
\newblock \bibinfo{title}{The present and the future of the kdd cup
  competition: an outsider's perspective}.
\bibitem[{Bergstra and Bengio(2012)}]{bergstra}
\bibinfo{author}{Bergstra, J.}, \bibinfo{author}{Bengio, Y.},
  \bibinfo{year}{2012}.
\newblock \bibinfo{title}{Random search for hyper-parameter optimization}.
\newblock \bibinfo{journal}{The Journal of Machine Learning Research}
  \bibinfo{volume}{13}, \bibinfo{pages}{281--305}.
\bibitem[{Breiman(2001)}]{randomforest}
\bibinfo{author}{Breiman, L.}, \bibinfo{year}{2001}.
\newblock \bibinfo{title}{Random forests}.
\newblock \bibinfo{journal}{Mach. Learn.} \bibinfo{volume}{45},
  \bibinfo{pages}{5–32}.
\newblock \URLprefix \url{https://doi.org/10.1023/A:1010933404324},
  \DOIprefix\doi{10.1023/A:1010933404324}.
\bibitem[{Breunig et~al.(2000)Breunig, Kriegel, Ng and Sander}]{lof}
\bibinfo{author}{Breunig, M.}, \bibinfo{author}{Kriegel, H.P.},
  \bibinfo{author}{Ng, R.}, \bibinfo{author}{Sander, J.}, \bibinfo{year}{2000}.
\newblock \bibinfo{title}{Lof: Identifying density-based local outliers.}, in:
  \bibinfo{booktitle}{2000 ACM SIGMOD International Conference on Management of
  Data}, pp. \bibinfo{pages}{93--104}.
\newblock \DOIprefix\doi{10.1145/342009.335388}.
\bibitem[{Chen and Guestrin(2016)}]{chen2016xgboost}
\bibinfo{author}{Chen, T.}, \bibinfo{author}{Guestrin, C.},
  \bibinfo{year}{2016}.
\newblock \bibinfo{title}{Xgboost: A scalable tree boosting system}, in:
  \bibinfo{booktitle}{Proceedings of the 22nd acm sigkdd international
  conference on knowledge discovery and data mining}, pp.
  \bibinfo{pages}{785--794}.
\bibitem[{Chen et~al.(2015)Chen, He, Benesty, Khotilovich, Tang, Cho
  et~al.}]{xgboost}
\bibinfo{author}{Chen, T.}, \bibinfo{author}{He, T.}, \bibinfo{author}{Benesty,
  M.}, \bibinfo{author}{Khotilovich, V.}, \bibinfo{author}{Tang, Y.},
  \bibinfo{author}{Cho, H.}, et~al., \bibinfo{year}{2015}.
\newblock \bibinfo{title}{Xgboost: extreme gradient boosting}.
\newblock \bibinfo{journal}{R package version 0.4-2} \bibinfo{volume}{1}.
\bibitem[{Chen et~al.(2020)Chen, Shi, Wong and Yu}]{chen2020predicting}
\bibinfo{author}{Chen, T.}, \bibinfo{author}{Shi, X.}, \bibinfo{author}{Wong,
  Y.D.}, \bibinfo{author}{Yu, X.}, \bibinfo{year}{2020}.
\newblock \bibinfo{title}{Predicting lane-changing risk level based on
  vehicles’ space-series features: A pre-emptive learning approach}.
\newblock \bibinfo{journal}{Transportation research part C: emerging
  technologies} \bibinfo{volume}{116}, \bibinfo{pages}{102646}.
\bibitem[{Chung et~al.(2011)Chung, Walubita and Choi}]{chungyou}
\bibinfo{author}{Chung, Y.}, \bibinfo{author}{Walubita, L.},
  \bibinfo{author}{Choi, K.}, \bibinfo{year}{2011}.
\newblock \bibinfo{title}{Modeling accident duration and its mitigation
  strategies on south korean freeway systems}.
\newblock \bibinfo{journal}{Transportation Research Record Journal of the
  Transportation Research Board} \bibinfo{volume}{2178},
  \bibinfo{pages}{49--57}.
\newblock \DOIprefix\doi{10.3141/2178-06}.
\bibitem[{Chung et~al.(2010)Chung, Walubita and Choi}]{chung2010modeling}
\bibinfo{author}{Chung, Y.}, \bibinfo{author}{Walubita, L.F.},
  \bibinfo{author}{Choi, K.}, \bibinfo{year}{2010}.
\newblock \bibinfo{title}{Modeling accident duration and its mitigation
  strategies on south korean freeway systems}.
\newblock \bibinfo{journal}{Transportation research record}
  \bibinfo{volume}{2178}, \bibinfo{pages}{49--57}.
\bibitem[{Dietterich(2000)}]{Dietterich2000}
\bibinfo{author}{Dietterich, T.G.}, \bibinfo{year}{2000}.
\newblock \bibinfo{title}{Ensemble methods in machine learning}, in:
  \bibinfo{booktitle}{Multiple Classifier Systems},
  \bibinfo{publisher}{Springer Berlin Heidelberg}, \bibinfo{address}{Berlin,
  Heidelberg}. pp. \bibinfo{pages}{1--15}.
\bibitem[{Fix and Hodges(1951)}]{fix1951discriminatory}
\bibinfo{author}{Fix, E.}, \bibinfo{author}{Hodges, J.}, \bibinfo{year}{1951}.
\newblock \bibinfo{title}{Discriminatory analysis, nonparametric
  discrimination}.
\newblock \bibinfo{journal}{International Statistical Review} .
\bibitem[{Friedman(2000)}]{gbdt}
\bibinfo{author}{Friedman, J.}, \bibinfo{year}{2000}.
\newblock \bibinfo{title}{Greedy function approximation: A gradient boosting
  machine}.
\newblock \bibinfo{journal}{The Annals of Statistics} \bibinfo{volume}{29}.
\newblock \DOIprefix\doi{10.1214/aos/1013203451}.
\bibitem[{Government(2017)}]{arc2}
\bibinfo{author}{Government, A.}, \bibinfo{year}{2017}.
\newblock \bibinfo{title}{Road safety}.
\newblock \URLprefix \url{https://infrastructure.gov.au/roads/safety/}.
\bibitem[{Haule et~al.(2019)Haule, Sando, Lentz, Chuan and Alluri}]{Haule}
\bibinfo{author}{Haule, H.J.}, \bibinfo{author}{Sando, T.},
  \bibinfo{author}{Lentz, R.}, \bibinfo{author}{Chuan, C.H.},
  \bibinfo{author}{Alluri, P.}, \bibinfo{year}{2019}.
\newblock \bibinfo{title}{Evaluating the impact and clearance duration of
  freeway incidents}.
\newblock \bibinfo{journal}{International Journal of Transportation Science and
  Technology} \bibinfo{volume}{8}, \bibinfo{pages}{13 -- 24}.
\newblock \URLprefix
  \url{http://www.sciencedirect.com/science/article/pii/S2046043018300522},
  \DOIprefix\doi{https://doi.org/10.1016/j.ijtst.2018.06.005}.
\bibitem[{He et~al.(2013)He, Kamarianakis, Jintanakul and
  Wynter}]{he2013incident}
\bibinfo{author}{He, Q.}, \bibinfo{author}{Kamarianakis, Y.},
  \bibinfo{author}{Jintanakul, K.}, \bibinfo{author}{Wynter, L.},
  \bibinfo{year}{2013}.
\newblock \bibinfo{title}{Incident duration prediction with hybrid tree-based
  quantile regression}, in: \bibinfo{booktitle}{Advances in dynamic network
  modeling in complex transportation systems}. \bibinfo{publisher}{Springer},
  pp. \bibinfo{pages}{287--305}.
\bibitem[{Hojati et~al.(2012)Hojati, Ferreira, Charles and Kabit}]{hoj2}
\bibinfo{author}{Hojati, A.}, \bibinfo{author}{Ferreira, L.},
  \bibinfo{author}{Charles, P.}, \bibinfo{author}{Kabit, M.},
  \bibinfo{year}{2012}.
\newblock \bibinfo{title}{Analysing freeway traffic incident duration using an
  australian data set}.
\newblock \bibinfo{journal}{Road and Transport Research} \bibinfo{volume}{21},
  \bibinfo{pages}{16--28}.
\bibitem[{Hojati] et~al.(2014)Hojati], Ferreira, Washington, Charles and
  Shobeirinejad}]{Hojati2014}
\bibinfo{author}{Hojati], A.T.}, \bibinfo{author}{Ferreira, L.},
  \bibinfo{author}{Washington, S.}, \bibinfo{author}{Charles, P.},
  \bibinfo{author}{Shobeirinejad, A.}, \bibinfo{year}{2014}.
\newblock \bibinfo{title}{Modelling total duration of traffic incidents
  including incident detection and recovery time}.
\newblock \bibinfo{journal}{Accident Analysis \& Prevention}
  \bibinfo{volume}{71}, \bibinfo{pages}{296 -- 305}.
\newblock \URLprefix
  \url{http://www.sciencedirect.com/science/article/pii/S0001457514001791},
  \DOIprefix\doi{https://doi.org/10.1016/j.aap.2014.06.006}.
\bibitem[{Hou et~al.(2013)Hou, Lao, Wang, Zhang, Zhang and Li}]{Mecha2013}
\bibinfo{author}{Hou, L.}, \bibinfo{author}{Lao, Y.}, \bibinfo{author}{Wang,
  Y.}, \bibinfo{author}{Zhang, Z.}, \bibinfo{author}{Zhang, Y.},
  \bibinfo{author}{Li, Z.}, \bibinfo{year}{2013}.
\newblock \bibinfo{title}{Modeling freeway incident response time: A
  mechanism-based approach}.
\newblock \bibinfo{journal}{Transportation Research Part C: Emerging
  Technologies} \bibinfo{volume}{28}, \bibinfo{pages}{87 -- 100}.
\newblock \URLprefix
  \url{http://www.sciencedirect.com/science/article/pii/S0968090X12001519},
  \DOIprefix\doi{https://doi.org/10.1016/j.trc.2012.12.005}.
  \bibinfo{note}{euro Transportation: selected paper from the EWGT Meeting,
  Padova, September 2009}.
\bibitem[{Ke et~al.(2017)Ke, Meng, Finley, Wang, Chen, Ma, Ye and
  Liu}]{lightgbm}
\bibinfo{author}{Ke, G.}, \bibinfo{author}{Meng, Q.}, \bibinfo{author}{Finley,
  T.}, \bibinfo{author}{Wang, T.}, \bibinfo{author}{Chen, W.},
  \bibinfo{author}{Ma, W.}, \bibinfo{author}{Ye, Q.}, \bibinfo{author}{Liu,
  T.Y.}, \bibinfo{year}{2017}.
\newblock \bibinfo{title}{Lightgbm: A highly efficient gradient boosting
  decision tree}.
\newblock \bibinfo{journal}{Advances in neural information processing systems}
  \bibinfo{volume}{30}, \bibinfo{pages}{3146--3154}.
\bibitem[{Khattak et~al.(2016)Khattak, Liu, Wali, Li and
  Ng}]{khattak2016modeling}
\bibinfo{author}{Khattak, A.J.}, \bibinfo{author}{Liu, J.},
  \bibinfo{author}{Wali, B.}, \bibinfo{author}{Li, X.}, \bibinfo{author}{Ng,
  M.}, \bibinfo{year}{2016}.
\newblock \bibinfo{title}{Modeling traffic incident duration using quantile
  regression}.
\newblock \bibinfo{journal}{Transportation Research Record}
  \bibinfo{volume}{2554}, \bibinfo{pages}{139--148}.
\bibitem[{Knapen et~al.(2014)Knapen, Bellemans, Usman, Janssens and
  Wets}]{knapen2014within}
\bibinfo{author}{Knapen, L.}, \bibinfo{author}{Bellemans, T.},
  \bibinfo{author}{Usman, M.}, \bibinfo{author}{Janssens, D.},
  \bibinfo{author}{Wets, G.}, \bibinfo{year}{2014}.
\newblock \bibinfo{title}{Within day rescheduling microsimulation combined with
  macrosimulated traffic}.
\newblock \bibinfo{journal}{Transportation Research Part C: Emerging
  Technologies} \bibinfo{volume}{45}, \bibinfo{pages}{99--118}.
\bibitem[{Kuang et~al.(2019)Kuang, Yan, Zhu, Tu and Fan}]{kuang2019predicting}
\bibinfo{author}{Kuang, L.}, \bibinfo{author}{Yan, H.}, \bibinfo{author}{Zhu,
  Y.}, \bibinfo{author}{Tu, S.}, \bibinfo{author}{Fan, X.},
  \bibinfo{year}{2019}.
\newblock \bibinfo{title}{Predicting duration of traffic accidents based on
  cost-sensitive bayesian network and weighted k-nearest neighbor}.
\newblock \bibinfo{journal}{Journal of Intelligent Transportation Systems}
  \bibinfo{volume}{23}, \bibinfo{pages}{161--174}.
\bibitem[{Lee and Wei(2010)}]{lee2010computerized}
\bibinfo{author}{Lee, Y.}, \bibinfo{author}{Wei, C.H.}, \bibinfo{year}{2010}.
\newblock \bibinfo{title}{A computerized feature selection method using genetic
  algorithms to forecast freeway accident duration times}.
\newblock \bibinfo{journal}{Computer-Aided Civil and Infrastructure
  Engineering} \bibinfo{volume}{25}, \bibinfo{pages}{132--148}.
\bibitem[{li(2014)}]{li2014}
\bibinfo{author}{li, R.}, \bibinfo{year}{2014}.
\newblock \bibinfo{title}{Traffic incident duration analysis and prediction
  models based on the survival analysis approach}.
\newblock \bibinfo{journal}{IET Intelligent Transport Systems}
  \bibinfo{volume}{9}.
\newblock \DOIprefix\doi{10.1049/iet-its.2014.0036}.
\bibitem[{Li et~al.(2015)Li, Pereira and Ben-Akiva}]{li2015competing}
\bibinfo{author}{Li, R.}, \bibinfo{author}{Pereira, F.C.},
  \bibinfo{author}{Ben-Akiva, M.E.}, \bibinfo{year}{2015}.
\newblock \bibinfo{title}{Competing risks mixture model for traffic incident
  duration prediction}.
\newblock \bibinfo{journal}{Accident Analysis \& Prevention}
  \bibinfo{volume}{75}, \bibinfo{pages}{192--201}.
\bibitem[{Li et~al.(2018)Li, Pereira and Ben-Akiva}]{LiOverview2018}
\bibinfo{author}{Li, R.}, \bibinfo{author}{Pereira, F.C.},
  \bibinfo{author}{Ben-Akiva, M.E.}, \bibinfo{year}{2018}.
\newblock \bibinfo{title}{Overview of traffic incident duration analysis and
  prediction}.
\newblock \bibinfo{journal}{European transport research review}
  \bibinfo{volume}{10}, \bibinfo{pages}{22}.
\bibitem[{Li and Shang(2014)}]{HBM2014}
\bibinfo{author}{Li, R.}, \bibinfo{author}{Shang, P.}, \bibinfo{year}{2014}.
\newblock \bibinfo{title}{Incident duration modeling using flexible parametric
  hazard-based models}.
\newblock \bibinfo{journal}{Computational intelligence and neuroscience}
  \bibinfo{volume}{2014}, \bibinfo{pages}{723427}.
\newblock \URLprefix \url{http://dx.doi.org/10.1155/2014/723427},
  \DOIprefix\doi{10.1155/2014/723427}.
\bibitem[{Li(2018)}]{li2018analysis}
\bibinfo{author}{Li, X.}, \bibinfo{year}{2018}.
\newblock \bibinfo{title}{Analysis of large-scale traffic incidents and en
  route diversions due to congestion on freeways} .
\bibitem[{Liu et~al.(2008)Liu, Ting and Zhou}]{isolation}
\bibinfo{author}{Liu, F.T.}, \bibinfo{author}{Ting, K.M.},
  \bibinfo{author}{Zhou, Z.H.}, \bibinfo{year}{2008}.
\newblock \bibinfo{title}{Isolation forest}, in: \bibinfo{booktitle}{2008
  eighth ieee international conference on data mining},
  \bibinfo{organization}{IEEE}. pp. \bibinfo{pages}{413--422}.
\bibitem[{Lopes et~al.(2013)Lopes, Bento, Pereira and
  Ben-Akiva}]{lopes2013dynamic}
\bibinfo{author}{Lopes, J.}, \bibinfo{author}{Bento, J.},
  \bibinfo{author}{Pereira, F.C.}, \bibinfo{author}{Ben-Akiva, M.},
  \bibinfo{year}{2013}.
\newblock \bibinfo{title}{Dynamic forecast of incident clearance time using
  adaptive artificial neural network models}.
\bibitem[{Lu(2021)}]{lu2021detecting}
\bibinfo{author}{Lu, Y.C.}, \bibinfo{year}{2021}.
\newblock \bibinfo{title}{Detecting outliers for improving the quality of
  incident duration prediction} .
\bibitem[{Lundberg and Lee(2017)}]{lundberg2017unified}
\bibinfo{author}{Lundberg, S.}, \bibinfo{author}{Lee, S.I.},
  \bibinfo{year}{2017}.
\newblock \bibinfo{title}{A unified approach to interpreting model
  predictions}.
\newblock \bibinfo{journal}{arXiv preprint arXiv:1705.07874} .
\bibitem[{Ma et~al.(2017)Ma, Ding, Luan, Wang and Wang}]{ma2017prioritizing}
\bibinfo{author}{Ma, X.}, \bibinfo{author}{Ding, C.}, \bibinfo{author}{Luan,
  S.}, \bibinfo{author}{Wang, Y.}, \bibinfo{author}{Wang, Y.},
  \bibinfo{year}{2017}.
\newblock \bibinfo{title}{Prioritizing influential factors for freeway incident
  clearance time prediction using the gradient boosting decision trees method}.
\newblock \bibinfo{journal}{IEEE Transactions on Intelligent Transportation
  Systems} \bibinfo{volume}{18}, \bibinfo{pages}{2303--2310}.
\bibitem[{Mao et~al.(2021)Mao, Mih{\u{a}}it{\u{a}}, Chen and
  Vu}]{mao2021boosted}
\bibinfo{author}{Mao, T.}, \bibinfo{author}{Mih{\u{a}}it{\u{a}}, A.S.},
  \bibinfo{author}{Chen, F.}, \bibinfo{author}{Vu, H.L.}, \bibinfo{year}{2021}.
\newblock \bibinfo{title}{Boosted genetic algorithm using machine learning for
  traffic control optimization}.
\newblock \bibinfo{journal}{IEEE Transactions on Intelligent Transportation
  Systems} .
\bibitem[{Mihaita et~al.(2019)Mihaita, Liu, Cai and Rizoiu}]{Arterial2019}
\bibinfo{author}{Mihaita, A.S.}, \bibinfo{author}{Liu, Z.},
  \bibinfo{author}{Cai, C.}, \bibinfo{author}{Rizoiu, M.},
  \bibinfo{year}{2019}.
\newblock \bibinfo{title}{Arterial incident duration prediction using a
  bi-level framework of extreme gradient-tree boosting}.
\newblock \bibinfo{journal}{CoRR} \bibinfo{volume}{abs/1905.12254}.
\newblock \URLprefix \url{http://arxiv.org/abs/1905.12254},
  \href{http://arxiv.org/abs/1905.12254}{\tt arXiv:1905.12254}.
\bibitem[{Moosavi et~al.(2019)Moosavi, Samavatian, Parthasarathy and
  Ramnath}]{moosavi2019countrywide}
\bibinfo{author}{Moosavi, S.}, \bibinfo{author}{Samavatian, M.H.},
  \bibinfo{author}{Parthasarathy, S.}, \bibinfo{author}{Ramnath, R.},
  \bibinfo{year}{2019}.
\newblock \bibinfo{title}{A countrywide traffic accident dataset}.
\newblock \bibinfo{journal}{arXiv preprint arXiv:1906.05409} .
\bibitem[{Nguyen et~al.(2017)Nguyen, Cai and Chen}]{Nguyen2017}
\bibinfo{author}{Nguyen, H.}, \bibinfo{author}{Cai, C.}, \bibinfo{author}{Chen,
  F.}, \bibinfo{year}{2017}.
\newblock \bibinfo{title}{Automatic classification of traffic incident's
  severity using machine learning approaches}.
\newblock \bibinfo{journal}{IET Intelligent Transport Systems}
  \bibinfo{volume}{11}, \bibinfo{pages}{615--623}.
\bibitem[{Schrank and Lomax(2002)}]{arc1}
\bibinfo{author}{Schrank, D.}, \bibinfo{author}{Lomax, T.},
  \bibinfo{year}{2002}.
\newblock \bibinfo{title}{The 2002 urban mobility report (college station, tx:
  Texas transportation institute, texas a\&m university, june)}.
\bibitem[{Shafiei et~al.(2020)Shafiei, Mihaita, Nguyen, Bentley and
  Cai}]{shafiei2020short}
\bibinfo{author}{Shafiei, S.}, \bibinfo{author}{Mihaita, A.},
  \bibinfo{author}{Nguyen, H.}, \bibinfo{author}{Bentley, C.},
  \bibinfo{author}{Cai, C.}, \bibinfo{year}{2020}.
\newblock \bibinfo{title}{Short-term traffic prediction under non-recurrent
  incident conditions integrating data-driven models and traffic simulation},
  in: \bibinfo{booktitle}{Transportation Research Board 99th Annual Meeting},
  pp.~\bibinfo{pages}{--}.
\bibitem[{Smith and Smith(2001)}]{smith2001forecasting}
\bibinfo{author}{Smith, K.}, \bibinfo{author}{Smith, B.}, \bibinfo{year}{2001}.
\newblock \bibinfo{title}{Forecasting the Clearance Time of Freeway Accidents
  Final report of ITS Center project: Incident Duration Forecasting}.
\newblock \bibinfo{type}{Technical Report}. Smart Travel Lab Report No.
  STL-2001-01.
\bibitem[{Sullivan(1997)}]{sullivan1997new}
\bibinfo{author}{Sullivan, E.C.}, \bibinfo{year}{1997}.
\newblock \bibinfo{title}{New model for predicting freeway incidents and
  incident delays}.
\newblock \bibinfo{journal}{Journal of Transportation Engineering}
  \bibinfo{volume}{123}, \bibinfo{pages}{267--275}.
\bibitem[{of~the United States Department~of Transportation(2017)}]{mutcd}
\bibinfo{author}{of~the United States Department~of Transportation, F.H.A.},
  \bibinfo{year}{2017}.
\newblock \bibinfo{title}{The manual on uniform traffic control devices}.
\newblock \URLprefix
  \url{https://mutcd.fhwa.dot.gov/htm/2009/part6/part6i.htm}.
\bibitem[{Valenti et~al.(2010)Valenti, Lelli and
  Cucina}]{valenti2010comparative}
\bibinfo{author}{Valenti, G.}, \bibinfo{author}{Lelli, M.},
  \bibinfo{author}{Cucina, D.}, \bibinfo{year}{2010}.
\newblock \bibinfo{title}{A comparative study of models for the incident
  duration prediction}.
\newblock \bibinfo{journal}{European Transport Research Review}
  \bibinfo{volume}{2}, \bibinfo{pages}{103--111}.
\bibitem[{Wali et~al.()Wali, Khattak and Liu}]{waliheterogeneity}
\bibinfo{author}{Wali, B.}, \bibinfo{author}{Khattak, A.J.},
  \bibinfo{author}{Liu, J.}, .
\newblock \bibinfo{title}{Heterogeneity assessment in incident duration
  modelling: Implications for development of practical strategies for small \&
  large scale incidents} .
\bibitem[{Wen et~al.(2018)Wen, Mih{\u{a}}i{\c{t}}{\u{a}}, Nguyen, Cai and
  Chen}]{wen2018integrated}
\bibinfo{author}{Wen, T.}, \bibinfo{author}{Mih{\u{a}}i{\c{t}}{\u{a}}, A.S.},
  \bibinfo{author}{Nguyen, H.}, \bibinfo{author}{Cai, C.},
  \bibinfo{author}{Chen, F.}, \bibinfo{year}{2018}.
\newblock \bibinfo{title}{Integrated incident decision-support using traffic
  simulation and data-driven models}.
\newblock \bibinfo{journal}{Transportation research record}
  \bibinfo{volume}{2672}, \bibinfo{pages}{247--256}.
\bibitem[{Wen et~al.(2013)Wen, Chen, Xiong, Han and Chen}]{wen2013traffic}
\bibinfo{author}{Wen, Y.}, \bibinfo{author}{Chen, S.Y.},
  \bibinfo{author}{Xiong, Q.Y.}, \bibinfo{author}{Han, R.B.},
  \bibinfo{author}{Chen, S.Y.}, \bibinfo{year}{2013}.
\newblock \bibinfo{title}{Traffic incident duration prediction based on
  k-nearest neighbor}, in: \bibinfo{booktitle}{Applied Mechanics and
  Materials}, \bibinfo{organization}{Trans Tech Publ}. pp.
  \bibinfo{pages}{1675--1681}.
\bibitem[{Yi et~al.(2019)Yi, Su, Liu, Quddus and Chen}]{yi2019machine}
\bibinfo{author}{Yi, D.}, \bibinfo{author}{Su, J.}, \bibinfo{author}{Liu, C.},
  \bibinfo{author}{Quddus, M.}, \bibinfo{author}{Chen, W.H.},
  \bibinfo{year}{2019}.
\newblock \bibinfo{title}{A machine learning based personalized system for
  driving state recognition}.
\newblock \bibinfo{journal}{Transportation Research Part C: Emerging
  Technologies} \bibinfo{volume}{105}, \bibinfo{pages}{241--261}.
\bibitem[{Yu and Xia(2012)}]{yubin}
\bibinfo{author}{Yu, B.}, \bibinfo{author}{Xia, Z.}, \bibinfo{year}{2012}.
\newblock \bibinfo{title}{A methodology for freeway incident duration
  prediction using computerized historical database}, in:
  \bibinfo{booktitle}{The Twelfth COTA International Conference of
  Transportation Professionals}, pp. \bibinfo{pages}{3463--3474}.
\newblock \DOIprefix\doi{10.1061/9780784412442.351}.
\bibitem[{Zhan et~al.(2011)Zhan, Gan and Hadi}]{zhan2}
\bibinfo{author}{Zhan, C.}, \bibinfo{author}{Gan, A.}, \bibinfo{author}{Hadi,
  M.}, \bibinfo{year}{2011}.
\newblock \bibinfo{title}{Prediction of lane clearance time of freeway
  incidents using the m5p tree algorithm}.
\newblock \bibinfo{journal}{IEEE Transactions on Intelligent Transportation
  Systems} \bibinfo{volume}{12}, \bibinfo{pages}{1549--1557}.
\newblock \DOIprefix\doi{10.1109/TITS.2011.2161634}.
\bibitem[{Zou et~al.(2018)Zou, Ye, Henrickson, Tang and Wang}]{Zou1}
\bibinfo{author}{Zou, Y.}, \bibinfo{author}{Ye, X.},
  \bibinfo{author}{Henrickson, K.}, \bibinfo{author}{Tang, J.},
  \bibinfo{author}{Wang, Y.}, \bibinfo{year}{2018}.
\newblock \bibinfo{title}{Jointly analyzing freeway traffic incident clearance
  and response time using a copula-based approach}.
\newblock \bibinfo{journal}{Transportation Research Part C: Emerging
  Technologies} \bibinfo{volume}{86}, \bibinfo{pages}{171--182}.
\newblock \URLprefix
  \url{https://www.sciencedirect.com/science/article/pii/S0968090X17303108},
  \DOIprefix\doi{https://doi.org/10.1016/j.trc.2017.11.004}.

\end{thebibliography}
